\PassOptionsToPackage{table}{xcolor}
\documentclass[]{afu}
\usepackage{amsmath,amssymb,amsthm}
\usepackage{fontawesome5}
\definecolor{primary}{HTML}{1B2A4A}
\usepackage{fix-cm}

\usepackage{xcolor}
\usepackage{multirow}
\usepackage{colortbl}
\usepackage{tabularx}
\usepackage{longtable}
\usepackage{xltabular}
\usepackage{tcolorbox}
\usepackage{enumitem}
\usepackage{pifont}
\usepackage{wrapfig}
\usepackage{needspace}
\makeatletter
\newcommand{\finishfigurewrap}{\par
  \ifnum\c@WF@wrappedlines>1
    \vspace{\dimexpr\baselineskip*\c@WF@wrappedlines-\baselineskip\relax}%
  \fi
  \WFclear}
\makeatother
\usepackage{url}
\usepackage{makecell}
\usepackage{float}
\usepackage{tikz}
\usetikzlibrary{arrows.meta,positioning,fit}

\definecolor{mygray}{gray}{0.9}
\definecolor{myblue}{HTML}{F0FFFF}
\definecolor{myblue_light}{HTML}{EAF6FF}
\definecolor{myblue_dark}{HTML}{2B57A0}
\definecolor{myblue_lightv1}{HTML}{F2F7E0}
\definecolor{mygreen}{RGB}{60, 179, 113}
\definecolor{myyellow_light}{RGB}{255, 255, 224}
\definecolor{lightgreen}{HTML}{D3E5C7}
\definecolor{deepgreen}{RGB}{46,139,87}
\definecolor{mydarkgreen}{rgb}{0,0.6,0}

\definecolor{promptpurple}{HTML}{70528F}
\definecolor{promptpurplelight}{HTML}{F7F3FB}
\definecolor{promptpurpletitle}{HTML}{E8DDF3}
\newtcolorbox[auto counter, number within=section]{prompt}[1][]{
    breakable,
    colback=promptpurplelight,
    colframe=promptpurple,
    colbacktitle=promptpurpletitle,
    coltitle=promptpurple,
    fonttitle=\bfseries,
    fontupper=\normalsize,
    boxsep=3pt,
    before skip=4pt, after skip=4pt,
    left=2pt,
    right=2pt,
    top=2pt,
    bottom=2pt,
    boxrule=1pt,
    #1,
}

\usepackage[ruled,vlined]{algorithm2e}
\definecolor{MedEvoAlgorithmBlue}{HTML}{287DB1}

\newcommand{\targetmodel}{Qwen3.5-35B-A3B}
\newcommand{\targetparams}{35B (3B active)}

\DeclareRobustCommand{\medicalmark}{\textcolor[HTML]{287DB1}{\faBriefcaseMedical}}
\DeclareRobustCommand{\repairmark}{\textcolor[HTML]{B87333}{\faWrench}}
\newcommand{\modelmarks}[2]{\makebox[1.05em][c]{#1}\makebox[1.05em][c]{#2}\hspace{0.2em}}

\newcommand{\tablechange}[1]{{\scriptsize #1}}

\title{MedEvo: Medical Reasoning Evolution through Knowledge Navigation}
\renewcommand{\titlelist}{%
  \makebox[\linewidth][c]{{\Large\bfseries
  \llap{\raisebox{-0.5\height}{\includegraphics[height=2.1em]{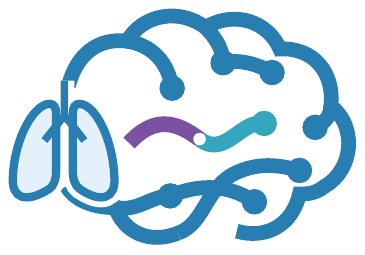}}\hspace{0.7em}}%
  \raisebox{-0.5\height}{\shortstack[c]{MedEvo: Medical Reasoning Evolution\\[3pt]
through Knowledge Navigation}}}}%
}
\renewcommand{\authorlist}{%
  {\small\normalfont\centering
  \mbox{\mbox{\href{https://openreview.net/profile?id=\string~Wang_Xiaoliang2}{\textcolor{black}{Xiaoliang Wang}}\textsuperscript{1,2,*}},\hspace{0.25em}\mbox{\href{https://openreview.net/profile?id=\string~Zhuo_Chen3}{\textcolor{black}{Zhuo Chen}}\textsuperscript{2,*}},\hspace{0.25em}\mbox{\href{https://openreview.net/profile?id=\string~Chuanyang_Zheng2}{\textcolor{black}{Chuanyang Zheng}}\textsuperscript{2}},\hspace{0.25em}\mbox{\href{https://openreview.net/profile?id=\string~Quanbin_Wang1}{\textcolor{black}{Quanbin Wang}}\textsuperscript{2}},\hspace{0.25em}\mbox{\href{https://openreview.net/profile?id=\string~Jiangwei_Lao1}{\textcolor{black}{Jiangwei Lao}}\textsuperscript{2}},\hspace{0.25em}\mbox{\href{https://openreview.net/profile?id=\string~Yingao_Liu1}{\textcolor{black}{Yingao Liu}}\textsuperscript{2}}}\\[2pt]
  \mbox{\mbox{\href{https://openreview.net/profile?id=\string~Jiajia_Liu1}{\textcolor{black}{Jiajia Liu}}\textsuperscript{2}},\hspace{0.25em}\mbox{\href{https://openreview.net/profile?id=\string~Jinming_Cao1}{\textcolor{black}{Jinming Cao}}\textsuperscript{2}},\hspace{0.25em}\mbox{\href{https://openreview.net/profile?id=\string~Honglin_Lin2}{\textcolor{black}{Honglin Lin}}\textsuperscript{3,4}},\hspace{0.25em}\mbox{\href{https://openreview.net/profile?id=\string~Zhihong_Lu2}{\textcolor{black}{Zhihong Lu}}\textsuperscript{2}},\hspace{0.25em}\mbox{\href{https://openreview.net/profile?id=\string~Hao_Jin4}{\textcolor{black}{Hao Jin}}\textsuperscript{1}},\hspace{0.25em}\mbox{\href{https://openreview.net/profile?id=\string~Jian_Wang25}{\textcolor{black}{Jian Wang}}\textsuperscript{2}},\hspace{0.25em}\mbox{Jinjie Gu\textsuperscript{2}},\hspace{0.25em}\mbox{\href{https://openreview.net/profile?id=\string~Lijun_Wu1}{\textcolor{black}{Lijun Wu}}\textsuperscript{2,3,\ensuremath{\dagger}}}}\par
  }%
}
\renewcommand{\affiliationlist}{%
  {\small\normalfont\centering
  \textsuperscript{1}Zhejiang University\quad
  \textsuperscript{2}Ant Group, AFu Team\\[1pt]
  \textsuperscript{3}Shanghai Artificial Intelligence Laboratory\quad
  \textsuperscript{4}Shanghai Jiao Tong University\par}%
}
\renewcommand{\contributionlist}{%
  {\footnotesize\normalfont\centering\vspace{1mm}
  \textsuperscript{*}Equal contribution\quad
  \textsuperscript{\ensuremath{\dagger}}Corresponding author\par}%
}
\renewcommand{\metadatalist}{%
  {\small\begin{tabularx}{\linewidth}{@{}l@{\enspace}X@{}}
  \textbf{Correspondence:} & Lijun Wu, \email{wlj554215@antgroup.com}\\
  \textbf{Project:} & \url{https://pork-pay.github.io/MedEvo/}\\
  \textbf{Code:} & \url{https://github.com/pork-pay/MedEvo}
  \end{tabularx}}%
}
\hypersetup{pdftitle={MedEvo: Medical Reasoning Evolution through Knowledge Navigation},
  pdfauthor={Xiaoliang Wang, Zhuo Chen, Chuanyang Zheng, Quanbin Wang, Jiangwei Lao, Yingao Liu, Jiajia Liu, Jinming Cao, Honglin Lin, Zhihong Lu, Hao Jin, Jian Wang, Jinjie Gu, Lijun Wu}}

\abstract{
Medical multimodal reasoning models are typically trained on \textbf{teacher-generated rationales}, discarding the valid reasoning in students' own failed trajectories, while general-domain correction methods offer little guidance for errors in image reading, anatomical context, or differential diagnosis. We propose \textbf{MedEvo}, which uses \textbf{medical knowledge}, namely a tag taxonomy and error-type diagnosis, to navigate both \textbf{trajectory repair and data selection}. For each failed rollout, a \textbf{gold-blind auditor} locates the first consequential error and provides a minimal, answer-free hint; the same student \textbf{retains the valid prefix and regenerates the erroneous suffix}. After each round, MedEvo identifies capability gaps across medical tags on a \textbf{held-out feedback set} and prioritizes related questions for new rollouts and repairs. With \textbf{supervised fine-tuning (SFT) alone}, MedEvo raises Qwen3.5-9B mean accuracy across seven benchmarks from \textbf{65.57\% to 69.12\%} (47.40\% to \textbf{54.50\%} on the multimodal track of MedXpertQA (MedXpertQA-MM)), surpassing teacher-distillation SFT (66.93\%). It also raises Qwen3.5-35B and Qwen3.5-397B mean accuracy from 67.91\% to \textbf{73.04\%} and from 71.83\% to \textbf{74.95\%}, respectively. Our analyses show that failed trajectories contain reasoning worth preserving, that \textbf{repair success does not imply learning value}, and that \textbf{independent feedback is key to sustained improvement}.
}

\let\paperabstracttext\abstractlist
\renewcommand{\abstractlist}{%
  {\setlength{\parskip}{0pt}\vspace{3mm}
   {\centering\bfseries Abstract\par}
   \vspace{2mm}\paperabstracttext\par}%
}

\begin{document}
\maketitle

\section{Introduction}
\label{sec:introduction}

\begin{figure}[!t]
    \centering
    \includegraphics[width=\linewidth]{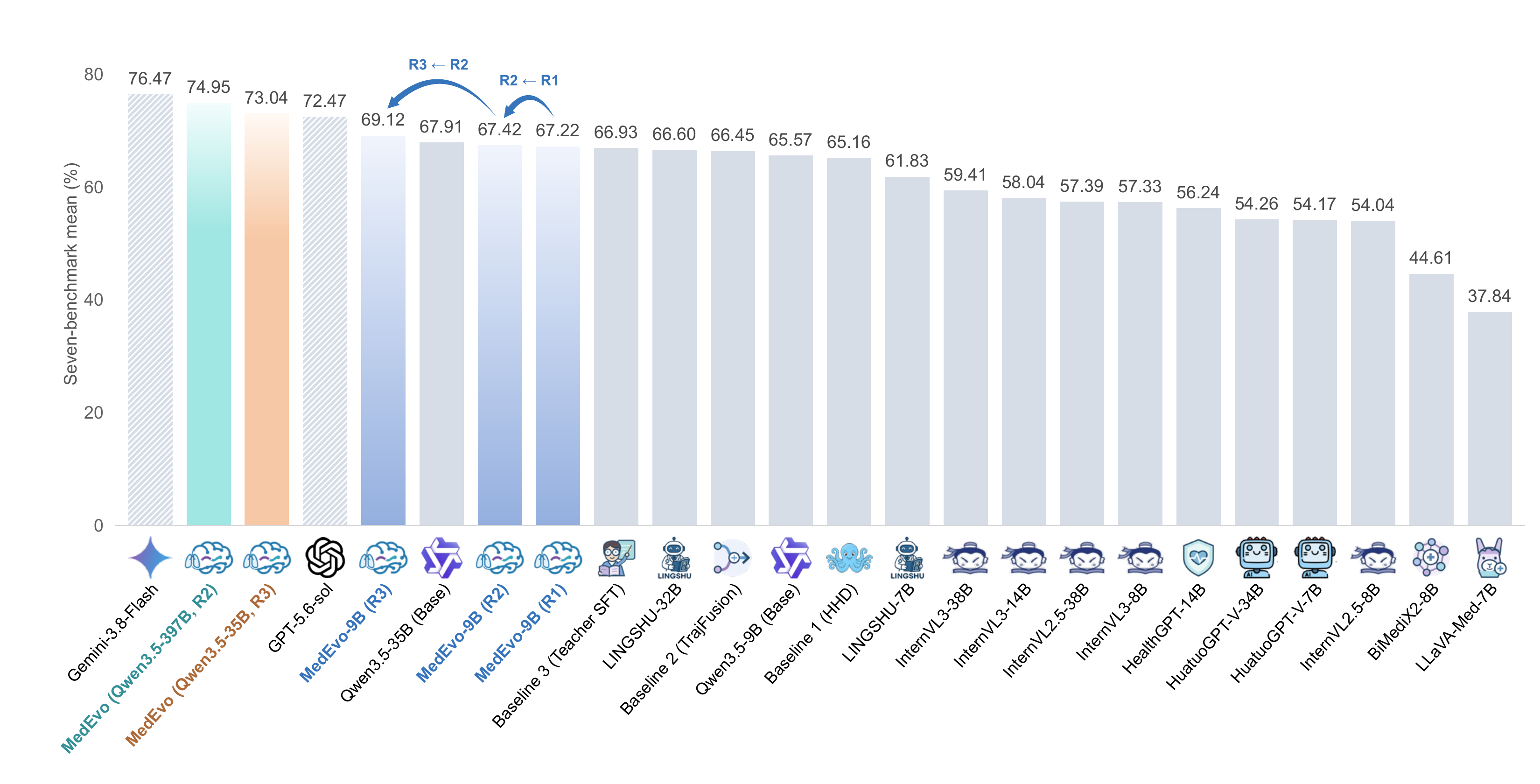}
    \caption{\textbf{Medical multimodal reasoning across model scales.} Mean accuracy on seven benchmarks. MedEvo improves 9B/35B/397B Qwen3.5 models. B: billion parameters; Hindsight Hint Distillation (HHD)~\citep{DBLP:journals/corr/abs-2605-11556}.}
    \label{fig:intro_benchmark_overview}
\end{figure}

Medical multimodal reasoning requires connecting visual findings with anatomical context, disease knowledge, and clinical decisions, and expert-level questions remain challenging for current medical vision-language models~\citep{li2023llavamed,moor2023medflamingo,DBLP:conf/icml/ZuoQLCZHZ0025}. Despite recognizing several findings, a model can fail through image misreading, overlooked clinical clues, or unsupported differential diagnosis. Such \textbf{failures are often local}: the reasoning before the mistake can remain valid, so preserving valid steps and recovering from the point of failure directly improves medical reasoning~\citep{tyen2023errors,an2023lema}.

Existing approaches typically use supervised fine-tuning (SFT) to train students on teacher-generated rationales or on answer-correct student rollouts~\citep{zelikman2022star,singh2023restem,li2026rethinkingopd}. Discarding a failed trajectory loses its valid reasoning, while replacing it with a complete teacher solution gives the student a path it did not produce. Methods that learn from failures, such as mistake--correction pairs~\citep{an2023lema}, trajectory fusion~\citep{DBLP:conf/acl/DengTLLWLX26}, and Hindsight Hint Distillation (HHD)~\citep{DBLP:journals/corr/abs-2605-11556}, were developed for mathematics and software engineering, and offer little guidance when an error depends on medical images, anatomy, or differential diagnosis. Moreover, a model's failures reveal which medical capabilities need further training, yet this signal is rarely used to guide the next round of supervision. We therefore ask two questions: how can we repair reasoning while \textbf{preserving valid steps}, and how can \textbf{failures guide subsequent learning}?

\begin{figure}[htbp]
    \centering
    \includegraphics[width=\linewidth]{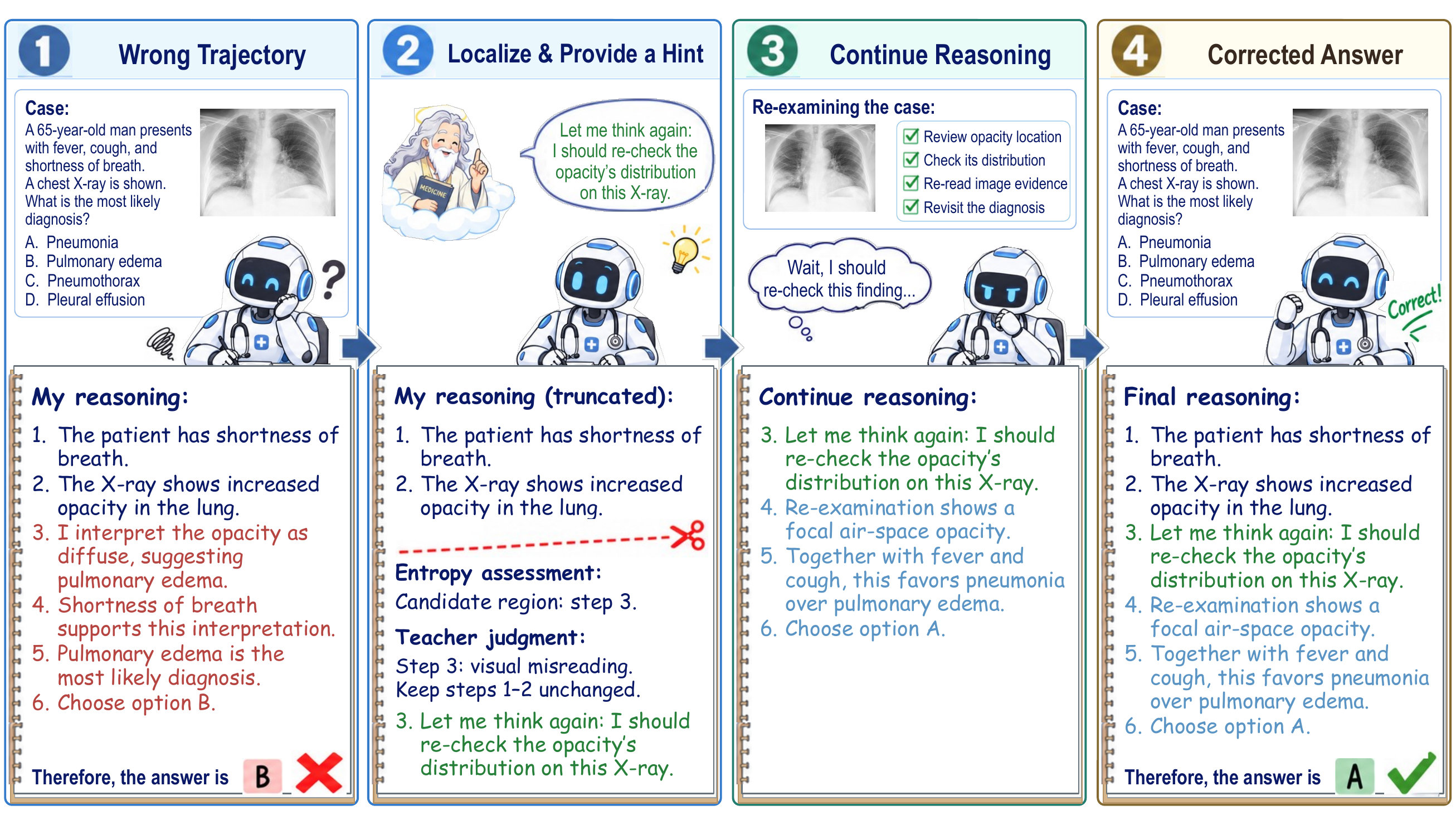}
    \caption{\textbf{Local trajectory repair.} A minimal, answer-blind hint guides the student to recover from a failed reasoning step while retaining its supported prefix. Localization and admission checks are detailed in \S\,\ref{sec:main-localization} and \S\,\ref{sec:main-admission}.}
    \label{fig:repair_intuition}
\end{figure}

We introduce MedEvo through the principle of \textbf{Knowledge Navigation}: using structured medical knowledge and failure diagnosis to guide \textbf{where to intervene, how to recover, and what to learn next}. These questions motivate local recovery decisions and learning decisions reflecting the student's remaining weaknesses. Medical tags locate a failure in its clinical context, while error types identify the reasoning operation that needs correction. We call this navigation because these signals direct a path from an observed failure to a targeted intervention, at two connected levels: \textbf{within a trajectory} and \textbf{across training rounds} (Fig.~\ref{fig:method}). For a failed rollout, a gold-blind auditor identifies the first consequential error, preserves the valid prefix, and provides a \textbf{minimal, answer-free hint}; the same student then regenerates the erroneous suffix, so the correction remains grounded in its own reasoning (Fig.~\ref{fig:repair_intuition}). After each round, MedEvo measures performance and errors across medical tags on a \textbf{held-out feedback set} and prioritizes questions for fresh rollouts and repairs, so that recurring failures shape subsequent training.

Across seven medical multimodal benchmarks (Fig.~\ref{fig:intro_benchmark_overview}), MedEvo improves the mean accuracy of Qwen3.5-9B, Qwen3.5-35B, and Qwen3.5-397B from 65.57\%, 67.91\%, and 71.83\% to \textbf{69.12\%, 73.04\%, and 74.95\%}, respectively. With Qwen3.5-9B, it outperforms teacher-distillation SFT (69.12\% vs.\ 66.93\%); with Qwen3.5-35B and Qwen3.5-397B, it surpasses GPT-5.6-sol (72.47\%); Gemini-3.8-Flash has the highest mean accuracy in this comparison (76.47\%).

Our contributions are threefold:
\begin{itemize}[leftmargin=*,nosep]
    \item \textbf{Medically informed local repair.} Within trajectories, Knowledge Navigation uses a medical tag taxonomy and error-type diagnosis to locate the first consequential error and a safe rollback boundary; a gold-blind auditor provides an answer-free hint for same-student suffix regeneration that preserves the valid prefix.
    \item \textbf{Feedback-driven data evolution.} Across rounds, Knowledge Navigation uses feedback: capability gaps across medical tags and error types on a held-out feedback set target questions and failures for rollout and repair.
    \item \textbf{Empirical gains and findings.} With SFT alone, MedEvo improves Qwen3.5-9B, Qwen3.5-35B, and Qwen3.5-397B across seven benchmarks. Our analyses show that failed trajectories contain reasoning worth preserving, that repair success does not imply learning value, and that independent feedback is key to sustained improvement.
\end{itemize}

\begin{figure}[!t]
    \centering
    \includegraphics[width=\linewidth]{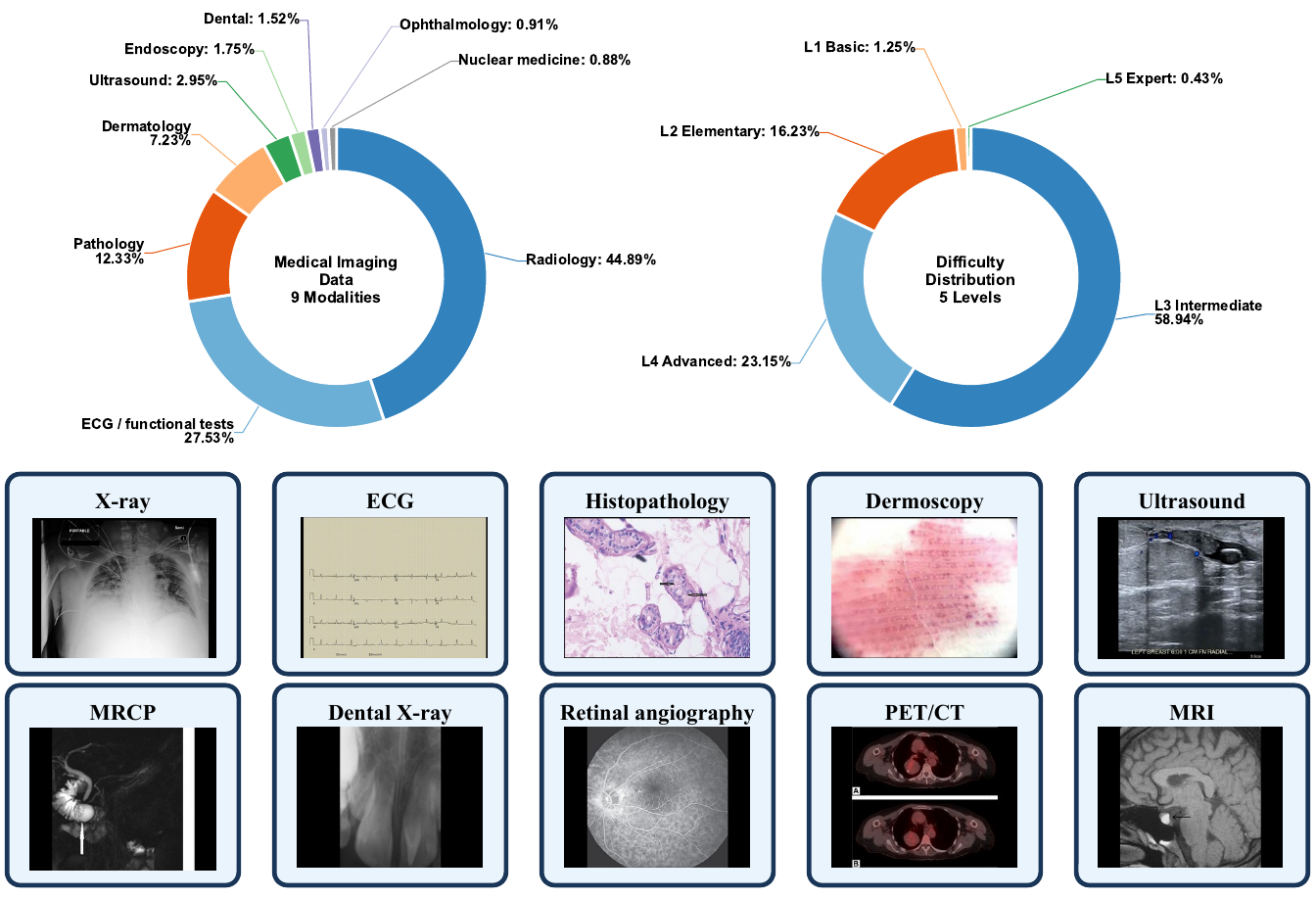}
    \caption{\textbf{Medical multimodal data.} Top: modality and difficulty; bottom: examples. Electrocardiogram (ECG); magnetic resonance cholangiopancreatography (MRCP); positron emission tomography/computed tomography (PET/CT); magnetic resonance imaging (MRI). Examples include both anatomical images and physiological recordings.}
    \label{fig:dataset_overview}
\end{figure}

\Needspace{8\baselineskip}
\section{Related Work}
\label{sec:related_work}
\subsection{Medical Multimodal Reasoning and Supervision Quality}
\label{sec:related_medical}\label{sec:related_taxonomy}
Medical vision-language models use biomedical instruction tuning~\citep{li2023llavamed,moor2023medflamingo}, generalist training in Lingshu~\citep{xu2025lingshu}, and bilingual BiMediX2 training~\citep{mullappilly2024bimedix2biomedicalexpertlmm}. MedGemma builds medical vision-language models on Gemma~3~\citep{sellergren2025medgemma}. Structured knowledge can also be integrated with multimodal learning through knowledge graphs~\citep{DBLP:journals/corr/abs-2402-05391}. Beyond answer accuracy, \textbf{process-level audits} separate visual recognition, knowledge application, and evidence integration~\citep{DBLP:journals/corr/abs-2505-11462,DBLP:journals/corr/abs-2605-28301}, and rubric-based evaluations can miss hallucinations, motivating \textbf{factuality checks}~\citep{farrow2026rubricsfail}.

Supervision quality depends on \textbf{reasoning content and student compatibility}: process supervision provides step-level feedback~\citep{ICLR2024_aca97732}, on-policy distillation studies model compatibility~\citep{li2026rethinkingopd}, and ChainPrune~\citep{pan2026chainprune} reduces redundant reasoning while distinguishing structural efficiency from token length. JOLT makes student compatibility explicit: even a higher-performing privileged teacher can induce an unhelpful student update, motivating joint teacher--student training with regularization toward the student's current policy~\citep{ardywibowo2026goodselfteachermeetsstudent}. MedEvo diagnoses failures with a medical codebook, screening repairs for answer correctness, valid boundaries, integrity, and student-relative learnability.

\subsection{Learning from Failures and Localized Repair}
\label{sec:related_correction}\label{sec:related_trajectories}
Failed trajectories can supply supervision: Learning from Mistakes (LEMA) trains on mistake--correction pairs~\citep{an2023lema}, TrajFusion combines incorrect and correct trajectories~\citep{DBLP:conf/acl/DengTLLWLX26}, and HHD guides recovery with concise hints~\citep{DBLP:journals/corr/abs-2605-11556}. Identifying an error can enable its correction~\citep{tyen2023errors}, and backtracking distillation and selective regeneration study where correction should begin~\citep{DBLP:journals/corr/abs-2605-19433,DBLP:journals/corr/abs-2608-24338}. MedEvo differs in three respects: the corrected suffix is \textbf{generated by the same student} rather than written by a teacher; the auditor \textbf{never accesses the reference answer}; and both localization and hints are conditioned on medical tags and error types. Self-Retrospection Distillation uses completed interactions, including failed attempts, to supervise pre-interaction predictions of required knowledge and likely pitfalls in the same policy~\citep{zhang2026selfretrospectiondistillationturningposthoc}. MedEvo uses the repaired trajectory as its training target: it retains the supported prefix and learns from the student's regenerated continuation.

\subsection{Data Evolution and Recursive Self-Improvement}
\label{sec:related_rsi}
Self-Taught Reasoner (STaR) and Reinforced Self-Training with Expectation--Maximization (ReST$^{\mathrm{EM}}$) build supervision by filtering self-generated reasoning for correctness~\citep{zelikman2022star,singh2023restem}, and micro-reflective self-distillation adds capability-dependent trajectory selection~\citep{DBLP:journals/corr/abs-2606-18844}. MediSkill-Evo~\citep{wu2026mediskillevo} instead evolves validated clinical knowledge banks without updating the backbone. Work on frozen medical models similarly evolves external procedures, supported facts, and visual references, checking candidate updates on both new and previously seen cases while leaving model parameters unchanged~\citep{he2026frozenmodelsevolvingexpertise}. MedEvo follows STaR and ReST$^{\mathrm{EM}}$ self-training but differs in two respects: rather than STaR's rationalization, which supplies the correct answer as a hint~\citep{zelikman2022star}, it repairs failures with answer-free hints; and it selects each round's data using \textbf{medical-tag and error-type feedback on a fixed feedback set}, linking local recovery to what the student should learn next. Knowledge Navigation couples repair and feedback selection.

Appendix~\ref{app:broader-related-work} reviews medical hallucinations, reasoning supervision, correction, and self-evolution.

\section{Method}
\label{sec:method}\label{sec:overview}
\begin{figure*}[t]
\centering
\includegraphics[width=\textwidth]{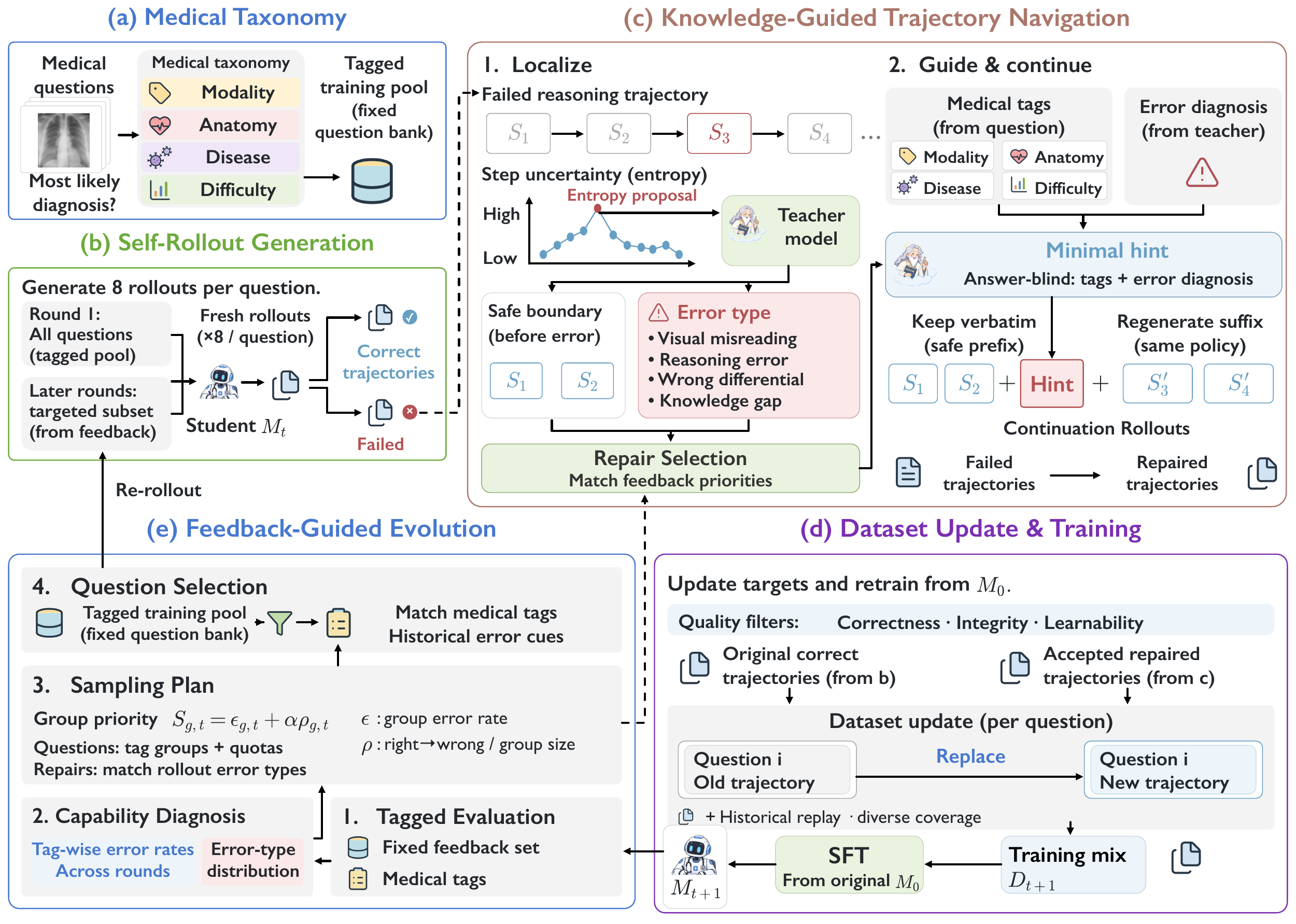}
\caption{\textbf{Knowledge Navigation in MedEvo.} (a) Medical tags organize questions. (b) The current student generates fresh rollouts. (c) Knowledge-guided trajectory navigation localizes errors and provides recovery hints through a gold-blind auditor; the student preserves the valid prefix and regenerates the suffix. (d) Admitted trajectories update the training set, retaining valid historical targets; the next model trains from $M_0$. (e) Fixed-set feedback guides question selection for the next round.}
\label{fig:method}
\end{figure*}

\paragraph{Knowledge Navigation.}\label{sec:knowledge-navigation}
We define \textbf{Knowledge Navigation} as a feedback-guided process that uses structured domain knowledge and failure diagnosis to choose \textbf{where to intervene, how to recover, and what to learn next}. MedEvo instantiates this process with a medical tag taxonomy and an error codebook: tags organize the clinical context, error types identify the failed reasoning operation, and their associated recovery operations guide intervention. Within a trajectory, the process selects a safe rollback boundary and an answer-free hint while preserving supported reasoning (\S\,\ref{sec:main-repair}). Across rounds, measured capability gaps guide question selection and repair priorities (\S\,\ref{sec:main-evolution}). The term navigation therefore describes the link from a structured diagnosis to an actionable decision and its subsequent reassessment. The taxonomy supplies the reference structure; student failures guide learning (Fig.~\ref{fig:method}).

\paragraph{Learning protocol.} A question $x_i=(I_i,q_i,O_i)$ consists of one or more medical images $I_i$, the clinical context and question $q_i$, and the answer options $O_i$. Following STaR and ReST$^{\mathrm{EM}}$~\citep{zelikman2022star,singh2023restem}, each round $t$ uses the current student $M_t=\pi_{\theta_t}$ to generate trajectories, collects correct rollouts and accepted repairs into the training set $D_{t+1}$, and fine-tunes the next student from the original model $M_0$ rather than from $M_t$. The training pool, feedback set, and test benchmarks are kept separate throughout.

\Needspace{4\baselineskip}
\subsection{Medical Taxonomy and Failure Diagnosis}
\label{sec:tags}
\subsubsection{Medical Question Tags}
We assign each question a four-dimensional hierarchical tag
\begin{equation}
z_i=(z_i^{\mathrm{mod}},z_i^{\mathrm{anat}},z_i^{\mathrm{dis}},z_i^{\mathrm{diff}}),
\label{eq:main-medical-tags}
\end{equation}
covering modality, anatomy, disease, and difficulty. Tags serve two roles. During repair, the auditor receives only the \textbf{observable subset} $z_i^{\mathrm{obs}}$, i.e., tags supported by the input itself, excluding answer-derived labels such as the diagnosed disease. Across rounds, tag-level accuracy on the feedback set identifies capability gaps and guides question selection. See Appendix~\ref{app:taxonomy} for the tag hierarchy and annotation rules.

\Needspace{4\baselineskip}
\subsubsection{Error Types and Their Relation to Repair}
\label{sec:main-error-types}
We diagnose failures at the process level, building on work that separates medical reasoning from knowledge and audits intermediate steps~\citep{DBLP:journals/corr/abs-2505-11462,DBLP:journals/corr/abs-2605-28301}. Our codebook defines \textbf{four parent error types} with five subtypes each, twenty in total (Appendix~\ref{app:error-codebook}). Each type is paired with a recovery operation used for hint construction (\S\,\ref{sec:main-auditing}): visual misreading calls for perceptual re-examination; reasoning error for reconstructing the evidence-to-decision link; differential diagnosis error for comparing discriminative features; and knowledge gap for a narrowly scoped knowledge cue. Whereas tags describe the question, error types describe what went wrong in a specific trajectory; because diagnoses vary across rollouts and checkpoints, \textbf{each fresh failure is diagnosed independently}.

\subsection{Knowledge-Guided Trajectory Navigation}
\phantomsection\label{sec:main-repair}
\subsubsection{Student Rollouts and Uncertainty-Based Localization}
\phantomsection\label{sec:main-localization}
Tyen et al.~\citep{tyen2023errors} show that large language models can correct reasoning errors once the error location is given; we therefore \textbf{localize an error before regenerating}. For each selected question, $M_t$ samples eight trajectories, and answer-correct ones proceed to quality screening (\S\,\ref{sec:main-admission}). An incorrect or unparsable trajectory is segmented into semantically complete steps
\begin{equation}
r^-=(s_1,\dots,s_K),
\label{eq:main-step-segmentation}
\end{equation}
such as image observation, feature interpretation, mechanism, differential diagnosis, and option mapping. The goal is to keep the steps before the \textbf{first consequential error} and regenerate the rest.

To propose where to audit, we compute the student's token entropy
\begin{equation}
H_j=-\sum_{v\in\mathcal V}\pi_{\theta_t}(v\mid x,r^-_{<j})\log\pi_{\theta_t}(v\mid x,r^-_{<j}),
\label{eq:main-token-entropy}
\end{equation}
where $\mathcal V$ is the vocabulary, $j$ indexes tokens, and $r^-_{<j}$ is the token prefix. Logs are natural, with $0\log0=0$. For each nonempty step $s_k$, let $\mathcal J_k$ be its token-index set and $H_{s_k}=(H_j)_{j\in\mathcal J_k}$. We combine their \textbf{mean and 90th percentile} $Q_{0.9}$, and measure the increase over the preceding $m_k=\min(m,k-1)$ steps:
\begin{align}
E^*_k&=\lambda\,\frac{1}{|\mathcal J_k|}\sum_{j\in\mathcal J_k}H_j+(1-\lambda)\,Q_{0.9}(H_{s_k}),\\
\Delta E_k&=\begin{cases}0,&k=1,\\ E^*_k-\dfrac{1}{m_k}\displaystyle\sum_{\ell=k-m_k}^{k-1}E^*_\ell,&2\le k\le K,\end{cases}
\label{eq:main-step-entropy}
\end{align}
For $k\ge2$, $m_k>0$. After min--max normalizing both profiles (constant profiles map to zero), we select the anchor
\begin{equation}
k^*\in\operatorname*{arg\,max}_{1\le k\le K}\big[\widetilde E^*_k+\widetilde{\Delta E}_k\big],
\end{equation}
and the audit window covers steps $k^*\pm1$, within valid indices, using $\lambda=0.5$ and $m=3$ (Appendix~\ref{app:localization-details}). \textbf{Entropy proposes the audit region}; the auditor locates errors and boundaries using images, questions, and reasoning (\S\,\ref{sec:main-auditing}).

\subsubsection{Gold-Blind Auditing and Medical Error Diagnosis}
\label{sec:main-auditing}
Building on hint-guided recovery in HHD~\citep{DBLP:journals/corr/abs-2605-11556}, an external auditor $\mathcal A$ (GPT-5.6-sol) turns the diagnosis into concise guidance. The auditor is gold-blind: it receives the question and images, the failed trajectory, the entropy proposal $\mathcal E_t(x,r^-)$, and the observable tags $z^{\mathrm{obs}}$, but \textbf{no reference answer, reference rationale, or answer-derived metadata}. It returns
\begin{equation}
(a,e,b,h)=\mathcal A\big(x,r^-,\mathcal E_t(x,r^-),z^{\mathrm{obs}}\big),
\label{eq:main-auditor}
\end{equation}
where $a\in\{\textsc{Repair},\textsc{Skip}\}$ is the action. For \textsc{Repair}, $e$ is the \textbf{earliest consequential error} together with its diagnosed type, $b$ is the last fully valid step before $e$, and $h$ is a \textbf{minimal recovery hint}. For \textsc{Skip}, $(e,b,h)$ are unused and no continuation is generated. The hint instantiates the recovery operation of the diagnosed type (\S\,\ref{sec:main-error-types}) in the observable context. For example, a perceptual hint may ask the student to reinspect a region or waveform attribute without stating the finding, and a differential hint may request comparison of a discriminative feature without naming the diagnosis. Hints must not reveal the answer, an answer option, or a corrected rationale. The audit prompt appears in Appendix~\ref{app:stage1}.

\subsubsection{Prefix Preservation and Same-Student Continuation}
Like LEMA~\citep{an2023lema}, we train on corrected mistakes, but the correction is generated by the student itself. The boundary splits the failed trajectory into a \textbf{preserved prefix and an erroneous suffix},
\begin{equation}
r^-=p_b\oplus s^-,\qquad p_b=(s_1,\dots,s_b).
\label{eq:main-prefix-split}
\end{equation}
Here $b\in\{0,\ldots,K-1\}$, $p_0$ is empty, and $\oplus$ denotes token-sequence concatenation. \textbf{We keep $p_b$ verbatim}, append the hint rendered as a natural-language recovery cue $g(h)$, and let the same checkpoint continue:
\begin{equation}
s^+\sim\pi_{\theta_t}\big(\cdot\mid x,p_b\oplus g(h)\big),\qquad r^*=p_b\oplus g(h)\oplus s^+.
\label{eq:main-continuation}
\end{equation}
The question stays in the user message, $p_b\oplus g(h)$ prefills an unfinished assistant message, and the structured audit is stored only as metadata. With the prefix fixed, we sample up to five suffixes per failed trajectory, \textbf{stopping at the first admitted continuation}, and send answer-correct continuations to admission (\S\,\ref{sec:main-admission}; Appendix~\ref{app:stage2}).

\subsubsection{Repair Admission and Training Targets}
\label{sec:main-admission}\label{sec:formulation:learnability}
A continuation is admitted only if it passes \textbf{answer, boundary, integrity, and learnability checks}. Learnability is scored on the \textbf{new suffix alone}, so that a valid prefix cannot mask the difficulty of the correction. Adapting the instruction-following difficulty of Li et al.~\citep{li2024quantity}, we measure how much the suffix depends on the task input:
\begin{equation}
D_{\mathrm{repair}}=\frac{\mathcal L_t(s^+\mid x,p_b)}{\mathcal L_t(s^+\mid p_b)},
\label{eq:main-suffix-difficulty}
\end{equation}
For a token sequence $s$ and context $c$, $\mathcal L_t(s\mid c)=-\sum_{j=1}^{|s|}\log\pi_{\theta_t}(s_j\mid c,s_{<j})$, in nats. Both terms are \textbf{computed without the hint}; the denominator omits $x$ but retains $p_b$. A smaller ratio indicates a larger relative loss reduction from explicitly conditioning on $x$ beyond the prefix, which may itself contain task information. Both suffixes must be nonempty, all losses finite, and the denominator strictly positive. We also measure how strongly the student still prefers its original wrong suffix, using per-token losses to account for length differences:
\begin{equation}
\Delta_{\mathrm{repair}}=\frac{\mathcal L_t(s^+\mid x,p_b)}{|s^+|}-\frac{\mathcal L_t(s^-\mid x,p_b)}{|s^-|}.
\label{eq:main-preference-gap}
\end{equation}
For a repair candidate $\xi=(x,r^-,r^*,b)$, the gap is measured in nats per token. A positive value means the student assigns lower average loss to the wrong suffix than to the repair. The learnability gate $L_\xi$ requires $D_{\mathrm{repair}}\in[0.35,0.85]$ and $\Delta_{\mathrm{repair}}\in[0.02,0.30]$, with \textbf{thresholds calibrated on the feedback set} and frozen before final evaluation (Appendix~\ref{app:training-details}). Admission is
\begin{equation}
\operatorname{Accept}(\xi)=C_\xi\wedge A_\xi\wedge N_\xi\wedge L_\xi,
\label{eq:main-admission-gate}
\end{equation}
Here $C_\xi$ checks correctness, $A_\xi$ validates the \textbf{prefix and boundary}, and $N_\xi$ rejects unsupported additions or repetition.

Each admitted repair is trained as the complete response $r^*=p_b\oplus g(h)\oplus s^+$, including the recovery cue, with \textbf{loss on all assistant tokens}:
\begin{equation}
\mathcal L_{\mathrm{repair}}(\theta)=-\sum_{j=1}^{|r^*|}\log\pi_\theta(r^*_j\mid x,r^*_{<j}).
\label{eq:main-repair-loss}
\end{equation}
Correct rollouts share this objective; image and user tokens are excluded from the loss. If no repair is admitted, we retain a qualified correct rollout or a valid historical target; if neither is available, we add none. \textbf{Evaluation excludes hints and audit outputs.}

\subsection{Feedback-Driven Evolution of Supervision}
\label{sec:main-evolution}
Across-round Knowledge Navigation follows the iterative self-training pattern of STaR and ReST$^{\mathrm{EM}}$~\citep{zelikman2022star,singh2023restem}, but uses \textbf{medical-tag and error-type feedback on a fixed set} to allocate generation and repair effort. The cycle is assessment, targeted selection, local repair, screened dataset update, and retraining. The taxonomy and error codebook remain fixed; \textbf{selection priorities and the supervision dataset evolve} as the student changes.

\subsubsection{First-Round Coverage}
The training pool $\mathcal T$ and a separate, fixed feedback set $V_{\mathrm{feedback}}$ are drawn from public medical datasets (\S\,\ref{sec:exp:setup}). In the first round, $M_0$ generates \textbf{rollouts for every question} in $\mathcal T$; correct trajectories enter screening, failures are repaired as in \S\,\ref{sec:main-repair}, and accepted targets form $D_1$, giving $M_1=\operatorname{SFT}(M_0,D_1)$.

\subsubsection{Feedback across Medical Tags and Error Types}
Before constructing $D_{t+1}$, we evaluate $M_t$ on the feedback set, including $M_0$ at initialization. For a supported tag group $g$ with $n_g>0$, the error rate is
\begin{equation}
\epsilon_{g,t}=\frac{1}{n_g}\sum_{i\in V_{\mathrm{feedback},g}}(1-c_{i,t}),\qquad n_g=|V_{\mathrm{feedback},g}|,
\label{eq:main-feedback}
\end{equation}
where $c_{i,t}\in\{0,1\}$ indicates correctness. Comparing consecutive checkpoints separates corrections, regressions, persistent errors, and retained successes (Appendix~\ref{app:feedback-selection-details}), and diagnosing failed responses yields each group's error-type distribution. Groups below a support threshold back off to their parent tag. Feedback questions guide selection and threshold calibration but \textbf{never enter training}, and test results never influence data construction.

\subsubsection{Selecting Questions and Updating the Dataset}
\label{sec:dataset-update}
From the second round on, we allocate the generation budget across groups by the rule-based priority
\begin{equation}
S_{g,t}=\epsilon_{g,t}+\alpha\,\rho_{g,t},\qquad \rho_{g,t}=\frac{N^{10}_{g,t}}{n_g},
\label{eq:main-priority}
\end{equation}
where $t\ge1$ and $N^{10}_{g,t}$ counts correct-to-incorrect transitions from $M_{t-1}$ to $M_t$ and $\alpha\geq0$ weights regressions. Quotas are proportional to $S_{g,t}$, capped by the available unique questions, and rounded deterministically; unused capacity is redistributed, and each multi-tag question is counted once (Appendix~\ref{app:planning-details}). Within each group, candidates are retrieved by tag match and ranked by historical error records. Selected questions receive \textbf{fresh rollouts} from $M_t$, their failures are diagnosed anew, and failures whose error types match the feedback priorities are repaired first. Feedback thus decides where effort is spent, while the checks in \S\,\ref{sec:main-admission} decide what enters training.

The dataset keeps \textbf{at most one active target per question}. An admitted candidate replaces the existing target only if a blinded comparison judges it better in evidence support, reasoning consistency, and conciseness; otherwise the existing target is kept, and all versions are logged separately. Writing each target as a pair $(x,r)$ keyed by question ID, with $D_0=\varnothing$, replaced targets $R_t\subseteq D_t$, accepted replacements $U_{t+1}$, and new targets $A_{t+1}$ for previously unrepresented questions,
\begin{equation}
D_{t+1}=(D_t\setminus R_t)\cup U_{t+1}\cup A_{t+1},
\label{eq:dataset-update}
\end{equation}
\begin{equation}
M_{t+1}=\operatorname{SFT}(M_0,D_{t+1}).
\label{eq:outer-update}
\end{equation}
$U_{t+1}$ contains exactly one replacement for each question in $R_t$; $A_{t+1}$ contains one target per new question, disjoint by question ID from $D_t$. Each round restarts from $M_0$ with a fresh optimizer and learning-rate schedule, so \textbf{experience is carried only through the evolving dataset}. Algorithm~\ref{alg:medevo} summarizes the procedure.

\begin{algorithm}[H]
\fontsize{9.5}{12}\selectfont
\DontPrintSemicolon
\SetAlgoNoLine
\SetAlgoVlined
\SetArgSty{textnormal}
\SetAlCapNameFnt{\normalfont}
\SetAlCapFnt{\bfseries}
\SetAlgoCaptionSeparator{:}
\SetKwInput{Require}{Require}
\SetKw{Return}{return}
\SetKw{Break}{break}\SetKw{Continue}{continue}
\SetCommentSty{MedEvoAlgorithmComment}
\caption{MedEvo Training with Local Repair and Feedback-Guided Data Evolution}
\label{alg:medevo}
\Require{Initial student $M_0$; training pool $\mathcal T$ with verification labels; fixed feedback set $V_{\mathrm{feedback}}$; gold-blind auditor $\mathcal A$; rounds $R$; selection and generation budgets.}
$\mathcal G\leftarrow\operatorname{AssignMedicalTags}(\mathcal T,V_{\mathrm{feedback}})$\;
$D_0\leftarrow\varnothing$; $H\leftarrow\varnothing$ \tcp*[r]{Active targets and trajectory history}
\For{$t\leftarrow 0$ \KwTo $R-1$}{
  \tcp{Feedback assessment and question selection}
  $F_t\leftarrow\operatorname{Evaluate}(M_t,V_{\mathrm{feedback}},\mathcal G)$\;
  \eIf{$t=0$}{
    $Q_t\leftarrow\mathcal T$ \tcp*[r]{Broad first-round coverage}
  }{
    $P_t\leftarrow\operatorname{PlanQuotas}(F_t,F_{t-1},\mathcal G)$\;
    $Q_t\leftarrow\operatorname{RetrieveUnique}(\mathcal T,P_t,H)$\;
  }
  $D_{t+1}\leftarrow D_t$\;
  \tcp{Fresh student rollouts and local trajectory repair}
  \ForEach{$x\in Q_t$}{
    $\mathcal R_x\leftarrow\operatorname{Rollout}(M_t,x,8)$\;
    $C_x\leftarrow\operatorname{ScreenCorrect}(x,\mathcal R_x)$\;
    \ForEach{failed $r^-\in\operatorname{PrioritizeFailures}(\mathcal R_x,F_t)$ within budget}{
      $u\leftarrow\operatorname{StepUncertainty}(M_t,x,r^-)$\;
      $(a,e,b,h)\leftarrow\mathcal A(x,r^-,u,z^{\mathrm{obs}}(x))$ \tcp*[r]{Observable tags only}
      $H\leftarrow\operatorname{RecordHistory}(H,x,r^-,a,e,b,h)$\;
      \If{$a=\textsc{Skip}$}{\Continue\;}
      $p_b\leftarrow\operatorname{ValidPrefix}(r^-,b)$\;
      \For{$j\leftarrow 1$ \KwTo $5$}{
        $s^+\sim M_t(\,\cdot\mid x,p_b\oplus g(h))$\;
        $r^*\leftarrow p_b\oplus g(h)\oplus s^+$\;
        $\xi\leftarrow(x,r^-,r^*,b)$\;
        \If{$\operatorname{Accept}(\xi)$ \normalfont{(Eq.~\ref{eq:main-admission-gate})}}{
          $C_x\leftarrow C_x\cup\{r^*\}$\;
          \Break\;
        }
      }
    }
    \tcp{Keep one active target per question; retain valid history on ties}
    $D_{t+1}\leftarrow\operatorname{RetainOrReplace}(D_{t+1},x,C_x)$\;
  }
  \tcp{Full-response SFT from the same initial student}
  $M_{t+1}\leftarrow\operatorname{SFT}(M_0,D_{t+1})$ \tcp*[r]{Reset optimizer and schedule}
}
\Return{$\{M_t,D_t\}_{t=1}^{R}$}\;
\end{algorithm}

\begingroup
\renewcommand{\targetmodel}{Qwen3.5-9B}
\renewcommand{\targetparams}{9B}
\Needspace{8\baselineskip}
\section{Experiments}
\label{sec:exp}
\subsection{Experimental Setup}
\label{sec:exp:setup}

\paragraph{Models and data.} Our main backbone is Qwen3.5-9B~\citep{qwen3.5}; Qwen3.5-35B and Qwen3.5-397B test scalability (\S\,\ref{sec:current:scaling}). The training pool contains 81,293 questions from the training splits of public medical datasets, spanning nine modality categories and five difficulty levels (Fig.~\ref{fig:dataset_overview}), and is deduplicated by question and source identifiers, patient/case identifiers when available, image hashes, and near-duplicate clusters (Appendix~\ref{app:contracts}). A separate, fixed set of 5,000 modality-stratified questions serves as the feedback set and \textbf{never enters training}. Dataset roles and training configurations are given in Appendices~\ref{app:dataset-statistics} and~\ref{app:manifest}.

\newsavebox{\repairstatstable}
\sbox{\repairstatstable}{%
\fontsize{9.5}{11.5}\selectfont
\setlength{\tabcolsep}{2pt}
\renewcommand{\arraystretch}{1.18}
\begin{tabular}{@{}lrrr@{}}
\toprule
\multicolumn{1}{@{}c}{\textbf{Training round}} & \multicolumn{1}{c}{\textbf{R1}} & \multicolumn{1}{c}{\textbf{R2}} & \multicolumn{1}{c@{}}{\textbf{R3}} \\
\multicolumn{1}{@{}c}{Generator} & \multicolumn{1}{c}{R0} & \multicolumn{1}{c}{R1} & \multicolumn{1}{c@{}}{R2} \\
\midrule
\rowcolor[HTML]{E4F1FB}
\textbf{Sent for repair} & \textbf{27,233}\makebox[0.55em][l]{} & \textbf{25,831}\makebox[0.55em][l]{\textcolor[HTML]{E34B43}{\tablechange{$\downarrow$}}} & \textbf{23,620}\makebox[0.55em][l]{\textcolor[HTML]{E34B43}{\tablechange{$\downarrow$}}} \\
Share of question bank & 33.50\%\makebox[0.55em][l]{} & 31.78\%\makebox[0.55em][l]{} & 29.06\%\makebox[0.55em][l]{} \\
Successfully repaired & 19,567\makebox[0.55em][l]{} & 18,258\makebox[0.55em][l]{} & 16,797\makebox[0.55em][l]{} \\
\textbf{Repair success rate} & \textbf{71.85\%}\makebox[0.55em][l]{} & \textbf{70.68\%}\makebox[0.55em][l]{\textcolor[HTML]{E34B43}{\tablechange{$\downarrow$}}} & \textbf{71.11\%}\makebox[0.55em][l]{\textcolor[HTML]{287DB1}{\tablechange{$\uparrow$}}} \\
Mean repair attempts & 3.37\makebox[0.55em][l]{} & 2.48\makebox[0.55em][l]{} & 2.12\makebox[0.55em][l]{} \\
\bottomrule
\end{tabular}}
\begin{table}[!htbp]
\centering
\begin{minipage}{\wd\repairstatstable}
\centering
\captionsetup{width=\linewidth}
\caption{\textbf{Repair workload across rounds.} Rate: fraction repaired; arrows compare rounds.}
\label{tab:current_counts}
\usebox{\repairstatstable}
\end{minipage}
\end{table}

\finishfigurewrap
{\emergencystretch=1em
\paragraph{Evaluation.} We use MedUniEval~\citep{jiang2025hulumed}, the Hulu-Med implementation built on Lingshu's MedEvalKit~\citep{xu2025lingshu}, for medical visual question answering (VQA) on PMC-VQA~\citep{zhang2023pmcvqa}, VQA-RAD~\citep{lau2018vqarad}, SLAKE~\citep{liu2021slake}, PathVQA~\citep{he2020pathvqa}, OmniMedVQA~\citep{hu2024omnimedvqa}, the multimodal track of MedXpertQA (MedXpertQA-MM)~\citep{DBLP:conf/icml/ZuoQLCZHZ0025}, and the medical subset of Massive Multi-discipline Multimodal Understanding (MMMU), denoted MMMU-Med~\citep{yue2023mmmu}. All models share the \textbf{answering prompt and parser} (Appendix~\ref{app:shared_qa_prompt}). We report accuracy (\%) and the unweighted seven-benchmark mean; accuracy gains and $\Delta$ denote absolute differences in percentage points.\par}

\finishfigurewrap
\newlength{\mainresultstablewidth}
\begin{table}[!htbp]
\centering\fontsize{9.5}{11.5}\selectfont
\caption{\textbf{Medical reasoning benchmarks.} General-purpose and medical model accuracy (\%). MedEvo: R3. Avg.: unweighted seven-benchmark mean; MedXQA: MedXpertQA-MM; MMMU-Med: the medical subset of MMMU. \medicalmark: medical models; \repairmark: trajectory-repair methods. Blue: MedEvo. Bold/underline: top two per group and benchmark, with shared ranks for ties.}
\label{tab:current_main}\label{tab:main}
\setlength{\tabcolsep}{2pt}
\renewcommand{\arraystretch}{1.0}
\renewcommand{\modelmarks}[2]{\makebox[1.05em][c]{#1}\makebox[1.05em][c]{#2}\hspace{0.2em}\hfill}
\sbox{2}{Baseline 2 (TrajFusion)~{\small\citep{DBLP:conf/acl/DengTLLWLX26}}}
\sbox{0}{%
\begin{tabular}{>{\raggedleft\arraybackslash}p{\dimexpr\wd2+2.3em\relax}rrrrrrrr}
\toprule
\multicolumn{1}{c}{\textbf{Model}} & \multicolumn{1}{c}{\makecell{PMC-\\VQA}} & \multicolumn{1}{c}{\makecell{VQA-\\RAD}} & \multicolumn{1}{c}{SLAKE} & \multicolumn{1}{c}{PathVQA} & \multicolumn{1}{c}{\makecell{OmniMed\\VQA}} & \multicolumn{1}{c}{\textbf{MedXQA}} & \multicolumn{1}{c}{\makecell{MMMU-\\Med}} & \multicolumn{1}{c}{\textbf{Avg.}} \\
\midrule
\rowcolor[HTML]{ECECEC}
\multicolumn{9}{c}{\textit{Proprietary models}} \\
\midrule
\modelmarks{}{}GPT-5.6-sol & \underline{60.43} & \underline{70.29} & \textbf{81.34} & \underline{54.84} & \underline{76.67} & \underline{77.50} & \underline{86.24} & \underline{72.47} \\
\modelmarks{}{}Gemini-3.8-Flash & \textbf{64.52} & \textbf{74.72} & \textbf{81.34} & \textbf{58.58} & \textbf{80.76} & \textbf{86.30} & \textbf{89.10} & \textbf{76.47} \\
\midrule
\rowcolor[HTML]{ECECEC}
\multicolumn{9}{c}{\textit{Open-source models ($<10$B)}} \\
\midrule
\modelmarks{\medicalmark}{}LLaVA-Med-7B~{\small\citep{li2023llavamed}} & 30.50 & 53.70 & 48.00 & 38.80 & 44.30 & 20.30 & 29.30 & 37.84 \\
\modelmarks{\medicalmark}{}HuatuoGPT-V-7B~{\small\citep{DBLP:journals/corr/abs-2406-19280}} & 53.30 & 67.00 & 67.80 & 48.00 & 74.20 & 21.60 & 47.30 & 54.17 \\
\modelmarks{\medicalmark}{}BiMediX2-8B~{\small\citep{mullappilly2024bimedix2biomedicalexpertlmm}} & 43.50 & 49.20 & 57.70 & 37.00 & 63.30 & 21.80 & 39.80 & 44.61 \\
\modelmarks{}{}InternVL2.5-8B~{\small\citep{DBLP:journals/corr/abs-2412-05271}} & 51.30 & 59.40 & 69.00 & 42.10 & 81.30 & 21.70 & 53.50 & 54.04 \\
\modelmarks{}{}InternVL3-8B~{\small\citep{DBLP:journals/corr/abs-2504-10479}} & 53.80 & 65.40 & 72.80 & 48.60 & 79.10 & 22.40 & 59.20 & 57.33 \\
\modelmarks{\medicalmark}{}LINGSHU-7B~{\small\citep{xu2025lingshu}} & 56.30 & 67.90 & \textbf{83.10} & \textbf{61.90} & 82.90 & 26.70 & 54.00 & 61.83 \\
\modelmarks{}{}Qwen3.5-9B (Base)~{\small\citep{qwen3.5}} & 59.20 & 68.10 & 76.80 & 48.80 & 82.20 & 47.40 & 76.50 & 65.57 \\
\rowcolor[HTML]{F0F0F0}
\modelmarks{}{\repairmark}Baseline 1 (HHD)~{\small\citep{DBLP:journals/corr/abs-2605-11556}} & 58.93 & 67.65 & 71.45 & 50.12 & 82.35 & 48.75 & 76.90 & 65.16 \\
\rowcolor[HTML]{F0F0F0}
\modelmarks{}{\repairmark}Baseline 2 (TrajFusion)~{\small\citep{DBLP:conf/acl/DengTLLWLX26}} & 59.32 & 69.12 & 77.23 & 49.23 & \underline{83.46} & 49.56 & 77.20 & 66.45 \\
\rowcolor[HTML]{F0F0F0}
\modelmarks{}{}Baseline 3 (Teacher SFT) & \underline{61.23} & \underline{70.15} & 71.60 & 53.20 & 81.90 & \underline{52.30} & \underline{78.14} & \underline{66.93} \\
\rowcolor[HTML]{E4F1FB}
\modelmarks{\medicalmark}{\repairmark}\textbf{MedEvo (9B)} & \textbf{62.68} & \textbf{70.40} & \underline{78.23} & \underline{53.34} & \textbf{85.41} & \textbf{54.50} & \textbf{79.28} & \textbf{69.12} \\
\midrule
\rowcolor[HTML]{ECECEC}
\multicolumn{9}{c}{\textit{Open-source models ($>10$B)}} \\
\midrule
\modelmarks{\medicalmark}{}HealthGPT-14B~{\small\citep{DBLP:conf/icml/LinZLYY0H00ST0L25}} & 56.40 & 65.00 & 66.10 & \underline{56.70} & 75.20 & 24.70 & 49.60 & 56.24 \\
\modelmarks{\medicalmark}{}HuatuoGPT-V-34B~{\small\citep{DBLP:journals/corr/abs-2406-19280}} & 56.60 & 61.40 & 69.50 & 44.40 & 74.00 & 22.10 & 51.80 & 54.26 \\
\modelmarks{}{}InternVL3-14B~{\small\citep{DBLP:journals/corr/abs-2504-10479}} & 54.10 & 66.30 & 72.80 & 48.00 & 78.90 & 23.10 & 63.10 & 58.04 \\
\modelmarks{}{}InternVL2.5-38B~{\small\citep{DBLP:journals/corr/abs-2412-05271}} & 57.20 & 61.40 & 70.30 & 46.90 & 79.90 & 24.40 & 61.60 & 57.39 \\
\modelmarks{}{}InternVL3-38B~{\small\citep{DBLP:journals/corr/abs-2504-10479}} & 56.60 & 65.40 & 72.70 & 51.00 & 79.80 & 25.20 & 65.20 & 59.41 \\
\modelmarks{\medicalmark}{}LINGSHU-32B~{\small\citep{xu2025lingshu}} & 57.90 & \textbf{76.50} & \textbf{89.20} & \textbf{65.90} & \underline{83.40} & 30.90 & 62.30 & 66.59 \\
\modelmarks{}{}Qwen3.5-35B (Base)~{\small\citep{qwen3.5}} & \underline{62.40} & 70.50 & 70.10 & 49.15 & 83.20 & \underline{60.00} & \underline{80.00} & \underline{67.91} \\
\rowcolor[HTML]{E4F1FB}
\modelmarks{\medicalmark}{\repairmark}\textbf{MedEvo (35B)} & \textbf{64.63} & \underline{75.83} & \underline{80.30} & 56.01 & \textbf{83.97} & \textbf{66.50} & \textbf{84.02} & \textbf{73.04} \\
\bottomrule
\end{tabular}}
\global\mainresultstablewidth=\wd0
\usebox{0}
\end{table}

\paragraph{Baselines.} On Qwen3.5-9B, we compare with two methods that learn from failed trajectories, HHD~\citep{DBLP:journals/corr/abs-2605-11556} and TrajFusion~\citep{DBLP:conf/acl/DengTLLWLX26}, and with \textbf{teacher-distillation SFT} on rationales generated by Qwen-3.8-Max. External general-purpose and medical models provide additional context (Appendix~\ref{app:baseline-details}).

\subsection{Main Results}
\label{sec:exp:main}
Tab.~\ref{tab:current_main} reports the results. On Qwen3.5-9B, MedEvo raises mean accuracy from 65.57\% to 69.12\% and \textbf{improves all seven benchmarks}, with the largest gain on MedXpertQA-MM~\citep{DBLP:conf/icml/ZuoQLCZHZ0025} (47.40\% $\rightarrow$ 54.50\%). It outperforms HHD~\citep{DBLP:journals/corr/abs-2605-11556}, TrajFusion~\citep{DBLP:conf/acl/DengTLLWLX26}, and teacher-distillation SFT, whose mean accuracies are 65.16\%, 66.45\%, and 66.93\%, respectively. On Qwen3.5-35B, mean accuracy rises from 67.91\% to 73.04\% (\S\,\ref{sec:current:scaling}). A question-level paired bootstrap (10,000 resamples; Tab.~\ref{tab:paired-bootstrap}, right) yields \textbf{95\% confidence intervals (CIs) that exclude zero} on every benchmark. These intervals capture question-level uncertainty; variation across training runs is reported in Tab.~\ref{tab:seed-stability}.
\begin{table}[!htbp]
\centering\small
\setlength{\abovecaptionskip}{2pt}
\setlength{\belowcaptionskip}{2pt}
\caption{\textbf{Evolution and statistical uncertainty (Qwen3.5-9B).} Left: accuracy (\%) across rounds, each trained from Base (R0). Right: absolute gain R3$-$R0 and 95\% bootstrap confidence intervals (CIs) (10,000 paired question resamples). The mean CI resamples each benchmark independently before averaging. All gain CIs exclude zero. Arrows compare consecutive rounds; MedXQA annotations give absolute changes. Bold/underline: best/second-best per row.}
\label{tab:paired-bootstrap}
\label{tab:current_rounds}
\setlength{\tabcolsep}{3pt}
\renewcommand{\arraystretch}{1.0}
\begin{tabularx}{\mainresultstablewidth}{l*{4}{>{\raggedright\arraybackslash}X}!{\vrule width 0.4pt}rc}
\toprule
 & \multicolumn{4}{c!{\vrule width 0.4pt}}{\textbf{Round-wise accuracy}} & \multicolumn{2}{c}{\textbf{Paired bootstrap}} \\
\cmidrule(lr){2-5}\cmidrule(lr){6-7}
Benchmark & R0 & R1 & R2 & R3 & $\Delta$ (R3--R0) & 95\% CI \\
\midrule
PMC-VQA & 59.20 & 59.54\,\textcolor[HTML]{287DB1}{\tablechange{$\uparrow$}} & \underline{60.97}\,\textcolor[HTML]{287DB1}{\tablechange{$\uparrow$}} & \textbf{62.68}\,\textcolor[HTML]{287DB1}{\tablechange{$\uparrow$}} & \tablechange{+3.48} & [+1.9, +5.1] \\
VQA-RAD & 68.10 & \underline{68.47}\,\textcolor[HTML]{287DB1}{\tablechange{$\uparrow$}} & 68.06\,\textcolor[HTML]{E34B43}{\tablechange{$\downarrow$}} & \textbf{70.40}\,\textcolor[HTML]{287DB1}{\tablechange{$\uparrow$}} & \tablechange{+2.30} & [+0.8, +3.8] \\
SLAKE & 76.80 & 76.92\,\textcolor[HTML]{287DB1}{\tablechange{$\uparrow$}} & \underline{77.34}\,\textcolor[HTML]{287DB1}{\tablechange{$\uparrow$}} & \textbf{78.23}\,\textcolor[HTML]{287DB1}{\tablechange{$\uparrow$}} & \tablechange{+1.43} & [+0.2, +2.7] \\
PathVQA & 48.80 & \underline{52.38}\,\textcolor[HTML]{287DB1}{\tablechange{$\uparrow$}} & 51.27\,\textcolor[HTML]{E34B43}{\tablechange{$\downarrow$}} & \textbf{53.34}\,\textcolor[HTML]{287DB1}{\tablechange{$\uparrow$}} & \tablechange{+4.54} & [+2.6, +6.5] \\
OmniMedVQA & \underline{82.20} & 80.81\,\textcolor[HTML]{E34B43}{\tablechange{$\downarrow$}} & 81.58\,\textcolor[HTML]{287DB1}{\tablechange{$\uparrow$}} & \textbf{85.41}\,\textcolor[HTML]{287DB1}{\tablechange{$\uparrow$}} & \tablechange{+3.21} & [+1.5, +4.9] \\
\rowcolor[HTML]{E4F1FB}
\textbf{MedXQA} & 47.40 & 52.56\,\textcolor[HTML]{287DB1}{\tablechange{$\uparrow5.16$}} & \underline{52.79}\,\textcolor[HTML]{287DB1}{\tablechange{$\uparrow0.23$}} & \textbf{54.50}\,\textcolor[HTML]{287DB1}{\tablechange{$\uparrow1.71$}} & \tablechange{+7.10} & [+5.2, +9.0] \\
MMMU-Med & 76.50 & \underline{79.83}\,\textcolor[HTML]{287DB1}{\tablechange{$\uparrow$}} & \textbf{79.91}\,\textcolor[HTML]{287DB1}{\tablechange{$\uparrow$}} & 79.28\,\textcolor[HTML]{E34B43}{\tablechange{$\downarrow$}} & \tablechange{+2.78} & [+1.3, +4.3] \\
\rowcolor[HTML]{E4F1FB}
\textbf{Avg.} & 65.57 & 67.22\,\textcolor[HTML]{287DB1}{\tablechange{$\uparrow$}} & \underline{67.42}\,\textcolor[HTML]{287DB1}{\tablechange{$\uparrow$}} & \textbf{69.12}\,\textcolor[HTML]{287DB1}{\tablechange{$\uparrow$}} & \tablechange{+3.55} & [+2.4, +4.7] \\
\bottomrule
\end{tabularx}
\end{table}

\Needspace{8\baselineskip}
\subsection{Accuracy Gains and Repair Workload}
\label{sec:current:rounds}
Mean accuracy increases from 65.57\% to 67.22\%, 67.42\%, and 69.12\% across rounds (Tab.~\ref{tab:current_rounds}, left), while the number of questions sent for repair falls from 27,233 to 23,620 and repair success remains near 71\% (Tab.~\ref{tab:current_counts}). Thus, the later rounds improve benchmark performance with a smaller question-level repair workload; \S\,\ref{sec:current:evolution} isolates the role of feedback-guided selection. Individual benchmarks still fluctuate, as illustrated by PathVQA~\citep{he2020pathvqa}. Modality and error-type statistics on the fixed feedback set guide subsequent selection (Appendix~\ref{app:exp-supplement}); the \textbf{seven held-out benchmarks assess generalization}.

\FloatBarrier
\Needspace{14\baselineskip}
\section{Ablation Studies and Analysis}
\label{sec:ablation}
We analyze repair, feedback, scale, and \textbf{learning value}, using Qwen3.5-9B except in scaling experiments~\citep{qwen3.5}.
\Needspace{8\baselineskip}
\subsection{Repair Design in the First Round}
\label{sec:current:navigation}
\begin{table}[!htbp]
\centering\fontsize{9.5}{11.5}\selectfont
\captionsetup{width=0.94\linewidth}
\setlength{\abovecaptionskip}{2pt}
\setlength{\belowcaptionskip}{2pt}
\caption{\textbf{Repair design (Qwen3.5-9B, R1).} Repair success and downstream accuracy (\%); Base accuracy is 65.57\%. Type-guided suffix continuation performs best on both. Red $\downarrow$: below the full method; bold: best.}
\label{tab:current_nav}
\setlength{\tabcolsep}{3pt}
\renewcommand{\arraystretch}{1.0}
\sbox{0}{%
\begin{tabularx}{0.94\linewidth}{>{\raggedright\arraybackslash}Xll}
\toprule
Method & \makecell{Repair\\success} & \makecell{Mean\\accuracy} \\
\midrule
Hint + full rewrite & 57.80\textcolor[HTML]{E34B43}{\tablechange{$\downarrow$}} & 63.20\textcolor[HTML]{E34B43}{\tablechange{$\downarrow$}} \\
No-hint rewrite & 36.70\textcolor[HTML]{E34B43}{\tablechange{$\downarrow$}} & 61.16\textcolor[HTML]{E34B43}{\tablechange{$\downarrow$}} \\
Generic hint + suffix continuation & 64.50\textcolor[HTML]{E34B43}{\tablechange{$\downarrow$}} & 64.20\textcolor[HTML]{E34B43}{\tablechange{$\downarrow$}} \\
\rowcolor[HTML]{E4F1FB}
\textbf{Type-guided hint + suffix continuation (Ours)} & \textbf{71.80} & \textbf{67.22} \\
\bottomrule
\end{tabularx}}
\usebox{0}
\end{table}
Tab.~\ref{tab:current_nav} compares four first-round designs varying regeneration scope (full rewrite vs.\ suffix continuation) and hint specificity (none, generic, or type-guided). We measure repair success and downstream mean accuracy. \textbf{Type-guided suffix continuation performs best} with 71.80\% repair success and 67.22\% mean accuracy. Generic-hint continuation achieves 64.50\% repair success and 64.20\% accuracy; hint-guided full rewriting achieves 57.80\% and 63.20\%. Both prefix preservation and error-type conditioning contribute.

\Needspace{12\baselineskip}
\subsection{Contribution of Feedback across Rounds}
\label{sec:current:evolution}
\begin{wraptable}{r}{0.48\textwidth}
\vspace{-\intextsep}
\centering\fontsize{9.5}{11.5}\selectfont
\captionsetup{skip=3pt}
\caption{\textbf{Feedback ablation (Qwen3.5-9B).} Mean accuracy (\%) for branches sharing the R1 generator and historical dataset (R1: 67.22\%); each round of SFT restarts from $M_0$. Full MedEvo achieves the highest R3 accuracy. Arrows compare with R1; $\Delta=\mathrm{R3}-\mathrm{R1}$.}
\label{tab:current_evo}
\setlength{\tabcolsep}{2.5pt}
\renewcommand{\arraystretch}{1.10}
\sbox{0}{%
\begin{tabularx}{\linewidth}{>{\raggedright\arraybackslash}Xrrr}
\toprule
Method & R2 & R3 & $\Delta$ \\
\midrule
No feedback (repair only) & 66.39\textcolor[HTML]{E34B43}{\tablechange{$\downarrow$}} & 66.67\textcolor[HTML]{E34B43}{\tablechange{$\downarrow$}} & \tablechange{$-0.55$} \\
Without medical tags & 67.32\textcolor[HTML]{287DB1}{\tablechange{$\uparrow$}} & 67.53\textcolor[HTML]{287DB1}{\tablechange{$\uparrow$}} & \tablechange{$+0.31$} \\
Without error-type feedback & 67.01\textcolor[HTML]{E34B43}{\tablechange{$\downarrow$}} & 66.21\textcolor[HTML]{E34B43}{\tablechange{$\downarrow$}} & \tablechange{$-1.01$} \\
\rowcolor[HTML]{E4F1FB}
\textbf{Full MedEvo} & \textbf{67.42}\textcolor[HTML]{287DB1}{\tablechange{$\uparrow$}} & \textbf{69.12}\textcolor[HTML]{287DB1}{\tablechange{$\uparrow$}} & \tablechange{$\mathbf{+1.90}$} \\
\bottomrule
\end{tabularx}}
\usebox{0}
\vspace{-0.5\baselineskip}
\end{wraptable}
Tab.~\ref{tab:current_evo} isolates the contribution of feedback. Sharing the R1 data-generation model and historical dataset (R1 accuracy: 67.22\%), four variants run two more rounds: repair without feedback, feedback without medical tags, feedback without error types, and full MedEvo. Full MedEvo reaches 69.12\% at R3 versus 66.67\%, 67.53\%, and 66.21\% for the three variants, respectively. Without feedback or without error-type feedback, accuracy falls below the shared R1 reference; without medical tags, it improves only marginally (67.22\% to 67.53\%). \textbf{Combined medical-tag and error-type feedback performs best}.
\finishfigurewrap

\subsection{Generalization across Model Scales}
\label{sec:current:scaling}
We apply MedEvo to Qwen3.5-35B for three rounds and to Qwen3.5-397B for two rounds, evaluating on the same seven benchmarks~\citep{DBLP:conf/icml/ZuoQLCZHZ0025,zhang2023pmcvqa,lau2018vqarad,liu2021slake,he2020pathvqa,hu2024omnimedvqa,yue2023mmmu} (Tab.~\ref{tab:current_35b}). On Qwen3.5-35B, mean accuracy rises from 67.91\% to 69.73\%, 70.81\%, and 73.04\%. MedXpertQA-MM~\citep{DBLP:conf/icml/ZuoQLCZHZ0025} peaks at 66.73\% in R2 and declines slightly to 66.50\% in R3, indicating benchmark-dependent optima. On Qwen3.5-397B, mean accuracy rises from 71.83\% to 74.71\% and 74.95\%, and MedXpertQA-MM from 67.65\% to 75.00\%. The R2 model improves six of seven benchmarks, with OmniMedVQA~\citep{hu2024omnimedvqa} essentially unchanged (85.10\% $\rightarrow$ 85.05\%). At both scales, \textbf{mean accuracy increases in every round}.
\begin{table}[!htbp]
\centering\small
\caption{\textbf{Scaling to larger models.} Accuracy (\%) for Qwen3.5-35B (left) and Qwen3.5-397B (right). Every round trains from Base (R0). Avg.: seven-benchmark mean. Arrows compare consecutive rounds; MedXQA annotations give absolute changes. Bold/underline: best/second-best within each model scale and benchmark.}
\label{tab:current_35b}
\label{tab:current_397b}
\setlength{\tabcolsep}{3pt}
\renewcommand{\arraystretch}{1.18}
\begin{tabularx}{\linewidth}{l*{4}{>{\raggedright\arraybackslash}X}!{\vrule width 0.4pt}*{3}{>{\raggedright\arraybackslash}X}}
\toprule
 & \multicolumn{4}{c!{\vrule width 0.4pt}}{\textbf{35B}} & \multicolumn{3}{c}{\textbf{397B}} \\
\cmidrule(lr){2-5}\cmidrule(lr){6-8}
Benchmark & R0 & R1 & R2 & R3 & R0 & R1 & R2 \\
\midrule
PMC-VQA & 62.40 & \underline{63.12}\,\textcolor[HTML]{287DB1}{\tablechange{$\uparrow$}} & 60.33\,\textcolor[HTML]{E34B43}{\tablechange{$\downarrow$}} & \textbf{64.63}\,\textcolor[HTML]{287DB1}{\tablechange{$\uparrow$}} & 63.47 & \underline{64.87}\,\textcolor[HTML]{287DB1}{\tablechange{$\uparrow$}} & \textbf{65.72}\,\textcolor[HTML]{287DB1}{\tablechange{$\uparrow$}} \\
VQA-RAD & 70.50 & 62.08\,\textcolor[HTML]{E34B43}{\tablechange{$\downarrow$}} & \underline{70.91}\,\textcolor[HTML]{287DB1}{\tablechange{$\uparrow$}} & \textbf{75.83}\,\textcolor[HTML]{287DB1}{\tablechange{$\uparrow$}} & 74.72 & \underline{76.05}\,\textcolor[HTML]{287DB1}{\tablechange{$\uparrow$}} & \textbf{77.15}\,\textcolor[HTML]{287DB1}{\tablechange{$\uparrow$}} \\
SLAKE & 70.10 & 78.14\,\textcolor[HTML]{287DB1}{\tablechange{$\uparrow$}} & \underline{79.18}\,\textcolor[HTML]{287DB1}{\tablechange{$\uparrow$}} & \textbf{80.30}\,\textcolor[HTML]{287DB1}{\tablechange{$\uparrow$}} & 77.38 & \underline{78.89}\,\textcolor[HTML]{287DB1}{\tablechange{$\uparrow$}} & \textbf{79.10}\,\textcolor[HTML]{287DB1}{\tablechange{$\uparrow$}} \\
PathVQA & 49.15 & \underline{52.71}\,\textcolor[HTML]{287DB1}{\tablechange{$\uparrow$}} & 50.87\,\textcolor[HTML]{E34B43}{\tablechange{$\downarrow$}} & \textbf{56.01}\,\textcolor[HTML]{287DB1}{\tablechange{$\uparrow$}} & 52.49 & \textbf{58.51}\,\textcolor[HTML]{287DB1}{\tablechange{$\uparrow$}} & \underline{57.90}\,\textcolor[HTML]{E34B43}{\tablechange{$\downarrow$}} \\
OmniMedVQA & 83.20 & \textbf{85.39}\,\textcolor[HTML]{287DB1}{\tablechange{$\uparrow$}} & \underline{85.22}\,\textcolor[HTML]{E34B43}{\tablechange{$\downarrow$}} & 83.97\,\textcolor[HTML]{E34B43}{\tablechange{$\downarrow$}} & \underline{85.10} & \textbf{85.15}\,\textcolor[HTML]{287DB1}{\tablechange{$\uparrow$}} & 85.05\,\textcolor[HTML]{E34B43}{\tablechange{$\downarrow$}} \\
\rowcolor[HTML]{E4F1FB}
\textbf{MedXQA} & 60.00 & 65.68\,\textcolor[HTML]{287DB1}{\tablechange{$\uparrow+5.68$}} & \textbf{66.73}\,\textcolor[HTML]{287DB1}{\tablechange{$\uparrow+1.05$}} & \underline{66.50}\,\textcolor[HTML]{E34B43}{\tablechange{$\downarrow-0.23$}} & 67.65 & \underline{74.40}\,\textcolor[HTML]{287DB1}{\tablechange{$\uparrow+6.75$}} & \textbf{75.00}\,\textcolor[HTML]{287DB1}{\tablechange{$\uparrow+0.60$}} \\
MMMU-Med & 80.00 & 81.02\,\textcolor[HTML]{287DB1}{\tablechange{$\uparrow$}} & \underline{82.41}\,\textcolor[HTML]{287DB1}{\tablechange{$\uparrow$}} & \textbf{84.02}\,\textcolor[HTML]{287DB1}{\tablechange{$\uparrow$}} & 82.03 & \textbf{85.10}\,\textcolor[HTML]{287DB1}{\tablechange{$\uparrow$}} & \underline{84.75}\,\textcolor[HTML]{E34B43}{\tablechange{$\downarrow$}} \\
\rowcolor[HTML]{E4F1FB}
\textbf{Avg.} & 67.91 & 69.73\,\textcolor[HTML]{287DB1}{\tablechange{$\uparrow$}} & \underline{70.81}\,\textcolor[HTML]{287DB1}{\tablechange{$\uparrow$}} & \textbf{73.04}\,\textcolor[HTML]{287DB1}{\tablechange{$\uparrow$}} & 71.83 & \underline{74.71}\,\textcolor[HTML]{287DB1}{\tablechange{$\uparrow$}} & \textbf{74.95}\,\textcolor[HTML]{287DB1}{\tablechange{$\uparrow$}} \\
\bottomrule
\end{tabularx}
\end{table}

\subsection{Learning Value of Repair Supervision}
\label{sec:current:extensions}
Beyond repair success, we examine three questions about \textbf{what makes repair supervision useful} (Fig.~\ref{fig:current_D}).

\paragraph{Which trajectories are worth repairing?} We compare the 8,000 questions with the longest failed trajectories against 8,000 random questions, each used in one round. The longest set has lower repair success (55.70\% vs.\ 67.50\%) but yields higher downstream accuracy (66.78\% vs.\ 64.35\%; Fig.~\ref{fig:current_D}a), favoring \textbf{learning-based trajectory selection}.

\paragraph{How many hints help?} All main experiments provide one hint per failed trajectory. Using the longest-8k subset from the previous comparison, we vary the number of hints per failed trajectory from 1 to 3, 5, and 8, keeping five continuation attempts. Mean accuracy is 66.78\% with one hint and 67.01\% with three, then falls to 66.74\% and 64.10\% with five and eight (Fig.~\ref{fig:current_D}b). Three hints yield the highest observed accuracy, 0.23 percentage points above one hint; additional hints reduce accuracy.

\begin{figure}[!htbp]
\centering
\includegraphics[width=\linewidth]{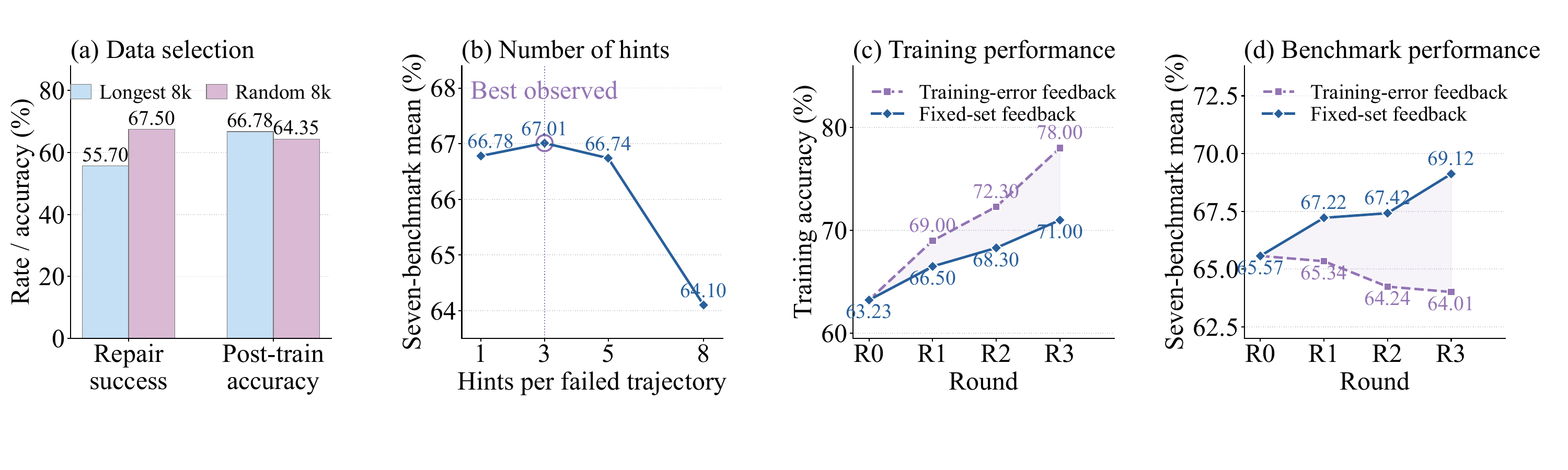}
\caption{\textbf{Selection, hints, and feedback (Qwen3.5-9B).} (a) Longest-8k versus random-8k questions: lower repair success can yield higher downstream accuracy. (b) Hint-count ablation uses the same longest-8k subset and run configuration as (a), with five continuation attempts; $k=1$ reproduces its 66.78\% single-hint baseline. Three hints perform best. (c,d) Training accuracy and seven-benchmark mean with training-error or fixed-set feedback. Feedback: 5,000 held-out questions.}
\label{fig:current_D}
\end{figure}

\begin{figure}[!htbp]
\centering
\includegraphics[width=\linewidth]{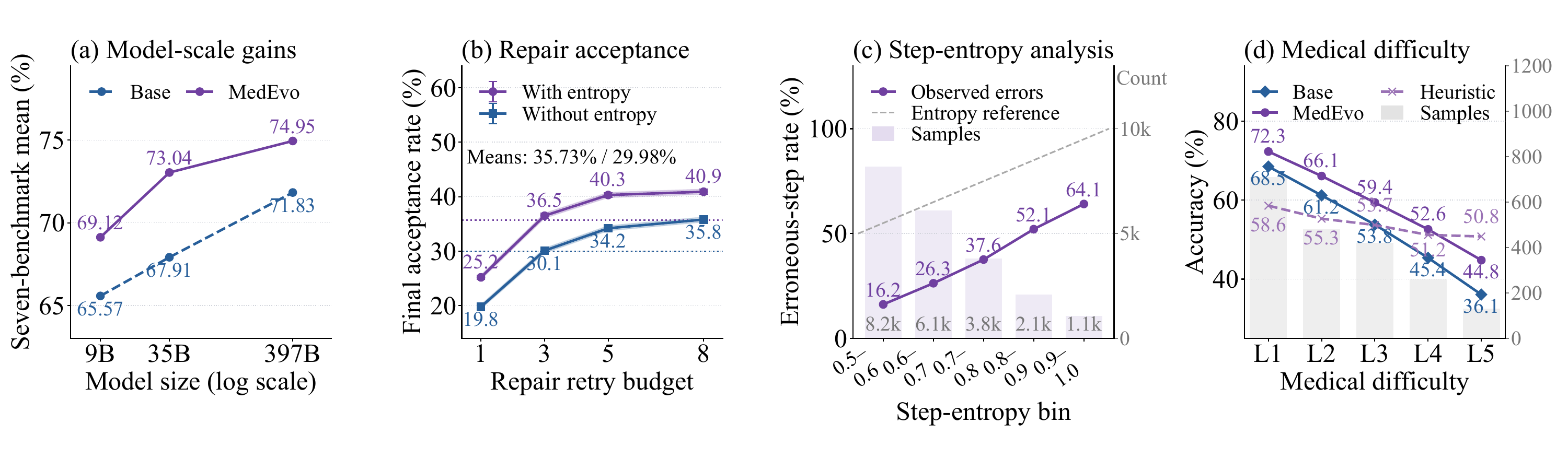}
\caption{\textbf{Scale, uncertainty, and difficulty.} (a) Mean accuracy after R3 for 9B/35B and R2 for 397B. (b) Repair acceptance with/without entropy localization; pointwise 95\% Wilson intervals use 45,000 attempts per setting. (c) Error rates and step counts by min--max-normalized step-entropy bin; the diagonal is $y=100x$. (d) Accuracy on a modality-stratified 2,000-question subset of the feedback set; counts use the right axis. Heuristic is a training-free HHD-inspired hint-recovery baseline (Appendix~\ref{app:analysis-settings}).}
\label{fig:four_panel_analysis}
\end{figure}

\paragraph{Which feedback guides evolution?} Training-error feedback raises training accuracy from 63.23\% to 78.00\% but lowers benchmark accuracy from 65.57\% to 64.01\%; fixed-set feedback yields 71.00\% training and 69.12\% benchmark accuracy (Fig.~\ref{fig:current_D}c,d). Failures supply repairs; \textbf{independent feedback reveals generalizable weaknesses}.\looseness=1\relax

\paragraph{Further analyses.} Fig.~\ref{fig:four_panel_analysis} summarizes additional repair analyses. Entropy localization increases final acceptance, defined as the fraction of repair candidates retained, at every continuation budget: from 19.8--35.8\% to 25.2--40.9\% for one to eight attempts (Fig.~\ref{fig:four_panel_analysis}b). The erroneous-step rate rises from 16.2\% to 64.1\% across increasing entropy bins, supporting \textbf{entropy as an audit prior} (Fig.~\ref{fig:four_panel_analysis}c). Relative to Base, MedEvo improves accuracy from 68.5\% to 72.3\% at L1 and from 36.1\% to 44.8\% at L5 (Fig.~\ref{fig:four_panel_analysis}d). MedEvo exceeds Heuristic at L1--L4, while Heuristic attains 50.8\% at L5 versus 44.8\% for MedEvo.

\Needspace{6\baselineskip}
\section{Reliability Analysis}
\label{sec:reliability}
We assess \textbf{repair reliability} through human audit and \textbf{MedEvo stability} across independent runs.
\subsection{Human Audit}
\label{sec:human-audit}
\begin{table}[!htbp]
\centering
\captionsetup{width=0.92\linewidth}
\fontsize{9.5}{11.5}\selectfont
\caption{\textbf{Human audit of repaired trajectories.} Adjudicated binary pass rates (\%) and pre-adjudication Cohen\textquotesingle s $\kappa$. The 95\% bootstrap CIs are for $\kappa$, resampling trajectories with the two independent ratings paired. V/R/D/K: visual, reasoning, differential-diagnosis, and knowledge errors. Repaired-output checks are shaded blue.}
\label{tab:human-audit}
\setlength{\tabcolsep}{2.5pt}
\renewcommand{\arraystretch}{1.15}
\begin{tabularx}{0.92\linewidth}{@{}>{\raggedright\arraybackslash}Xrccrrrr@{}}
\toprule
 & & \multicolumn{2}{c}{Agreement} & \multicolumn{4}{c}{Error family} \\
\cmidrule(lr){3-4}\cmidrule(l){5-8}
\rowcolor[HTML]{ECECEC}
Audit criterion & \multicolumn{1}{c}{Pass (\%)} & $\kappa$ & 95\% CI & \multicolumn{1}{c}{V} & \multicolumn{1}{c}{R} & \multicolumn{1}{c}{D} & \multicolumn{1}{c}{K} \\
\midrule
Original trajectory erroneous & 95.5 & 0.90 & [0.83, 0.97] & 96 & 96 & 94 & 96 \\
Valid repair boundary & 84.5 & 0.85 & [0.66, 0.88] & 88 & 84 & 82 & 84 \\
No answer leakage in hint & 97.2 & 0.94 & [0.87, 1.00] & 96 & 100 & 96 & 97 \\
\rowcolor[HTML]{E4F1FB}
Image-grounded repaired reasoning & \textbf{78.5} & 0.70 & [0.58, 0.82] & 76 & 80 & 76 & 82 \\
\rowcolor[HTML]{E4F1FB}
Correct repaired answer & \textbf{88.3} & 0.89 & [0.81, 0.97] & 88 & 89 & 88 & 88 \\
\bottomrule
\end{tabularx}
\end{table}
Two reviewers independently assess each trajectory using the same rubric and resolve disagreements through discussion. Pass rates use the resulting adjudicated binary labels. Cohen\textquotesingle s $\kappa$ uses the independent labels before adjudication. Tab.~\ref{tab:human-audit} reports results for the audited sample of repaired trajectories. Annotators identify 95.5\% of submitted trajectories as erroneous ($\kappa=0.90$) and validate 84.5\% of the selected repair boundaries ($\kappa=0.85$). Hints are answer-blind in 97.2\% of cases ($\kappa=0.94$). Repaired answers are judged correct in 88.3\% of cases ($\kappa=0.89$), and 78.5\% of repairs are judged \textbf{image-grounded}. Across the four error families, answer-correctness pass rates range from 88\% to 89\%, and grounding pass rates from 76\% to 82\%.

The audit assesses \textbf{complementary answer correctness and visual grounding}, with marginal pass rates of 88.3\% and 78.5\%, respectively. Agreement is higher for answer correctness ($\kappa=0.89$) than for grounding ($\kappa=0.70$). These findings support assessing answers and the visual evidence behind reasoning.

\subsection{Stability across Random Seeds}
\label{sec:seed-stability}
\begingroup
\normalsize
\newsavebox{\seedstabilitytable}
\setlength{\tabcolsep}{3pt}
\renewcommand{\arraystretch}{1.16}
\sbox{\seedstabilitytable}{%
\fontsize{9.5}{11.5}\selectfont
\begin{tabular}{lrrrlr}
\toprule
Model & Seed 1 & Seed 2 & Seed 3 & Mean $\pm$ SD & $\Delta$ \\
\midrule
Base & 65.57 & 65.57 & 65.57 & $65.57\pm0.00$ & --- \\
R1 & 67.22 & 67.05 & 67.38 & $67.22\pm0.17$ & \textcolor[HTML]{287DB1}{\tablechange{$\uparrow1.65$}} \\
R2 & \underline{67.42} & \underline{67.21} & \underline{67.60} & $\underline{67.41}\pm0.20$ & \textcolor[HTML]{287DB1}{\tablechange{$\uparrow1.84$}} \\
\rowcolor[HTML]{E4F1FB}
\textbf{R3} & \textbf{69.12} & \textbf{68.84} & \textbf{69.31} & $\boldsymbol{69.09\pm0.24}$ & \textcolor[HTML]{287DB1}{\tablechange{$\uparrow\mathbf{3.52}$}} \\
\bottomrule
\end{tabular}}

\Needspace{12\baselineskip}
\begin{wraptable}{r}{\wd\seedstabilitytable}
\vspace{-\intextsep}
\centering
\caption{\textbf{Stability across training runs (Qwen3.5-9B).} Seven-benchmark mean accuracy (\%), three runs; all improve through R3. Sample standard deviation (SD); $\Delta$: absolute gain over Base. Bold/underline: top two rounds; blue: R3.}
\label{tab:seed-stability}
\usebox{\seedstabilitytable}

\vspace{-0.5\baselineskip}
\end{wraptable}
Tab.~\ref{tab:seed-stability} reports three independent training runs from the same Base checkpoint. The \textbf{improvement is highly reproducible}. Every run follows the ordering Base $<$ R1 $<$ R2 $<$ R3, and R3 averages $69.09\%\pm0.24$, compared with 65.57\% for Base. Even the weakest run reaches 68.84\%, so the improvement far exceeds the variation across seeds. All three runs improve more from R2 to R3 than from R1 to R2.

Each round retrains from $M_0$, so \textbf{the evolving dataset carries information across rounds}. Longer training runs can characterize the timing and saturation of these gains.
\finishfigurewrap
\endgroup

\FloatBarrier
\endgroup

\section{Conclusion and Discussion}
\label{sec:conclusion}
We introduced MedEvo to study how failed medical reasoning trajectories can become useful supervision. \textbf{Knowledge Navigation} links local repair to subsequent data selection: a gold-blind auditor provides a minimal, answer-free hint, the student preserves supported reasoning and regenerates the erroneous suffix, and a fixed feedback set directs later repairs toward remaining weaknesses. \textbf{With SFT alone}, MedEvo improves Qwen3.5-9B, Qwen3.5-35B, and Qwen3.5-397B across seven benchmarks, with consistent gains across the three Qwen3.5-9B runs.

The central lesson from these experiments is that \textbf{correcting a response and constructing useful supervision are different objectives}. Preserving valid reasoning, selecting repairs for learning value, and using independent feedback each affect what the student learns. Their contribution should therefore be assessed through downstream learning, alongside repair success and evidence grounding.

\subsection{Insights for Medical Self-Evolution}
\label{sec:discussion}
The practical question is how to allocate repair effort: which parts of a failed trajectory remain useful, which corrections should become training targets, and which capability gaps warrant new data.

\paragraph{From diagnosis to intervention.} Medical tags describe the clinical context, while error types identify a recovery operation for the current trajectory. The two guide different decisions: tag-level feedback prioritizes questions, and local error diagnosis guides the hint and rollback boundary. The repair and feedback ablations (Tabs.~\ref{tab:current_nav} and~\ref{tab:current_evo}) support this division of roles within MedEvo.

\paragraph{Evolution through supervision.} The taxonomy and error codebook remain fixed in the present implementation; \textbf{selection priorities and accepted supervision change with the student}. Each round retrains from $M_0$ on the updated dataset, so information passes across rounds through supervision. Regressions and persistent weaknesses then redirect subsequent repair effort.

\paragraph{Failed trajectories retain valuable reasoning.} Suffix continuation outperforms full rewriting in both repair success and downstream accuracy (Tab.~\ref{tab:current_nav}), supporting the value of \textbf{preserving the valid prefix}. In the trajectory-selection experiment, the longest-8k subset has lower repair success than random selection (55.70\% vs.\ 67.50\%) but yields higher downstream accuracy (66.78\% vs.\ 64.35\%; Fig.~\ref{fig:current_D}a). Repair success alone would therefore rank these two choices differently from their downstream learning outcomes.

\paragraph{Repairability and learning value are different objectives.} The comparison above shows a divergence between repair success and learning value in the tested subsets; it does not isolate trajectory length from question difficulty. In the hint-count experiment, three hints yield the highest observed accuracy, only 0.23 percentage points above one hint, while five and eight hints perform worse (Fig.~\ref{fig:current_D}b). \textbf{Guidance should be calibrated against subsequent learning}, rather than increased solely to ease correction. Localization has a related distinction: the erroneous-step rate rises from 16.2\% to 64.1\% across entropy bins (Fig.~\ref{fig:four_panel_analysis}c), making entropy a useful inspection cue in this analysis, while the evidence-grounded audit determines whether an error is present and where regeneration should begin.

{\emergencystretch=1em
\paragraph{The source of feedback matters.} Training-error feedback improves fit to the training data while benchmark accuracy declines; fixed-set feedback improves benchmark accuracy across rounds (Fig.~\ref{fig:current_D}c,d). In these experiments, \textbf{training failures supply repair candidates, while independent feedback helps prioritize capability gaps}. The distinction prevents progress on the training examples from serving as the sole criterion for allocating later repair effort.\par}

\paragraph{Answer correctness and grounding are complementary.} The human audit assesses answer correctness and visual grounding separately (\S\,\ref{sec:human-audit}). Its results motivate checking \textbf{evidence use alongside accuracy} and hint non-leakage when admitting repaired trajectories as supervision.

\subsection{Limitations and Future Directions}
\label{sec:limitations}
\textbf{The current implementation has a fixed knowledge structure}. Its taxonomy and error codebook constrain which weaknesses can be represented, and a strong proprietary auditor supplies the diagnoses. Evaluating open-source auditors and separating the effects of trajectory length from question difficulty would clarify these dependencies. Our benchmarks use multiple-choice and short-answer formats; deployment requires \textbf{clinical evaluation}.

\paragraph{From diagnosis to grounded recovery.} Future work could study \textbf{evidence-grounded recovery} through source-attributed retrieval, factuality checks, and image-region supervision, evaluating answer leakage, grounding, and downstream learning jointly.

\paragraph{Adaptive navigation.} Persistent failures could guide expert-validated updates to the taxonomy and recovery operations. Longer runs on independent data would help assess the regression, cost, and saturation of such \textbf{knowledge-structure evolution}.

\paragraph{Transferable learning strategies.} Further studies could test repair transfer across modalities, specialties, and institutions, while accounting for annotation and auditor budgets. Relevant outcomes include transfer, rare-condition coverage, and robustness under distribution shift.

\paragraph{Declaration of Artificial Intelligence (AI) Assistance.}
AI tools were used solely for language editing, manuscript formatting, and figure preparation. They were not involved in method development or experimental design. All AI-assisted content was reviewed and revised by the authors, who take full responsibility for the final manuscript.

\paragraph{Reproducibility Statement.}
The main text and appendices describe the method, prompts, training configurations, and evaluation procedures. The present release is limited to the manuscript; the implementation and training-data files are not released. Use of the underlying public datasets remains subject to their licenses and access restrictions.

\bibliographystyle{plainnat}
\setcitestyle{numbers}
\bibliography{ref}

\Needspace{18\baselineskip}
\beginappendix
\definecolor{AppendixTopic}{HTML}{70528F}
\definecolor{AppendixAction}{HTML}{285F9A}
\newcommand{\appkey}[1]{\textcolor{AppendixAction}{\textbf{#1}}}
\newcommand{\apptopic}[1]{\textcolor{AppendixTopic}{\textbf{#1}}}
\newcommand{\casestage}[1]{\par\smallskip\noindent\apptopic{#1}\par\nobreak}

\section*{Appendix Contents and Reading Guide}
This guide links to each supplementary section and its starting page.
\par\vspace{7pt}

\begingroup
\normalsize
\setlength{\tabcolsep}{5pt}
\setlength{\aboverulesep}{3.5pt}
\setlength{\belowrulesep}{3.5pt}
\renewcommand{\arraystretch}{1.13}
\newcommand{\guideentry}[3]{\ref{#1} & \textbf{\nameref{#1}}. #3 & \pageref{#1}\\\addlinespace[3pt]}
\begin{xltabular}{\linewidth}{@{}p{0.085\linewidth} X r@{}}
\toprule
Section & Contents & Page\\
\midrule
\endfirsthead
\toprule
Section & Contents (continued) & Page\\
\midrule
\endhead
\endfoot
\endlastfoot
\guideentry{app:method-overview}{Method Details}{Algorithms, mathematical formulation, and implementation of navigation and local repair.}
\guideentry{app:failure-segmentation}{Failure Definition and Step Segmentation}{Failed answers, semantic steps, and prefix boundaries.}
\guideentry{app:localization-details}{Entropy-Based Localization}{Step scores, audit windows, and proposal assessment.}
\guideentry{app:audit-operators}{Gold-Blind Audit and Recovery Operators}{Auditor inputs, recovery hints, and cue rendering.}
\guideentry{app:training-details}{Continuation, Admission, and Training Targets}{Budgets, learnability gates, admission, and fallback.}
\guideentry{app:planning-details}{Feedback Statistics, Quota Allocation, and Dataset Update}{Transitions, quotas, retrieval, and target replacement.}
\guideentry{app:analysis-settings}{Settings of Analysis Experiments}{Repair, feedback, hint-count, and entropy comparisons.}
\midrule
\guideentry{app:experiments-overview}{Dataset Statistics}{Dataset sizes and roles, round-wise construction counts, and detailed feedback-set measurements.}
\guideentry{app:exp-supplement}{Additional Experimental Details}{Evolution of twenty error subtypes, plus round-wise modality and error-family statistics.}
\guideentry{app:detailed-error-statistics}{Detailed Error Statistics for the Current Rounds}{Numerical rates for all twenty subtypes and repair attempts, successes, and success rates by error family.}
\midrule
\guideentry{app:taxonomy-overview}{Medical Taxonomy and Error Categories}{Medical tags, error definitions, information-access rules, and reproducibility specifications.}
\guideentry{app:taxonomy}{Taxonomy, Permissions, and Execution Contracts}{Four-axis tag dictionary; four error families and twenty subtypes; access matrix; structured records and plan validation; training configuration and evaluation scope.}
\midrule
\guideentry{sec:appendix}{Prompts}{Templates for answering questions, planning feedback-driven rounds, auditing failures, and continuing repaired trajectories.}
\guideentry{app:shared_qa_prompt}{Shared Question-Answering Prompt}{The common question-answering instruction.}
\guideentry{app:feedback-agent-prompt}{Main-Agent Feedback and Next-Round Planning}{Inputs, constraints, and structured outputs for diagnosing weaknesses and planning the next training round.}
\guideentry{app:stage1}{Stage 1 Audit Prompt}{Failure diagnosis, localization, and construction of an error-conditioned recovery cue.}
\guideentry{app:stage2}{Stage 2 Continuation Prompt}{Same-student continuation from the retained prefix and recovery cue.}
\midrule
\guideentry{app:specialty}{Case Study}{Worked radiograph examples showing the question, image, retained prefix, hint, and continuation.}
\guideentry{app:case08}{Case 08: Chest radiograph}{Chest-image example of local trajectory repair.}
\guideentry{app:case31}{Case 31: Elbow radiograph}{Elbow-image example of local trajectory repair.}
\midrule
\guideentry{app:broader-related-work}{Additional Related Work}{Extended medical-AI background, hallucinations, supervision quality, correction methods, and broader self-evolution settings.}
\guideentry{app:baseline-details}{Additional Baseline Details}{Backbones, comparison models, trajectory-supervision baselines, and the evaluation protocol.}
\bottomrule
\end{xltabular}
\addtocounter{table}{-1}
\endgroup

\par\medskip

\section{Method Details}
\label{app:method-overview}
This appendix provides implementation details that complement \S\,\ref{sec:method}. Within round $t$, \appkey{the generator $M_t=\pi_{\theta_t}$ is frozen}: it produces rollouts, entropy profiles, continuations, and learnability scores, while the next student is trained from $M_0$ (Eq.~\ref{eq:outer-update}).

\subsection{Failure Definition and Step Segmentation}
\label{app:failure-segmentation}
Given a question $x=(I,q,O)$ with reference answer $y\in O$, $M_t$ samples a trajectory $r$ and a parsed answer $\hat y$. A trajectory $r^-$ fails if $\hat y\ne y$ or no valid answer can be parsed. Each failed trajectory is segmented into $K$ \appkey{semantically complete steps} (Eq.~\ref{eq:main-step-segmentation}), typically covering image observation, feature interpretation, mechanism, differential diagnosis, and option mapping. A boundary $b\in\{0,\dots,K-1\}$ splits $r^-$ into the preserved prefix $p_b=(s_1,\dots,s_b)$ and the erroneous suffix $s^-=(s_{b+1},\dots,s_K)$ (Eq.~\ref{eq:main-prefix-split}), where $b=0$ denotes an empty prefix.

\subsection{Entropy-Based Localization}
\label{app:localization-details}
\paragraph{Step scores.} Token entropy $H_j$, the step score $E_k^*$, and the local surge $\Delta E_k$ follow Eqs.~\ref{eq:main-token-entropy}--\ref{eq:main-step-entropy}, with entropy measured in nats. We use $\lambda=0.5$ and $m=3$, with $m_k=\min(m,k-1)$ for $k>1$ and $\Delta E_1=0$. Combining the mean with the 90th percentile $Q_{0.9}$ balances average uncertainty with an upper-tail signal within each step.

\paragraph{Anchor and window.} Both profiles are \appkey{min--max normalized} within each trajectory, giving $\widetilde E_k^*$ and $\widetilde{\Delta E}_k$; a constant profile maps to zero. The anchor is $k^*\in\operatorname*{arg\,max}_{1\le k\le K}(\widetilde E_k^*+\widetilde{\Delta E}_k)$, and the audit window is
\begin{equation}
\mathcal W(k^*)=\{s_{\max(1,\,k^*-w_\ell)},\dots,s_{\min(K,\,k^*+w_r)}\},\qquad w_\ell=w_r=1.
\label{eq:appendix-window}
\end{equation}
The step preceding the window defines a provisional boundary
\begin{equation}
b_0=\max(0,\;k^*-w_\ell-1).
\label{eq:appendix-provisional-boundary}
\end{equation}
Both $\mathcal W(k^*)$ and $b_0$ are proposals; \appkey{the auditor sets the final boundary} $b$ (Appendix~\ref{app:audit-operators}).

\paragraph{Proposal assessment.} For every audited trajectory, the auditor classifies the entropy proposal as HIT (the first consequential error lies at or adjacent to $k^*$), PARTIAL\_HIT (the window covers the propagated error, but the unsupported premise begins earlier), MISS (the first error lies outside the window), or NON\_ERROR\_UNCERTAINTY (the high-entropy region is medically and logically valid). These logged labels support Fig.~\ref{fig:four_panel_analysis}c.

\subsection{Gold-Blind Audit and Recovery Operators}
\label{app:audit-operators}
\paragraph{Auditor.} We use GPT-5.6-sol as the auditor $\mathcal A$ in Eq.~\ref{eq:main-auditor}. Its inputs are the images, the question and options, the failed trajectory, the entropy profile ($E_k^*$, normalized scores, $\Delta E_k$, $k^*$, and $\mathcal W(k^*)$), and \appkey{observable tags with provenance}. The reference answer, reference rationale, answer-derived tags, and retrieval metadata are \appkey{excluded through allowlist serialization} (Appendix~\ref{app:permissions}). The auditor returns a structured record (Appendix~\ref{app:stage1}) containing an action in $\{\text{\textsc{Repair}},\text{\textsc{Skip}}\}$, the first consequential error $e$ with its type and subtype, the boundary $b$ with a verbatim ending span (from $s_b$ if $b>0$, empty if $b=0$), and the hint $h$. It returns \textsc{Skip} when only answer formatting or option selection is wrong and no reasoning revision is justified. The final boundary may differ from $b_0$. The auditor \appkey{traces evidence dependencies backward} and moves the boundary earlier when a later entropy surge stems from an upstream unsupported premise; conversely, it retains a high-entropy step that is medically and logically valid.

\paragraph{Recovery operators.} An error type selects a recovery operator $o$; the text hint $h$ instantiates that operator in natural language. The operator belongs to
\begin{equation}
o\in\{\mathcal P_{\mathrm{inspect}},\ \mathcal R_{\mathrm{reconstruct}},\ \mathcal D_{\mathrm{disambiguate}},\ \mathcal K_{\mathrm{bridge}}\},
\label{eq:appendix-operators}
\end{equation}
each paired with a parent error type (\S\,\ref{sec:main-error-types}); subtype-level operations are listed in Appendix~\ref{app:error-codebook}. Operators define recovery actions.
\begin{itemize}[leftmargin=*]
\item \appkey{Perceptual re-examination} ($\mathcal P_{\mathrm{inspect}}$) names a region or attribute to revisit: lesion distribution, boundary shape, waveform morphology, stain localization, or cross-view consistency, without stating the finding.
\item \appkey{Inference reconstruction} ($\mathcal R_{\mathrm{reconstruct}}$) asks students to link evidence, mechanism, diagnosis, and decision, for example when supportive and causal treatments are confused or care steps are misordered.
\item \appkey{Differential disambiguation} ($\mathcal D_{\mathrm{disambiguate}}$) requests comparison of discriminative features: contrast pattern, onset profile, lesion location, signature pathology, or an exclusion criterion, without naming the diagnosis.
\item \appkey{Knowledge bridging} ($\mathcal K_{\mathrm{bridge}}$) supplies a narrowly scoped concept, such as a diagnostic criterion, drug mechanism, or immunohistochemical localization rule. It may state the concept but must not map it to an answer option.
\end{itemize}
\paragraph{Rendering.} The cue $g(h)$ consists of a natural transition followed by the hint, with flexible wording. Error labels and audit explanations are \appkey{stored as metadata} and are never shown to the student.

\subsection{Continuation, Admission, and Training Targets}
\label{app:training-details}
\paragraph{Continuation budget.} The current checkpoint generates eight rollouts per selected question. Each failed trajectory selected for repair is audited once and receives a single hint. With $p_b\oplus g(h)$ fixed, up to five suffixes are sampled (Eq.~\ref{eq:main-continuation}); \appkey{sampling stops at the first admitted continuation}. The user message contains $x$, the unfinished assistant message contains $p_b\oplus g(h)$, and generation continues within the same message \appkey{without repeating or modifying the prefix} (Appendix~\ref{app:stage2}).

\paragraph{Learnability gate.} Using the suffix scores $D_{\mathrm{repair}}$ and $\Delta_{\mathrm{repair}}$ (Eqs.~\ref{eq:main-suffix-difficulty}--\ref{eq:main-preference-gap}), computed under $M_t$ \appkey{without the hint},
\begin{equation}
L_\xi=\mathbb{1}\!\left[D_{\min}\le D_{\mathrm{repair}}(\xi)\le D_{\max}\ \wedge\ \Delta_{\min}\le\Delta_{\mathrm{repair}}(\xi)\le\Delta_{\max}\right],
\label{eq:method-learnability-gate}
\end{equation}
with $[D_{\min},D_{\max}]=[0.35,0.85]$ and $[\Delta_{\min},\Delta_{\max}]=[0.02,0.30]$. The lower bound on $\Delta_{\mathrm{repair}}$ requires that the student still prefers its original wrong suffix, so the repair carries information the student has not yet acquired. The upper bound limits the increase in per-token loss relative to the wrong suffix. Thresholds are selected on the feedback set and \appkey{frozen before final evaluation}. Feedback-set questions are \appkey{excluded from training targets}. Thresholds, retained fractions, and generator hashes are logged with each repair record (Appendix~\ref{app:contracts}).

\paragraph{Admission checks.} Admission follows Eq.~\ref{eq:main-admission-gate}. $C_\xi$ compares the parsed answer with the reference label. $A_\xi$ \appkey{verifies that the exported prefix matches} the auditor's verbatim span and that $b$ precedes the diagnosed error. $N_\xi$ \appkey{rejects continuations} that introduce symptoms, history, measurements, or findings absent from the input, as well as repetition loops. Reject candidates with either suffix empty, nonfinite losses, or a nonpositive denominator in Eq.~\ref{eq:main-suffix-difficulty}.

\paragraph{Training targets and fallback.} Admitted repairs are exported as $r^*=p_b\oplus g(h)\oplus s^+$ and trained with Eq.~\ref{eq:main-repair-loss}; original correct rollouts and historical repairs use the same \appkey{full-response loss}, and image and user tokens are masked. If no repair is admitted for a question, a quality-screened correct rollout or the existing valid target is retained; if neither exists, the question contributes no target. \appkey{No hint or auditor output is used at evaluation}.

\subsection{Feedback Statistics, Quota Allocation, and Dataset Update}
\label{app:planning-details}\label{app:feedback-selection-details}
\paragraph{Feedback statistics.} Besides the error rate in Eq.~\ref{eq:main-feedback}, consecutive checkpoints yield transition counts
\begin{equation}
N^{uv}_{g,t}=\sum_{i\in V_{\mathrm{feedback},g}}\mathbb{1}\!\left[c_{i,t-1}=u\ \wedge\ c_{i,t}=v\right],\qquad u,v\in\{0,1\},\quad t\ge1,
\label{eq:appendix-transitions}
\end{equation}
which separate \appkey{corrected errors} ($N^{01}$), \appkey{regressions} ($N^{10}$), \appkey{persistent errors} ($N^{00}$), and \appkey{retained successes} ($N^{11}$). The priority in Eq.~\ref{eq:main-priority} normalizes regressions by $n_g$. For diagnostics, we also report the conditional regression rate $N^{10}_{g,t}/(N^{10}_{g,t}+N^{11}_{g,t})$ only when its denominator is positive; it is undefined and omitted when the preceding checkpoint has no correct examples. Groups comprise supported single-axis tags and predeclared intersections. A group with $n_g<n_{\min}$ backs off to its nearest supported ancestor, and identical resolved nodes are merged. Each question is counted once within a group, and overall statistics are computed over the deduplicated union of questions.

\paragraph{Quota allocation.} From the second round on, the budget $B_t$ is allocated in proportion to $S_{g,t}$ (Eq.~\ref{eq:main-priority}), using largest-remainder rounding with deterministic tie-breaking. Quotas are \appkey{capped by the available unique training questions}, and surplus capacity is \appkey{redistributed among eligible groups}. Each multi-tag question is charged to the budget once. If all scores are zero, questions are sampled uniformly from eligible unique questions. Token and auditor-call caps impose additional limits, and any unfilled quota is recorded. The priority and quota rule is Eq.~\ref{eq:main-priority}; the planning template in Appendix~\ref{app:feedback-agent-prompt} assembles a plan and tag-based retrieval requests subject to these numerical constraints. The validator in Appendix~\ref{app:contracts} checks the plan before execution.

\paragraph{Retrieval and repair priority.} Within each group, candidates are retrieved by tag match and ranked by historical error records. Candidates may include questions already used in training. Selected questions receive \appkey{fresh rollouts} from $M_t$, and their failures are diagnosed anew. Failures matching both the modality and the error type prioritized by feedback are repaired first, followed by modality-only or error-type-only matches. Matching uses \appkey{current diagnoses}. Priority determines where repair effort is spent; admission (Appendix~\ref{app:training-details}) determines what enters training.

\paragraph{Dataset update.} Questions are deduplicated by canonical ID, and each keeps \appkey{at most one active target}. An admitted candidate replaces the existing target only if a comparison auditor, blinded to round and model labels, judges it better in evidence support, reasoning consistency, and conciseness; \appkey{ties retain the existing target}. All versions and rollout-specific diagnoses are logged in a separate history. Each round retrains from $M_0$ on the complete updated dataset (Eqs.~\ref{eq:dataset-update}--\ref{eq:outer-update}) with a fresh optimizer and learning-rate schedule. Generation caps are fixed in advance, and plan records follow Appendix~\ref{app:contracts}.

\subsection{Settings of Analysis Experiments}
\label{app:analysis-settings}
Unless stated otherwise, analyses use Qwen3.5-9B with the training configuration in Tab.~\ref{tab:training35b} and evaluation in \S\,\ref{sec:exp:setup}.

\paragraph{Repair designs (Tab.~\ref{tab:current_nav}).} All variants are applied in the first round under the same rollout and continuation budgets. Hint + full rewrite uses the type-guided hint but regenerates the trajectory from the start of the response. No-hint rewrite regenerates the full trajectory without a hint. Generic hint + suffix continuation keeps the auditor's boundary but replaces the type-guided hint with a fixed generic cue.

\paragraph{Feedback ablations (Tab.~\ref{tab:current_evo}).} All variants share the R1 data-generation model and historical dataset and use the same budgets. R2 data are generated by the shared R1 model; R3 data are generated by each branch\textquotesingle s own R2 model. In every branch and round, SFT starts from $M_0$ with a fresh optimizer and learning-rate schedule. No feedback omits feedback-guided question selection. Without medical tags omits medical-tag feedback. Without error-type feedback uses tag-group quotas but does not match repairs to prioritized error types.

\paragraph{Trajectory selection (Fig.~\ref{fig:current_D}a).} The longest-8k subset contains the 8,000 questions whose failed trajectories are longest; the random-8k subset is sampled uniformly. Each subset is used for one round of continuation training from $M_0$.

\paragraph{Hint count (Fig.~\ref{fig:current_D}b).} Using the same longest-8k subset, $M_0$ initialization, and one-round training configuration as Fig.~\ref{fig:current_D}a, we use $k\in\{1,3,5,8\}$ hints per failure and five continuation attempts. The $k=1$ setting reproduces the single-hint baseline of 66.78\%.

\paragraph{Feedback source (Fig.~\ref{fig:current_D}c,d).} Training-error feedback selects by training errors, replacing feedback-set errors.

\paragraph{Retry budget and entropy (Fig.~\ref{fig:four_panel_analysis}b).} We vary the continuation budget over $\{1,3,5,8\}$ with and without entropy localization. Final acceptance is the admitted fraction over 45,000 attempts per setting, with 95\% Wilson intervals.

\paragraph{Step-entropy analysis (Fig.~\ref{fig:four_panel_analysis}c).} The horizontal axis uses min--max-normalized step entropy rather than raw entropy in nats. The five equal-width bins have boundaries $0.5,0.6,0.7,0.8,0.9,1.0$; the statistical unit is a reasoning step. Each bin reports the erroneous-step rate and the number of steps. The dashed line is the numerical reference $y=100x$, with $y$ expressed as a percentage, rather than an error-probability calibration target.

\paragraph{Difficulty analysis and Heuristic baseline (Fig.~\ref{fig:four_panel_analysis}d).} We sample 2,000 questions from the 5,000-question feedback set using modality-stratified sampling. The difficulty counts are 710, 480, 420, 260, and 130 for L1--L5, following the rubric in Appendix~\ref{app:tag-dictionary}. This subset is used only for difficulty analysis and does not enter training; capability-gap assessment in every round uses the full 5,000-question feedback set. Thus, this analysis characterizes performance on feedback-set questions.

Heuristic is an HHD-inspired hint-recovery variant~\citep{DBLP:journals/corr/abs-2605-11556}, evaluated on the same difficulty-analysis subset as MedEvo. It takes the image and question as input and performs direct inference without additional training. It uses hint-based recovery without error localization, prefix preservation, or feedback-driven data evolution. This variant serves as a difficulty-analysis baseline. At L5, it attains 50.8\% accuracy versus 44.8\% for MedEvo; MedEvo is higher at L1--L4. The L5 result indicates that HHD-style hint recovery remains competitive on the hardest stratum.

\FloatBarrier
\section{Dataset Statistics}
\label{app:experiments-overview}
\subsection{Dataset Roles and Evaluation Scope}
\phantomsection\label{app:dataset-statistics}
The study separates the training bank, the fixed feedback set, and the final evaluation benchmarks. Training questions supply supervision; feedback guides diagnosis and selection; benchmarks measure performance.

\begin{table}[htbp]
\centering\small
\caption{\textbf{Dataset roles and sizes.} Test coverage is counted in benchmarks.}
\begin{tabularx}{\linewidth}{l r X}
\toprule
\rowcolor[HTML]{ECECEC}
Resource & Size & Role\\
\midrule
Training bank & 81,293 questions & Source of original correct rollouts, accepted repairs, and retained valid targets.\\
\rowcolor[HTML]{E4F1FB}
Fixed feedback set & 5,000 questions & Medical capability diagnosis and subsequent data selection; excluded from training.\\
Final evaluation & 7 benchmarks & Independent downstream accuracy reporting.\\
\bottomrule
\end{tabularx}
\end{table}

The training bank is constructed from public medical training splits and deduplicated using question and source identifiers, available patient or case identifiers, image hashes, and near-duplicate clusters. See Appendix~\ref{app:contracts} for structured records and plan validation, and Appendix~\ref{app:permissions} for information-access rules. Fig.~\ref{fig:dataset_overview} shows nine modality categories and five difficulty levels; the tag hierarchy is described in Appendix~\ref{app:tag-dictionary}.

Evaluation uses MedUniEval from Hulu-Med~\citep{jiang2025hulumed}, built on Lingshu's MedEvalKit~\citep{xu2025lingshu}. The seven benchmarks are PMC-VQA~\citep{zhang2023pmcvqa}, VQA-RAD~\citep{lau2018vqarad}, SLAKE~\citep{liu2021slake}, PathVQA~\citep{he2020pathvqa}, OmniMedVQA~\citep{hu2024omnimedvqa}, MedXpertQA-MM~\citep{DBLP:conf/icml/ZuoQLCZHZ0025}, and the medical subset of MMMU~\citep{yue2023mmmu}. Their accuracies are averaged without weighting by benchmark size. Modality and error-family results below describe the feedback set, separately from benchmark evaluation.

\subsection{Round-wise Data Construction}
\label{app:construction-counts}

Tab.~\ref{tab:current_counts} highlights a declining repair workload across training rounds. The number of questions submitted for repair falls from \textbf{27,233 to 23,620}, a reduction of \textbf{3,613 questions (13.27\%)} relative to R1. Their share of the fixed question bank decreases from 33.50\% to 29.06\%. The unit of this workload measure is a submitted question.

The final training set remains fixed at \textbf{81,293 questions} in each round, combining qualified original correct trajectories with \textbf{accepted repairs} and retained valid historical targets. Thus, repair workload decreases while training coverage stays constant. The number of successfully repaired questions decreases from 19,567 to 16,797 as fewer questions are submitted. Repair success rates are 71.85\%, 70.68\%, and 71.11\%, remaining stable across rounds. With at most five continuation attempts per repair, the mean number of attempts also falls from \textbf{3.37 to 2.12}, indicating improved hint-guided correction efficiency. Each column identifies the generator and training round.

\subsection{Additional Experimental Details}
\label{app:exp-supplement}

\subsubsection{Evolution across Twenty Error Subtypes}
\label{app:subtype-evolution}
\begin{figure}[!htbp]
\centering
\includegraphics[width=0.85\linewidth]{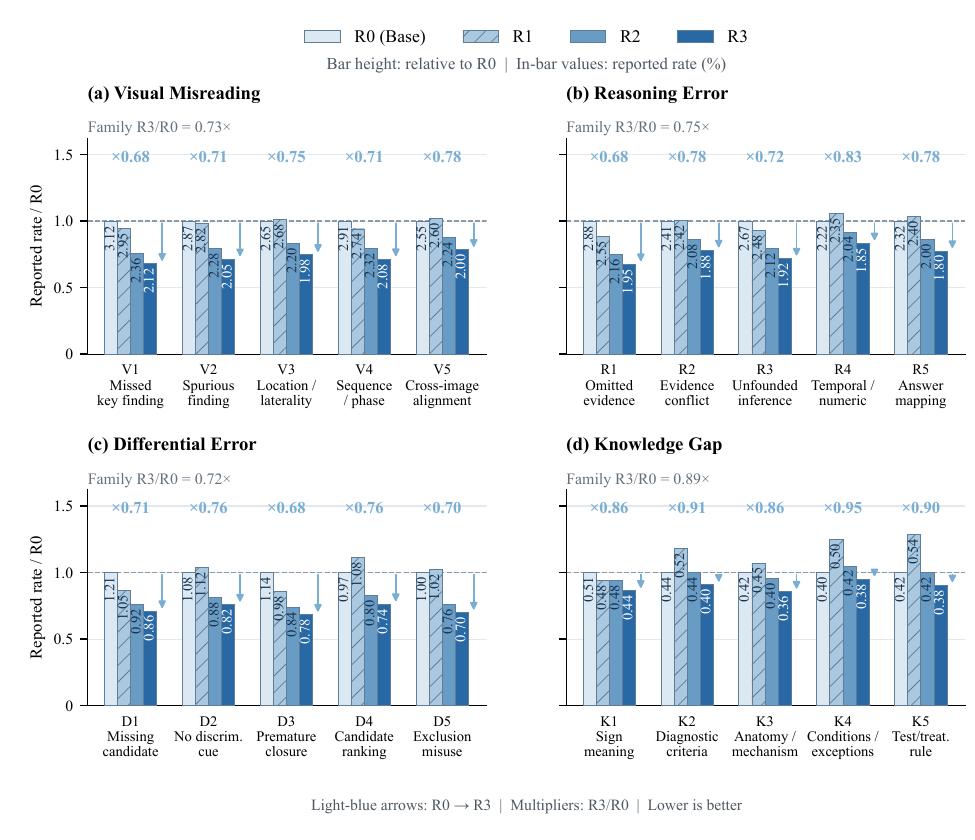}
\caption{\textbf{Error-subtype evolution (Qwen3.5-9B).} Bar heights normalize each subtype rate by R0; labels show original percentages. Arrows and multipliers compare R3 with R0. Family ratios divide summed R3 rates by summed R0 rates. Horizontal R1--R5 codes identify reasoning-error subtypes; legend R0--R3 labels identify training rounds.}
\label{fig:current_subtypes}
\end{figure}

Fig.~\ref{fig:current_subtypes} shows twenty subtypes across four families. Bars normalize rates by subtype R0 values; labels retain reported percentages. All subtypes are below R0 at R3. The knowledge-gap family has an R3/R0 ratio of approximately 0.89, compared with 0.72--0.75 for the other families, indicating a smaller relative reduction in the reported breakdown, conditional on a consistent counting denominator across rounds. Rates remain separate by subtype.

\subsubsection{Round-wise Feedback Statistics}
\label{app:feedback-tables}
Tab.~\ref{tab:current_modality} and~\ref{tab:current_errors} report the per-round modality error rates and error-family percentages; \S\,\ref{sec:current:rounds} summarizes the modality trends. Blue downward arrows indicate decreases, red upward arrows increases, and equality signs unchanged rates. The error-family percentages are reported separately from the modality-level error rates.

\begin{table}[!htbp]
\begin{minipage}[t]{0.49\linewidth}
\vspace{0pt}

\centering\fontsize{9.5}{11.5}\selectfont
\caption{\textbf{Modality-wise feedback errors.} Error rates (\%) on the fixed feedback set; $N$: group size. Arrows compare consecutive rounds. Bold/underline: two lowest rates; blue: R3.}
\label{tab:current_modality}
\setlength{\tabcolsep}{1.5pt}
\renewcommand{\arraystretch}{1.15}
\begin{tabular}{lrrrlrl>{\columncolor[HTML]{E4F1FB}}r>{\columncolor[HTML]{E4F1FB}}l}
\toprule
\rowcolor[HTML]{ECECEC}
Modality & N & R0 & \multicolumn{2}{c}{R1} & \multicolumn{2}{c}{R2} & \multicolumn{2}{c}{R3} \\
\midrule
Radiology & 1509 & 37.18 & 35.98 & \textcolor[HTML]{287DB1}{\tablechange{$\downarrow$}} & \textbf{31.28} & \textcolor[HTML]{287DB1}{\tablechange{$\downarrow$}} & \underline{32.21} & \textcolor[HTML]{E34B43}{\tablechange{$\uparrow$}} \\
Nuclear medicine & 234 & 34.62 & 26.50 & \textcolor[HTML]{287DB1}{\tablechange{$\downarrow$}} & \underline{24.79} & \textcolor[HTML]{287DB1}{\tablechange{$\downarrow$}} & \textbf{22.65} & \textcolor[HTML]{287DB1}{\tablechange{$\downarrow$}} \\
Ultrasound & 351 & 44.73 & \underline{39.03} & \textcolor[HTML]{287DB1}{\tablechange{$\downarrow$}} & 39.60 & \textcolor[HTML]{E34B43}{\tablechange{$\uparrow$}} & \textbf{36.18} & \textcolor[HTML]{287DB1}{\tablechange{$\downarrow$}} \\
Endoscopy & 247 & 32.39 & 30.36 & \textcolor[HTML]{287DB1}{\tablechange{$\downarrow$}} & \underline{27.13} & \textcolor[HTML]{287DB1}{\tablechange{$\downarrow$}} & \textbf{21.86} & \textcolor[HTML]{287DB1}{\tablechange{$\downarrow$}} \\
Pathology & 378 & 34.13 & 34.66 & \textcolor[HTML]{E34B43}{\tablechange{$\uparrow$}} & \underline{30.95} & \textcolor[HTML]{287DB1}{\tablechange{$\downarrow$}} & \textbf{28.31} & \textcolor[HTML]{287DB1}{\tablechange{$\downarrow$}} \\
Ophthalmology & 135 & 34.07 & 27.41 & \textcolor[HTML]{287DB1}{\tablechange{$\downarrow$}} & \underline{22.96} & \textcolor[HTML]{287DB1}{\tablechange{$\downarrow$}} & \textbf{20.00} & \textcolor[HTML]{287DB1}{\tablechange{$\downarrow$}} \\
Dermatology & 593 & 30.35 & 27.82 & \textcolor[HTML]{287DB1}{\tablechange{$\downarrow$}} & \underline{24.62} & \textcolor[HTML]{287DB1}{\tablechange{$\downarrow$}} & \textbf{23.10} & \textcolor[HTML]{287DB1}{\tablechange{$\downarrow$}} \\
Dental imaging & 308 & 40.26 & \underline{37.01} & \textcolor[HTML]{287DB1}{\tablechange{$\downarrow$}} & 37.66 & \textcolor[HTML]{E34B43}{\tablechange{$\uparrow$}} & \textbf{32.79} & \textcolor[HTML]{287DB1}{\tablechange{$\downarrow$}} \\
Functional tests & 1245 & 20.00 & 17.75 & \textcolor[HTML]{287DB1}{\tablechange{$\downarrow$}} & \underline{16.22} & \textcolor[HTML]{287DB1}{\tablechange{$\downarrow$}} & \textbf{15.50} & \textcolor[HTML]{287DB1}{\tablechange{$\downarrow$}} \\
\bottomrule
\end{tabular}
\end{minipage}\hfill
\begin{minipage}[t]{0.49\linewidth}
\vspace{0pt}

\centering\fontsize{9.5}{11.5}\selectfont
\caption{\textbf{Error-family feedback profiles.} Percentages sum the five subtypes in each family. These aggregates are not directly comparable to modality error rates. Notation follows Tab.~\ref{tab:current_modality}.}
\label{tab:current_errors}
\setlength{\tabcolsep}{1.5pt}
\renewcommand{\arraystretch}{1.15}
\begin{tabular}{lrrlrl>{\columncolor[HTML]{E4F1FB}}r>{\columncolor[HTML]{E4F1FB}}l}
\toprule
\rowcolor[HTML]{ECECEC}
Error family & R0 & \multicolumn{2}{c}{R1} & \multicolumn{2}{c}{R2} & \multicolumn{2}{c}{R3} \\
\midrule
Visual misreading & 14.10 & 13.79 & \textcolor[HTML]{287DB1}{\tablechange{$\downarrow$}} & \underline{11.40} & \textcolor[HTML]{287DB1}{\tablechange{$\downarrow$}} & \textbf{10.23} & \textcolor[HTML]{287DB1}{\tablechange{$\downarrow$}} \\
Reasoning error & 12.50 & 12.20 & \textcolor[HTML]{287DB1}{\tablechange{$\downarrow$}} & \underline{10.40} & \textcolor[HTML]{287DB1}{\tablechange{$\downarrow$}} & \textbf{9.40} & \textcolor[HTML]{287DB1}{\tablechange{$\downarrow$}} \\
Differential error & 5.40 & 5.25 & \textcolor[HTML]{287DB1}{\tablechange{$\downarrow$}} & \underline{4.20} & \textcolor[HTML]{287DB1}{\tablechange{$\downarrow$}} & \textbf{3.90} & \textcolor[HTML]{287DB1}{\tablechange{$\downarrow$}} \\
Knowledge gap & 2.19 & 2.49 & \textcolor[HTML]{E34B43}{\tablechange{$\uparrow$}} & \underline{2.16} & \textcolor[HTML]{287DB1}{\tablechange{$\downarrow$}} & \textbf{1.96} & \textcolor[HTML]{287DB1}{\tablechange{$\downarrow$}} \\
\bottomrule
\end{tabular}
\end{minipage}
\end{table}

\Needspace{6\baselineskip}
\subsection{Detailed Error Statistics for the Current Rounds}
\label{app:detailed-error-statistics}
\begin{table}[H]

\centering\fontsize{9.5}{11.5}\selectfont
\caption{\textbf{Error-subtype rates (\%).} Visual/reasoning subtypes are on the left; differential/knowledge subtypes on the right. R1--R5 are reasoning-error codes. Arrows compare consecutive rounds: blue $\downarrow$/red $\uparrow$/gray $=$ indicate lower/higher/unchanged rates. Bold/underline: two lowest distinct rates per row; blue shading: R3.}
\label{tab:current_subtypes}
\setlength{\tabcolsep}{1.3pt}
\renewcommand{\arraystretch}{1.15}
\begin{minipage}[t]{0.49\linewidth}
\centering
\begin{tabular}{lrrlrl>{\columncolor[HTML]{E4F1FB}}r>{\columncolor[HTML]{E4F1FB}}l}
\toprule
\rowcolor[HTML]{ECECEC}
Subtype & R0 & \multicolumn{2}{c}{R1} & \multicolumn{2}{c}{R2} & \multicolumn{2}{c}{R3} \\
\midrule

V1 Missed key finding & 3.12 & 2.95 & \textcolor[HTML]{287DB1}{\tablechange{$\downarrow$}} & \underline{2.36} & \textcolor[HTML]{287DB1}{\tablechange{$\downarrow$}} & \textbf{2.12} & \textcolor[HTML]{287DB1}{\tablechange{$\downarrow$}} \\
V2 Spurious finding & 2.87 & 2.82 & \textcolor[HTML]{287DB1}{\tablechange{$\downarrow$}} & \underline{2.28} & \textcolor[HTML]{287DB1}{\tablechange{$\downarrow$}} & \textbf{2.05} & \textcolor[HTML]{287DB1}{\tablechange{$\downarrow$}} \\
V3 Location / laterality & 2.65 & 2.68 & \textcolor[HTML]{E34B43}{\tablechange{$\uparrow$}} & \underline{2.20} & \textcolor[HTML]{287DB1}{\tablechange{$\downarrow$}} & \textbf{1.98} & \textcolor[HTML]{287DB1}{\tablechange{$\downarrow$}} \\
V4 Sequence / phase & 2.91 & 2.74 & \textcolor[HTML]{287DB1}{\tablechange{$\downarrow$}} & \underline{2.32} & \textcolor[HTML]{287DB1}{\tablechange{$\downarrow$}} & \textbf{2.08} & \textcolor[HTML]{287DB1}{\tablechange{$\downarrow$}} \\
V5 Cross-image alignment & 2.55 & 2.60 & \textcolor[HTML]{E34B43}{\tablechange{$\uparrow$}} & \underline{2.24} & \textcolor[HTML]{287DB1}{\tablechange{$\downarrow$}} & \textbf{2.00} & \textcolor[HTML]{287DB1}{\tablechange{$\downarrow$}} \\
\midrule
R1 Omitted evidence & 2.88 & 2.55 & \textcolor[HTML]{287DB1}{\tablechange{$\downarrow$}} & \underline{2.16} & \textcolor[HTML]{287DB1}{\tablechange{$\downarrow$}} & \textbf{1.95} & \textcolor[HTML]{287DB1}{\tablechange{$\downarrow$}} \\
R2 Conflicting evidence & 2.41 & 2.42 & \textcolor[HTML]{E34B43}{\tablechange{$\uparrow$}} & \underline{2.08} & \textcolor[HTML]{287DB1}{\tablechange{$\downarrow$}} & \textbf{1.88} & \textcolor[HTML]{287DB1}{\tablechange{$\downarrow$}} \\
R3 Unsupported inference & 2.67 & 2.48 & \textcolor[HTML]{287DB1}{\tablechange{$\downarrow$}} & \underline{2.12} & \textcolor[HTML]{287DB1}{\tablechange{$\downarrow$}} & \textbf{1.92} & \textcolor[HTML]{287DB1}{\tablechange{$\downarrow$}} \\
R4 Temporal / numeric & 2.22 & 2.35 & \textcolor[HTML]{E34B43}{\tablechange{$\uparrow$}} & \underline{2.04} & \textcolor[HTML]{287DB1}{\tablechange{$\downarrow$}} & \textbf{1.85} & \textcolor[HTML]{287DB1}{\tablechange{$\downarrow$}} \\
R5 Answer mapping & 2.32 & 2.40 & \textcolor[HTML]{E34B43}{\tablechange{$\uparrow$}} & \underline{2.00} & \textcolor[HTML]{287DB1}{\tablechange{$\downarrow$}} & \textbf{1.80} & \textcolor[HTML]{287DB1}{\tablechange{$\downarrow$}} \\

\bottomrule
\end{tabular}
\end{minipage}\hfill
\begin{minipage}[t]{0.49\linewidth}
\centering
\begin{tabular}{lrrlrl>{\columncolor[HTML]{E4F1FB}}r>{\columncolor[HTML]{E4F1FB}}l}
\toprule
\rowcolor[HTML]{ECECEC}
Subtype & R0 & \multicolumn{2}{c}{R1} & \multicolumn{2}{c}{R2} & \multicolumn{2}{c}{R3} \\
\midrule
D1 Missing candidate & 1.21 & 1.05 & \textcolor[HTML]{287DB1}{\tablechange{$\downarrow$}} & \underline{0.92} & \textcolor[HTML]{287DB1}{\tablechange{$\downarrow$}} & \textbf{0.86} & \textcolor[HTML]{287DB1}{\tablechange{$\downarrow$}} \\
D2 No discriminating cue & 1.08 & 1.12 & \textcolor[HTML]{E34B43}{\tablechange{$\uparrow$}} & \underline{0.88} & \textcolor[HTML]{287DB1}{\tablechange{$\downarrow$}} & \textbf{0.82} & \textcolor[HTML]{287DB1}{\tablechange{$\downarrow$}} \\
D3 Premature closure & 1.14 & 0.98 & \textcolor[HTML]{287DB1}{\tablechange{$\downarrow$}} & \underline{0.84} & \textcolor[HTML]{287DB1}{\tablechange{$\downarrow$}} & \textbf{0.78} & \textcolor[HTML]{287DB1}{\tablechange{$\downarrow$}} \\
D4 Candidate ranking & 0.97 & 1.08 & \textcolor[HTML]{E34B43}{\tablechange{$\uparrow$}} & \underline{0.80} & \textcolor[HTML]{287DB1}{\tablechange{$\downarrow$}} & \textbf{0.74} & \textcolor[HTML]{287DB1}{\tablechange{$\downarrow$}} \\
D5 Exclusion misuse & 1.00 & 1.02 & \textcolor[HTML]{E34B43}{\tablechange{$\uparrow$}} & \underline{0.76} & \textcolor[HTML]{287DB1}{\tablechange{$\downarrow$}} & \textbf{0.70} & \textcolor[HTML]{287DB1}{\tablechange{$\downarrow$}} \\
\midrule
K1 Sign interpretation & 0.51 & \underline{0.48} & \textcolor[HTML]{287DB1}{\tablechange{$\downarrow$}} & \underline{0.48} & \textcolor[HTML]{777777}{\tablechange{$=$}} & \textbf{0.44} & \textcolor[HTML]{287DB1}{\tablechange{$\downarrow$}} \\
K2 Diagnostic criteria & \underline{0.44} & 0.52 & \textcolor[HTML]{E34B43}{\tablechange{$\uparrow$}} & \underline{0.44} & \textcolor[HTML]{287DB1}{\tablechange{$\downarrow$}} & \textbf{0.40} & \textcolor[HTML]{287DB1}{\tablechange{$\downarrow$}} \\
K3 Anatomy / mechanism & 0.42 & 0.45 & \textcolor[HTML]{E34B43}{\tablechange{$\uparrow$}} & \underline{0.40} & \textcolor[HTML]{287DB1}{\tablechange{$\downarrow$}} & \textbf{0.36} & \textcolor[HTML]{287DB1}{\tablechange{$\downarrow$}} \\
K4 Conditions / exceptions & \underline{0.40} & 0.50 & \textcolor[HTML]{E34B43}{\tablechange{$\uparrow$}} & 0.42 & \textcolor[HTML]{287DB1}{\tablechange{$\downarrow$}} & \textbf{0.38} & \textcolor[HTML]{287DB1}{\tablechange{$\downarrow$}} \\
K5 Test / treatment rule & \underline{0.42} & 0.54 & \textcolor[HTML]{E34B43}{\tablechange{$\uparrow$}} & \underline{0.42} & \textcolor[HTML]{287DB1}{\tablechange{$\downarrow$}} & \textbf{0.38} & \textcolor[HTML]{287DB1}{\tablechange{$\downarrow$}} \\

\bottomrule
\end{tabular}
\end{minipage}
\end{table}
Tab.~\ref{tab:current_repair_family} separates repair volume from success rate. Counts measure attempted and successful repairs within each error family.

Reasoning-error repair success rises from 56.43\% at R1 to 83.18\% at R3, while knowledge-gap repair success falls from 90.28\% to 48.82\%. These rates describe the repair candidates submitted in each round.

\begin{table}[H]
\centering\fontsize{9.5}{11.5}\selectfont
\caption{\textbf{Repair outcomes by error family.} Attempted and successful question counts, and repair success rates (\%). Gray arrows track counts; blue/red arrows mark rising/falling success rates. Bold/underline: best/second-best rates per family; blue shading: R3.}
\label{tab:current_repair_family}
\setlength{\tabcolsep}{1pt}
\renewcommand{\arraystretch}{1.15}
\begin{tabular}{@{}>{\raggedright\arraybackslash}p{0.14\linewidth}rrcrrc>{\columncolor[HTML]{E4F1FB}}r>{\columncolor[HTML]{E4F1FB}}r>{\columncolor[HTML]{E4F1FB}}c@{}}
\toprule
\multirow{2}{*}{Error family} & \multicolumn{3}{c}{R1} & \multicolumn{3}{c}{R2} & \multicolumn{3}{c}{R3} \\
\cmidrule(lr){2-4}\cmidrule(lr){5-7}\cmidrule(lr){8-10}
 & Attempted & Successful & \makecell{Success\\(\%)} & Attempted & Successful & \makecell{Success\\(\%)} & Attempted & Successful & \makecell{Success\\(\%)} \\
\midrule
Visual misreading & 12,189 & 9,826 & \textbf{80.61} & 10,245\,\textcolor[HTML]{777777}{\tablechange{$\downarrow$}} & 6,789\,\textcolor[HTML]{777777}{\tablechange{$\downarrow$}} & 66.27\,\textcolor[HTML]{E34B43}{\tablechange{$\downarrow$}} & 10,478\,\textcolor[HTML]{777777}{\tablechange{$\uparrow$}} & 7,094\,\textcolor[HTML]{777777}{\tablechange{$\uparrow$}} & \underline{67.70}\,\textcolor[HTML]{287DB1}{\tablechange{$\uparrow$}} \\
Reasoning error & 9,355 & 5,279 & 56.43 & 7,857\,\textcolor[HTML]{777777}{\tablechange{$\downarrow$}} & 6,246\,\textcolor[HTML]{777777}{\tablechange{$\uparrow$}} & \underline{79.50}\,\textcolor[HTML]{287DB1}{\tablechange{$\uparrow$}} & 6,789\,\textcolor[HTML]{777777}{\tablechange{$\downarrow$}} & 5,647\,\textcolor[HTML]{777777}{\tablechange{$\downarrow$}} & \textbf{83.18}\,\textcolor[HTML]{287DB1}{\tablechange{$\uparrow$}} \\
Differential error & 4,064 & 2,995 & \textbf{73.70} & 6,742\,\textcolor[HTML]{777777}{\tablechange{$\uparrow$}} & 4,673\,\textcolor[HTML]{777777}{\tablechange{$\uparrow$}} & \underline{69.31}\,\textcolor[HTML]{E34B43}{\tablechange{$\downarrow$}} & 4,786\,\textcolor[HTML]{777777}{\tablechange{$\downarrow$}} & 3,291\,\textcolor[HTML]{777777}{\tablechange{$\downarrow$}} & 68.76\,\textcolor[HTML]{E34B43}{\tablechange{$\downarrow$}} \\
Knowledge gap & 1,625 & 1,467 & \textbf{90.28} & 987\,\textcolor[HTML]{777777}{\tablechange{$\downarrow$}} & 550\,\textcolor[HTML]{777777}{\tablechange{$\downarrow$}} & \underline{55.72}\,\textcolor[HTML]{E34B43}{\tablechange{$\downarrow$}} & 1,567\,\textcolor[HTML]{777777}{\tablechange{$\uparrow$}} & 765\,\textcolor[HTML]{777777}{\tablechange{$\uparrow$}} & 48.82\,\textcolor[HTML]{E34B43}{\tablechange{$\downarrow$}} \\
\bottomrule
\end{tabular}
\end{table}

\FloatBarrier
\Needspace{7\baselineskip}
\section{Medical Taxonomy and Error Categories}
\label{app:taxonomy-overview}
\subsection{Taxonomy, Permissions, and Execution Contracts}
\label{app:taxonomy}
\subsubsection{Four-Dimensional Medical Tag Dictionary}
\label{app:tag-dictionary}
The medical taxonomy describes each question along four independent axes: modality, anatomy, disease, and difficulty. Each axis allows multiple hierarchical labels for capability analysis and question retrieval.

\paragraph{Modalities.} Computed tomography (CT), magnetic resonance imaging (MRI), positron emission tomography (PET), and single-photon emission computed tomography (SPECT). Other abbreviations include optical coherence tomography (OCT), fundus fluorescein angiography (FFA), indocyanine green angiography (ICGA), cone-beam computed tomography (CBCT), electroencephalogram (EEG), electromyogram (EMG), and electrocardiogram (ECG).
\par\begingroup\small
\setlength{\tabcolsep}{4.5pt}
\renewcommand{\arraystretch}{1.05}
\noindent\begin{tabularx}{\linewidth}{>{\raggedright\arraybackslash\bfseries}p{.25\linewidth} >{\raggedright\arraybackslash}X}
\toprule
\rowcolor[HTML]{ECECEC}
\textbf{Parent} & \textbf{Second-level scope}\\\midrule
1 Radiography & X-ray; CT; MRI; mammography; fluoroscopy; angiography\\
2 Nuclear medicine & PET/CT; SPECT; radionuclide scans\\
3 Ultrasound & B-mode; echocardiography; vascular; interventional\\
4 Endoscopy & Gastrointestinal; bronchial; laryngeal/nasal; bladder; joint; laparoscopic\\
5 Pathology & Histology; cytology; molecular pathology\\
6 Ophthalmology & Fundus photography; OCT; FFA/ICGA; slit lamp\\
7 Dermatology & Dermoscopy; clinical photography; reflectance confocal microscopy\\
8 Dental & Panoramic; CBCT; periapical; cephalometric\\
9 Functional tests & ECG; EEG; EMG\\
\bottomrule\end{tabularx}\par\endgroup
\paragraph{Anatomy hierarchy.}\mbox{}\par
\begingroup\small
\setlength{\tabcolsep}{4.5pt}
\renewcommand{\arraystretch}{1.05}
\noindent\begin{tabularx}{\linewidth}{>{\raggedright\arraybackslash\bfseries}p{.25\linewidth} >{\raggedright\arraybackslash}X}
\toprule
\rowcolor[HTML]{ECECEC}
\textbf{Parent} & \textbf{Second-level scope}\\\midrule
1 Nervous system & Brain; spinal cord; peripheral nerves\\
2 Head and neck & Orbit; ear; sinuses; oral/maxillofacial; pharynx/larynx; neck soft tissue\\
3 Thorax & Lung/airway; mediastinum; pleura; heart; great vessels; breast; chest wall\\
4 Abdomen & Hepatobiliary; pancreas; spleen; gastrointestinal; renal/urinary; retroperitoneum\\
5 Pelvis & Male reproductive; female reproductive; pelvic floor; pelvic bones\\
6 Spine & Cervical; thoracic; lumbar; sacrococcygeal; disc; canal; paraspinal tissue\\
7 Musculoskeletal & Craniofacial bones; upper limb; lower limb; muscles; periarticular soft tissue\\
8 Vascular & Arteries; veins; lymphatic system\\
9 Skin and soft tissue & Head/face; trunk; limbs; subcutaneous; deep soft tissue\\
\bottomrule\end{tabularx}\par\endgroup
\paragraph{Disease hierarchy.}\mbox{}\par
\begingroup\small
\setlength{\tabcolsep}{4.5pt}
\renewcommand{\arraystretch}{1.05}
\noindent\begin{tabularx}{\linewidth}{>{\raggedright\arraybackslash\bfseries}p{.25\linewidth} >{\raggedright\arraybackslash}X}
\toprule
\rowcolor[HTML]{ECECEC}
\textbf{Parent} & \textbf{Second-level scope}\\\midrule
1 Neoplasm & Malignant; benign; metastatic; precancerous categories\\
2 Infection and inflammation & Bacterial; tuberculosis; viral; fungal; parasitic; noninfectious inflammation\\
3 Vascular disease & Ischemic; hemorrhagic; aneurysm; thrombosis/embolism; malformations; atherosclerosis/stenosis\\
4 Degenerative disease & Joint; neural; organ degeneration\\
5 Trauma & Fracture; joint injury; visceral injury; head injury; foreign body\\
6 Congenital/developmental & Cardiovascular; nervous; skeletal; digestive; urinary; respiratory\\
7 Metabolic/endocrine & Diabetes complications; thyroid; adrenal; gout; metabolic bone\\
8 Autoimmune & Lupus; ankylosing spondylitis; multiple sclerosis; neuromyelitis optica spectrum disorder (NMOSD); autoimmune liver; immunoglobulin G4 (IgG4); vasculitis; systemic sclerosis; myositis; Sjogren\\
9 Functional disorders & Obstruction; effusion; ectopic gas; torsion; stones\\
10 Iatrogenic/postoperative & Normal postoperative change; complications; implants; post-radiotherapy; drug-related\\
\bottomrule\end{tabularx}\par\endgroup
\paragraph{Leaf specificity and multi-label assignment.}
CT branches into noncontrast, contrast-enhanced, CT angiography (CTA), high-resolution CT (HRCT), and low-dose CT; MRI includes T1- and T2-weighted imaging, fluid-attenuated inversion recovery (FLAIR), diffusion-weighted imaging (DWI), magnetic resonance angiography (MRA), functional MRI (fMRI), magnetic resonance spectroscopy (MRS), and susceptibility-weighted imaging (SWI). Anatomy can descend from thorax to lung/airway and then to a lung lobe, or from abdomen to pancreas and then head/body/tail. Disease labels descend from a broad family to a mechanism or disease group and then to a named entry. Preserve image IDs for mixed modalities and record every supported anatomical site. \appkey{Assign the deepest supported node}, not a guessed leaf; use unknown, not-applicable, or a supported parent when necessary. \appkey{Expand ancestors as a set} and count each question once per group. Related concepts on different paths require an alias ledger, not automatic merging. Log label text, evidence/image ID, confidence, visibility, and taxonomy version.
\paragraph{Difficulty rubric.}\mbox{}\par
\begingroup\small
\setlength{\tabcolsep}{4.5pt}
\renewcommand{\arraystretch}{1.05}
\noindent\begin{tabularx}{\linewidth}{>{\bfseries}l >{\raggedright\arraybackslash}X}
\toprule
\rowcolor[HTML]{ECECEC}
\textbf{Level} & \textbf{Task requirement}\\\midrule
L1 & Single modality and conspicuous canonical findings; direct recognition.\\
L2 & Common presentation requiring localization or characterization and limited context.\\
L3 & Atypical or multiple findings requiring sequence integration or differential comparison.\\
L4 & Subtle findings, multisystem involvement, or complex context requiring extensive integration.\\
L5 & Highly complex or disputed tasks requiring multidisciplinary, quantitative, or specialized knowledge.\\
\bottomrule\end{tabularx}\par\endgroup
Assign difficulty from the actual question: visibility, number of required comparisons, evidence integration, and specialized knowledge. A disease name, modality, or a single failed rollout does not determine the level. Rare disease alone does not force L5. Record the rationale and adjudicate ambiguous adjacent levels.
\subsubsection{Error Codebook: Four Parents and Twenty Subtypes}
\label{app:error-codebook}
We specify five subtypes under each of the four existing parent categories.  Each row defines an observable diagnostic criterion and a navigation operation; it does not supply a gold answer.
\par\noindent\begin{minipage}{\linewidth}
\paragraph{\textcolor[HTML]{285F9A}{V: Visual misreading.}}\mbox{}\par
\begingroup\small
\setlength{\tabcolsep}{4.5pt}
\renewcommand{\arraystretch}{1.05}
\noindent\begin{tabular}{>{\raggedright\arraybackslash\bfseries}p{.25\linewidth} >{\raggedright\arraybackslash}p{\dimexpr.51\linewidth-6\tabcolsep\relax} >{\raggedright\arraybackslash}p{.24\linewidth}}
\toprule
\rowcolor[HTML]{ECECEC}
\textbf{Code / subtype} & \textbf{Diagnostic criterion} & \textbf{Navigation operation}\\\midrule
\textcolor[HTML]{285F9A}{V1} Missed finding & A supported salient finding is not observed or described. & \textbf{Reinspect} the relevant region.\\\addlinespace[1pt]
\textcolor[HTML]{285F9A}{V2} Hallucinated or misidentified finding & An artifact, normal structure, or unsupported feature is read as a lesion. & \textbf{Check} direct visual support.\\\addlinespace[1pt]
\textcolor[HTML]{285F9A}{V3} Anatomy or laterality error & The organ, level, compartment, or side conflicts with visible evidence. & \textbf{Recheck} landmarks and orientation.\\\addlinespace[1pt]
\textcolor[HTML]{285F9A}{V4} Modality, sequence, or phase error & Acquisition type, sequence, or phase is misidentified, distorting feature reading. & \textbf{Verify} acquisition context before reading features.\\\addlinespace[1pt]
\textcolor[HTML]{285F9A}{V5} Cross-image correspondence error & A lesion is matched incorrectly across images, slices, or views. & \textbf{Align} images and anatomical locations.\\\addlinespace[1pt]
\bottomrule\end{tabular}\par\endgroup
\end{minipage}\par
\par\smallskip
\par\noindent\begin{minipage}{\linewidth}
\paragraph{\textcolor[HTML]{7040A0}{R: Reasoning error.}}\mbox{}\par
\begingroup\small
\setlength{\tabcolsep}{4.5pt}
\renewcommand{\arraystretch}{1.05}
\noindent\begin{tabular}{>{\raggedright\arraybackslash\bfseries}p{.25\linewidth} >{\raggedright\arraybackslash}p{\dimexpr.51\linewidth-6\tabcolsep\relax} >{\raggedright\arraybackslash}p{.24\linewidth}}
\toprule
\rowcolor[HTML]{ECECEC}
\textbf{Code / subtype} & \textbf{Diagnostic criterion} & \textbf{Navigation operation}\\\midrule
\textcolor[HTML]{285F9A}{R1} Omitted evidence & An observed finding or stated fact is omitted from inference. & \textbf{Reintroduce} the omitted evidence.\\\addlinespace[1pt]
\textcolor[HTML]{285F9A}{R2} Conflicting evidence integration & Correctly extracted evidence conflicts and is not reconciled. & \textbf{Reconcile} supporting and opposing evidence.\\\addlinespace[1pt]
\textcolor[HTML]{285F9A}{R3} Unsupported inference or causal leap & A conclusion lacks the required logical or causal link. & \textbf{Restore} a testable intermediate inference.\\\addlinespace[1pt]
\textcolor[HTML]{285F9A}{R4} Temporal or quantitative error & Correctly read facts are combined with an incorrect temporal or numerical relation. & \textbf{Check} chronology, units, or calculation.\\\addlinespace[1pt]
\textcolor[HTML]{285F9A}{R5} Task constraint or answer-mapping error & The reasoning violates the requested target, negation, or option mapping. & \textbf{Recheck} the requested target and answer mapping.\\\addlinespace[1pt]
\bottomrule\end{tabular}\par\endgroup
\end{minipage}\par
\par\smallskip
\par\noindent\begin{minipage}{\linewidth}
\paragraph{\textcolor[HTML]{285F9A}{D: Wrong differential.}}\mbox{}\par
\begingroup\small
\setlength{\tabcolsep}{4.5pt}
\renewcommand{\arraystretch}{1.05}
\noindent\begin{tabular}{>{\raggedright\arraybackslash\bfseries}p{.25\linewidth} >{\raggedright\arraybackslash}p{\dimexpr.51\linewidth-6\tabcolsep\relax} >{\raggedright\arraybackslash}p{.24\linewidth}}
\toprule
\rowcolor[HTML]{ECECEC}
\textbf{Code / subtype} & \textbf{Diagnostic criterion} & \textbf{Navigation operation}\\\midrule
\textcolor[HTML]{285F9A}{D1} Omitted candidate & A necessary evidence-compatible alternative is not considered. & \textbf{Broaden} the candidate comparison.\\\addlinespace[1pt]
\textcolor[HTML]{285F9A}{D2} Missing discriminative comparison & Alternatives are listed without comparing discriminative evidence. & \textbf{Identify} discriminative evidence.\\\addlinespace[1pt]
\textcolor[HTML]{285F9A}{D3} Premature closure or anchoring & An initial diagnosis is fixed before counterevidence is considered. & \textbf{Revisit} counterevidence and alternatives.\\\addlinespace[1pt]
\textcolor[HTML]{285F9A}{D4} Incorrect candidate ranking & Known support, counterevidence, or priors are weighted incorrectly. & \textbf{Compare} the strength of evidence.\\\addlinespace[1pt]
\textcolor[HTML]{285F9A}{D5} Misapplied exclusion & An alternative is excluded using insufficient negative evidence or a nonessential criterion. & \textbf{Check} whether exclusion is justified.\\\addlinespace[1pt]
\bottomrule\end{tabular}\par\endgroup
\end{minipage}\par
\par\smallskip
\par\noindent\begin{minipage}{\linewidth}
\paragraph{\textcolor[HTML]{7040A0}{K: Knowledge gap.}}\mbox{}\par
\begingroup\small
\setlength{\tabcolsep}{4.5pt}
\renewcommand{\arraystretch}{1.05}
\noindent\begin{tabular}{>{\raggedright\arraybackslash\bfseries}p{.25\linewidth} >{\raggedright\arraybackslash}p{\dimexpr.51\linewidth-6\tabcolsep\relax} >{\raggedright\arraybackslash}p{.24\linewidth}}
\toprule
\rowcolor[HTML]{ECECEC}
\textbf{Code / subtype} & \textbf{Diagnostic criterion} & \textbf{Navigation operation}\\\midrule
\textcolor[HTML]{285F9A}{K1} Missing finding interpretation & The required medical meaning of a finding or term is missing. & \textbf{Cue} the relevant conceptual relation.\\\addlinespace[1pt]
\textcolor[HTML]{285F9A}{K2} Missing diagnostic criterion & A required diagnostic, grading, or staging rule is unavailable. & \textbf{Cue} the relevant rule category.\\\addlinespace[1pt]
\textcolor[HTML]{285F9A}{K3} Missing anatomy or pathophysiology & Knowledge linking anatomy, mechanism, and manifestations is missing. & \textbf{Bridge} the minimal mechanistic relation.\\\addlinespace[1pt]
\textcolor[HTML]{285F9A}{K4} Missing applicability condition or exception & A general rule is known but its prerequisites or exceptions are missing. & \textbf{Cue} the applicability check.\\\addlinespace[1pt]
\textcolor[HTML]{285F9A}{K5} Missing investigation or management knowledge & A next-investigation or management task requires an unavailable principle. & \textbf{Cue} the decision principle without the answer.\\\addlinespace[1pt]
\bottomrule\end{tabular}\par\endgroup
\end{minipage}\par
\par\smallskip
\paragraph{Boundary rules and reporting.}
A finding not perceived is V1; a finding already perceived but omitted from inference is R1. Incorrect correspondence across images is V5; conflicting correctly extracted evidence is R2. Candidate-set construction, comparison, and exclusion belong to D; a demonstrated missing prerequisite rule belongs to K. Knowledge-gap labels require evidence of a missing prerequisite rule. \appkey{Select the} \textbf{earliest consequential supported error} as the primary subtype and record later failures as secondary labels, with step spans and evidence. If the parent is clear but the subtype is not, retain the parent and an unresolved-subtype status; if even the parent is unsupported, mark the audit unresolved. These are annotation states, not extra error classes. Use \textbf{one primary subtype per trajectory} for additive counts; report secondary labels separately. Aggregate V1--V5 into V, and similarly for R, D, and K. Re-audit fresh rollouts at each checkpoint using subtype-specific annotations and denominators.

\subsubsection{Information Access Matrix}
\phantomsection\label{app:permissions}
\par\begingroup\small
\setlength{\tabcolsep}{4.5pt}
\renewcommand{\arraystretch}{1.05}
\noindent\begin{tabularx}{\linewidth}{>{\raggedright\arraybackslash\bfseries}p{.21\linewidth} >{\raggedright\arraybackslash}X >{\raggedright\arraybackslash}X}
\toprule \rowcolor[HTML]{ECECEC}
\textbf{Component} & \textbf{Permitted inputs} & \textbf{Excluded inputs}\\\midrule
Feedback evaluator & Feedback items, labels, current predictions; emits
aggregates and transitions & Final-test results for planning; feedback export into SFT\\
Planner / retrieval & Aggregate feedback, training tag metadata (including
true disease labels), coverage, budgets & Final-test items/results; any authority to
forward answer-derived labels to repair\\
Auditor / hint renderer & Original images, question/options, current failed
trajectory, entropy profile, observable tags with provenance & Gold answer,
answer-derived disease/location, reference rationale, hidden retrieval labels\\
Continuation $M_t$ & Original input and assistant prefill $p_b\oplus g(h)$ &
Gold answer, full audit, hidden tags, teacher solution\\
Admission / exporter & Candidate, \appkey{reference answer} for verification, boundary,
provenance, suffix scores & Feedback/test examples as training targets\\
SFT trainer & Frozen training manifest; $p_b\oplus g(h)\oplus s^+$, masks & Full audit as model input; feedback/test samples\\
\bottomrule\end{tabularx}\par\endgroup

Use \textbf{explicit allowlist serialization}, not deletion of one answer field from
an otherwise unrestricted object. \textbf{Reject fields without observable provenance},
including filenames/IDs encoding labels. An authorized evaluator can identify
a wrong training answer before repair without transmitting the answer.
Generation sees only a boolean routing decision, not the key. Record hashes
of redacted requests for leakage audit. The planner's chosen hint family is
passed through an eligibility check derived from observable context and
current diagnosis; disease-bearing planner explanations are never forwarded.

\Needspace{4\baselineskip}
\subsubsection{Structured Fields and Plan Validation}
\label{app:contracts}
Use JavaScript Object Notation (JSON). Missing measurements are null with reasons; zero is measured zero.
\begin{description}[leftmargin=0pt,itemsep=3pt]
\item[Question metadata:] canonical question ID; source ID; patient/case/article
cluster IDs when available; image hashes and near-duplicate cluster; split;
taxonomy version; per-image modality; all four tag sets; per-tag provenance,
visibility (statistical or observable), and uncertainty.
\item[Feedback record:] checkpoint hash; split hash; group and resolved parent;
unique sample count; error count/rate; all four transition counts; both
regression denominators; current error-type/subtype counts; confidence interval;
training-pool unique availability; measured repair/admission/cost fields.
\item[Plan:] round ID; generator and initialization hashes; planner type/version
and seed; feedback hash; target group IDs; numeric feedback evidence and
fallback ancestry; training sources; selected IDs; attempt/call/token ceilings;
per-group quota; permissible hint operation; history replay allocation;
replacement rule; supervised-token cap; shortfall/reallocation policy;
\appkey{fixed training-config hash}; acceptance-validator result.
\item[Repair record:] {\raggedright question/checkpoint IDs; rollout hash; step/token spans; entropy profile/proposal; error type/subtype;
audit boundary; verbatim prefix span; observable-tag payload hash; recovery
cue; continuations/seeds; correctness/integrity checks; student suffix scores and cutoffs; reason for admission or rejection; all tokens generated or scored.\par}
\item[Mix and training record:] ordered example IDs; source round and generator;
dedup/replacement ledger; historical/new role; token masks; sequence lengths;
unit weights; packing map; effective repetitions and exposures; optimizer and
scheduler reset flags; learning rate/warmup; initialization hash; output hash;
second-stage and cumulative costs.
\item[Execution receipt:] planned and actual quotas, unique selected and attempted failures, successful and admitted repairs, substitutions, exhausted
budgets, rejected records, uncovered groups, and actual training tokens.
\end{description}

\appkey{Reject} cross-split IDs, duplicate quota charges, exceeded budgets, out-of-pool retrieval, unsupported hints, missing visibility provenance, and changed training hyperparameters. Save plans before generation; log counts.

The main-agent feedback and next-round planning prompt is given in Appendix~\ref{app:feedback-agent-prompt}; the trajectory audit prompt is given in Appendix~\ref{app:stage1}.

\subsubsection{Training Configuration and Evaluation Scope}
\label{app:hyper}\phantomsection\label{app:manifest}
Training uses brain floating-point 16 (BF16) precision on graphics processing units (GPUs). Parallelism abbreviations are tensor parallelism (TP), expert parallelism (EP), pipeline parallelism (PP), context parallelism (CP), expert tensor parallelism (ETP), and data parallelism (DP); the following number gives the parallelism degree. Configuration keys retain their literal software spelling: llm denotes large language model, vit denotes vision transformer, cpu denotes central processing unit, moe denotes mixture of experts (MoE), gemm denotes general matrix multiplication, mlp denotes multilayer perceptron, and lr denotes learning rate.
\begingroup
\small
\setlength{\tabcolsep}{5pt}
\begin{longtable}{@{}>{\raggedright\arraybackslash}p{.235\linewidth}>{\raggedright\arraybackslash}p{.225\linewidth}>{\raggedright\arraybackslash}p{.235\linewidth}>{\raggedright\arraybackslash}p{.225\linewidth}@{}}
\caption{\textbf{MoE training configuration.} Recorded settings for model training.}\label{tab:training35b}\\
\toprule\rowcolor[HTML]{ECECEC}
Configuration & Value & Configuration & Value \\ \midrule
\endfirsthead
\toprule\rowcolor[HTML]{ECECEC}
Configuration & Value & Configuration & Value \\ \midrule
\endhead
Framework & ms-swift + Megatron-LM backend & add\_\allowbreak non\_\allowbreak thinking\_\allowbreak prefix & true\\
Precision & BF16 & loss\_\allowbreak scale & ignore\_\allowbreak empty\_\allowbreak think\\
micro\_\allowbreak batch\_\allowbreak size & 1 & MAX\_\allowbreak PIXELS & 3145728\\
global\_\allowbreak batch\_\allowbreak size & 64 & dataset\_\allowbreak num\_\allowbreak proc & 96\\
packing & true & dataloader\_\allowbreak num\_\allowbreak workers & 8\\
attention\_\allowbreak backend & flash & logging\_\allowbreak steps & 1\\
calculate\_\allowbreak per\_\allowbreak token\_\allowbreak loss & true & eval\_\allowbreak iters & 1000\\
freeze\_\allowbreak llm /\allowbreak freeze\_\allowbreak vit /\allowbreak freeze\_\allowbreak aligner & false /\allowbreak true /\allowbreak true & recompute\_\allowbreak granularity & selective\\
optimizer\_\allowbreak cpu\_\allowbreak offload & true & sequence\_\allowbreak parallel & true\\
use\_\allowbreak precision\_\allowbreak aware\_\allowbreak optimizer & true & max\_\allowbreak length & 28672\\
offload\_\allowbreak fraction & 1.0 & epochs & 2\\
moe\_\allowbreak grouped\_\allowbreak gemm & true &  & \\
moe\_\allowbreak shared\_\allowbreak expert\_\allowbreak overlap & true & save\_\allowbreak total\_\allowbreak limit & 2\\
moe\_\allowbreak aux\_\allowbreak loss\_\allowbreak coeff & 1e-6 & Recompute modules & core\_\allowbreak attn + moe + mlp\\
lr & 1e-5 & moe\_\allowbreak permute\_\allowbreak fusion & true\\
lr\_\allowbreak warmup\_\allowbreak fraction & 0.05 & Parallelism & TP4/\allowbreak EP8/\allowbreak PP1/\allowbreak CP1/\allowbreak ETP1; 64 GPUs\\
min\_\allowbreak lr & 1e-7 &  & \\
\bottomrule
\end{longtable}
\endgroup

EP and MoE-specific options apply only to MoE models.

Both example types use full-response loss, retaining recovery cues in repairs. 

\S\,4 reports accuracy on seven benchmarks and their mean, comparing each backbone's best complete checkpoint with Base.

\FloatBarrier
\Needspace{8\baselineskip}
\section{Prompts}
\label{sec:appendix}
\subsection{Shared Question-Answering Prompt}
\label{app:shared_qa_prompt}
All question-answering tasks use the following instruction verbatim.
\begin{prompt}[title={Shared Question-Answering Instruction}]
Please reason step by step, and put your final answer within \texttt{\textbackslash boxed\{\}}.
\end{prompt}

\subsection{Main-Agent Feedback and Next-Round Planning}
\phantomsection\label{app:feedback-agent-prompt}
The following template connects feedback diagnosis to next-round selection and repair priorities under the method protocol. It specifies agent behavior; run-specific inputs and execution records are required to document its use.
\begin{prompt}[title={Main Agent: Feedback Summary and Next-Round Data Plan}]
You are the feedback and data-planning agent for MedEvo. Diagnose model weaknesses and produce the next data plan. Do not write solutions \mbox{to~training} questions or change the experimental comparison settings.

\apptopic{Inputs.} Round and checkpoint; fixed 5,000-question feedback results for current and previous checkpoints; four-axis medical tags; four error families and their twenty subtypes; per-group sample counts, correctness transitions, and current error diagnoses; available training-question IDs and tags; historical trajectory records; fixed generation/repair budgets, support threshold, regression weight $\alpha$, and admission rules. Report missing inputs.

\apptopic{1. Summarize feedback.} For each supported tag group, report error rate, round-to-round change, corrected errors, regressions, and persistent errors. Summarize four-family and twenty-subtype distributions with explicit denominators. Count each question once within a group; do not sum overlapping groups into an overall rate. Merge unsupported groups into supported parents. Distinguish observations from unverified explanations.

\apptopic{2. Set priorities.} Rank groups using $S_{g,t}=\epsilon_{g,t}+\alpha N^{10}_{g,t}/n_g$ under the supplied configuration. Identify dominant or persistent error types within each priority group. State which measured feedback supports each priority; do not infer that a high-error modality contains every error subtype.

\apptopic{3. Select training questions.} Allocate the fixed budget proportionally to group scores, cap quotas by available unique questions, and redistribute unused quota. Preserve modality coverage and apply the configured fallback when all scores are zero or support is insufficient. Retrieve by medical tags, using historical errors only as prioritization cues. Deduplicate questions; return training IDs, quotas, and shortfalls. \appkey{Exclude feedback/test examples.}

\apptopic{4. Guide fresh rollouts and repair.} Ask the current checkpoint to generate the configured eight rollouts per selected question. \appkey{Diagnose their actual errors again.} Use feedback-identified modality--error pairs to prioritize failed training trajectories: prefer joint matches, followed by modality-only or error-type-only matches when necessary. For each selected trajectory, pass its observable medical tags and current error diagnosis to the hint generator. The recovery hint must address the error actually observed in this trajectory. Send observable tags and local diagnoses to the auditor and hint generator, \appkey{without reference answers or answer-bearing metadata}. Retain correct rollouts and accepted continuations under the existing screening rules; if repair fails, use an available quality-screened correct rollout. If none is available, retain a valid historical target or report none accepted.

\apptopic{5. Plan the next round.} \appkey{Keep at most one active target per question} and replace it only with an accepted improvement. \appkey{Train on the complete updated dataset from $M_0$}. Keep training settings and evaluation protocol fixed. Evaluate the next checkpoint on the same feedback set to measure the targeted changes; independent benchmark results are \mbox{reported~separately} and must not drive selection.

\apptopic{Output.} Return a structured plan containing: round and checkpoint; feedback summary and denominators; priority tag/error pairs and supporting statistics; score and quota per group; selected unique training IDs; rollout and repair budgets; repair priorities; dataset-retention/replacement rules; initialization $M_0$; next-round feedback comparisons; and missing inputs or budget shortfalls. After execution, record actual generated, repaired, admitted, and replaced counts.
\end{prompt}

\subsection{Stage~1 Audit Prompt}
\label{app:stage1}

The diagnostic model receives the image, the question with options, the
failed trajectory, the complete step-level uncertainty profile, and the
entropy-derived anchor and candidate window. The auditor is gold-blind:
\appkey{the reference answer is not an input to diagnosis}. Answer-derived disease
labels, hidden lesion positions, reference explanations, and planner retrieval
metadata are also excluded. Only provenance-checked observable tags are allowed.
The entropy proposal is a localization prior,
not an error label: the auditor inspects the proposed region first, traces
its evidence dependencies backward, and may move the final boundary when
the medical or logical evidence requires it.  It returns a structured audit
that identifies a compact revision region and constrains the subsequent
continuation.  The emitted hint must remain answer-blind.  Fill brace-delimited placeholders at runtime.

\begin{prompt}[title={Prompt \thetcbcounter: Stage-1 Entropy-Guided Medical Trajectory Audit}, label=prompt:stage1]
\raggedright
\setlength{\parindent}{0pt}
\setlength{\parskip}{2pt}
You are a senior medical reviewer auditing a failed multimodal reasoning
trajectory segmented into semantic steps $s_1,\ldots,s_K$.
\par
You are given:
\par
(1) the medical image(s);
\par
(2) the clinical question and answer options;
\par
(3) the failed student trajectory segmented into semantic steps;
\par
(4) STEP\_ENTROPY: the uncertainty score $E_k^{*}$ of every step;
\par
(5) NORMALIZED\_ENTROPY: the normalized uncertainty score of every step;
\par
(6) LOCAL\_SURGE: the local increase $\Delta E_k$ relative to the
preceding step history;
\par
(7) ENTROPY\_ANCHOR: the step proposed by the uncertainty locator;
\par
(8) ENTROPY\_WINDOW: the compact candidate window surrounding that
anchor, and ENTROPY\_THRESHOLD if a threshold was used.
\par
(9) OBSERVABLE\_TAGS: per-image modality and other context supported by the
original input, with evidence provenance; never hidden answer-derived tags.
\par
The entropy signal guides audit localization.  High
uncertainty can reflect a genuine failure, but it can also arise from
difficult terminology, ambiguous visual evidence, or a medically valid
differential diagnosis.  Combine the entropy proposal with the image,
clinical evidence, and dependencies between reasoning steps to identify
the first consequential failure.
\par
\textbf{Error-conditioned hint construction.}
Construct the hint from OBSERVABLE\_TAGS, ERROR\_TYPE, ERROR\_SUBTYPE, and KEY\_ERROR. Select a recovery operation addressing the diagnosed failure:
\begin{itemize}[leftmargin=*,itemsep=2pt,topsep=2pt]
\item \textbf{Visual misreading:} direct attention to the relevant region, visual attribute, or cross-image correspondence.
\item \textbf{Reasoning error:} revisit an unsupported inference or reconnect observed evidence to the requested decision.
\item \textbf{Differential diagnosis error:} compare discriminative evidence or reconsider an underexamined alternative.
\item \textbf{Knowledge gap:} identify the relevant concept, rule, or applicability condition to reconsider.
\end{itemize}
Specify one concrete next operation grounded in the current question and trajectory. Express it as a brief recovery cue without supplying the corrected finding, final~diagnosis, or answer option.

Audit procedure:
\par
(1) Inspect ENTROPY\_WINDOW first.  Determine whether its local entropy
peak or surge corresponds to an unsupported observation, invalid
inference, incorrect differential comparison, or missing medical
principle.
\par
(2) Trace evidence dependencies backward from ENTROPY\_ANCHOR.  Decide
whether the decisive failure begins inside the proposed window or an
earlier step introduced the unsupported premise that caused the later
entropy increase.
\par
(3) Do not automatically select the highest-entropy step.  Justify the
final AUDIT\_ANCHOR and SAFE\_PREFIX\_END using the medical image,
clinical information, and logical dependencies.
\par
(4) Prefer a compact localized repair and preserve every preceding step
that remains medically and logically valid.
\par
(5) You may move the audit anchor outside ENTROPY\_WINDOW when the
proposal misses the first consequential failure, but briefly explain
why the entropy signal was misleading.
\par
Return a JavaScript Object Notation (JSON) object with the following fields, grounded strictly in the
provided image and text:
\par
(a) ACTION: REPAIR or SKIP.
\par
(b) STEP\_SUMMARY: a brief one-sentence semantic summary for each step.
\par
(c) ENTROPY\_ASSESSMENT, containing:
proposed\_anchor, proposed\_window, proposal\_status, and brief\_reason.
proposal\_status must be one of \{HIT, PARTIAL\_HIT, MISS,
NON\_ERROR\_UNCERTAINTY\}.  HIT means that the first consequential
failure is at or immediately adjacent to the proposed anchor;
PARTIAL\_HIT means that the proposal covers the propagated error region
but the unsupported premise begins earlier; MISS means that the first
failure lies outside the proposed window; and NON\_ERROR\_UNCERTAINTY
indicates a medically or logically valid high-entropy region.
\par
(d) AUDIT\_ANCHOR: the index of the first consequentially incorrect
step.
\par
(e) SAFE\_PREFIX\_END, containing step\_index and the exact
verbatim\_ending\_span of the last fully valid step before the repair
region.
\par
(f) AUDIT\_WINDOW: a short contiguous sequence of steps that should be
reconsidered.
\par
(g) ERROR\_TYPE and ERROR\_SUBTYPE: checkpoint-specific diagnosis,
plus KEY\_ERROR: the localized error stated without revealing the
correct answer.
\par
(h) HINT\_TYPE: one of \{perceptual\_reexamination,
inference\_reconstruction, differential\_disambiguation,
knowledge\_bridging\}.
\par
(i) MINIMAL\_HINT: a brief recovery cue with flexible wording, conditioned on observable medical tags and the diagnosed local error. Name the evidence or reasoning operation to revisit. Do not include the error-category label or audit explanation. Introduce the cue with a natural transition; no fixed phrase or first-person form is required.
\par
(j) IMAGE\_CHECK, containing applicable (true or false) and target (the
specific modality, anatomy, laterality, finding, relation, or
measurement to re-inspect).
\par
(k) KNOWLEDGE\_SCOPE: the narrow medical principle to recall, or null.
\par
(l) DIFFERENTIAL\_FOCUS: the discriminative axis to compare, or null.
\par
(m) STOP\_CONDITION: the evidence condition under which the continuation
should commit to an answer and stop.
\par
Constraints:
\par
- Do not access hidden disease labels or hidden lesion coordinates.
\par
- Do not state a gold answer, corrected diagnosis, or correct
option letter.
\par
- Do not provide a replacement chain of thought or teacher-authored
solution.
\par
- Do not introduce symptoms, history, tests, measurements, or visual
findings absent from the input.
\par
- Do not discard a valid step merely because its entropy is high, and do
not preserve an invalid step solely for low entropy.
\par
- SAFE\_PREFIX\_END must precede the first consequential failure.
\par
- AUDIT\_WINDOW should normally overlap ENTROPY\_WINDOW, but may deviate
when supported by explicit medical or logical evidence.
\par
- If only final-answer formatting or option selection is wrong and no
reasoning revision is justified, set ACTION to SKIP.
\end{prompt}

\subsection{Stage~2 Continuation Prompt}
\phantomsection\label{app:stage2}

The continuation input contains the original question and an unfinished assistant message with the retained prefix and an error-conditioned recovery cue. Stage 1 constructs the cue from observable medical tags, ERROR\_TYPE, ERROR\_SUBTYPE, and KEY\_ERROR. Feedback determines repair priority; the selected trajectory's own diagnosis determines cue content. The same student continues from this prefilled message.

\begin{prompt}[title={Prompt \thetcbcounter: Stage-2 Assistant-Prefill Continuation}, label=prompt:stage2]
\textbf{User message:}
\\
\{original medical image(s), question, and answer options\}
\\

\textbf{Assistant message:}
\\
\{safe\_prefix\}
\\
\{Natural transition\} \{MINIMAL\_HINT from Stage 1\}
\end{prompt}

MINIMAL\_HINT is inserted verbatim after the retained prefix. The student receives only the cue; error labels and audits remain metadata. Continue immediately after the cue in the same assistant message; preserve the prefix.

The SFT export retains the student-style recovery cue. Repaired and original correct examples use full-response loss in Equation~\ref{eq:main-repair-loss}, supervising all assistant-response tokens. No audit or external hint
is supplied at final evaluation. Record exported examples and token masks
with the \mbox{training~configuration}.

\section{Case Study}
\label{app:specialty}
We present two medical image question-answering cases, including the question, options, and excerpts from the student's initial reading and subsequent continuation, with the hint marked separately. We preserve the wording of the student outputs without language editing. Purple labels separate the original text, hint, and continuation; blue emphasis marks selected observations and reasoning actions in the quoted outputs.

\Needspace{4\baselineskip}
\subsection{Case 08: Chest radiograph}
\label{app:case08}
The initial reading is shown as a failed-rollout excerpt for comparison; it is not the safe prefix retained after rollback. The hint and continuation show how the initial interpretation is revised.
{\raggedright Clinical terms: anteroposterior (AP), electrocardiogram (ECG), \mbox{nasogastric (NG), chest radiograph (CXR), pulmonary embolism (PE), tuberculosis (TB).}\par}
\begin{prompt}[title={Original question and options},colback=myblue_light!35!white,colframe=myblue_dark!55!white,colbacktitle=myblue_light,coltitle=myblue_dark,fonttitle=\bfseries,fontupper=\small,boxrule=0.7pt,arc=2pt]
\raggedright\setlength{\parindent}{0pt}\setlength{\parskip}{2pt}
\begin{minipage}[t]{0.66\linewidth}\vspace{0pt}
A 67-year-old man is recovering in the surgical intensive care unit following an esophagectomy. A portable anteroposterior chest radiograph (Figure A) is obtained to evaluate his postoperative status. The patient has a right-sided chest tube in place. Based on the radiographic findings, which of the following is \mbox{the most likely cause of the bilateral lower lung opacities?\par}

Options: \par
A. Pleural effusion\par
B. Pulmonary embolism\par
C. Emphysema\par
D. Cavitary tuberculosis\par
E. Tension pneumothorax\par
Please reason step by step, and put your final answer within \textbackslash{}boxed\{\}.\par
\end{minipage}\hfill\begin{minipage}[t]{0.30\linewidth}\vspace{0pt}\centering
\includegraphics[width=\linewidth,height=0.22\textheight,keepaspectratio]{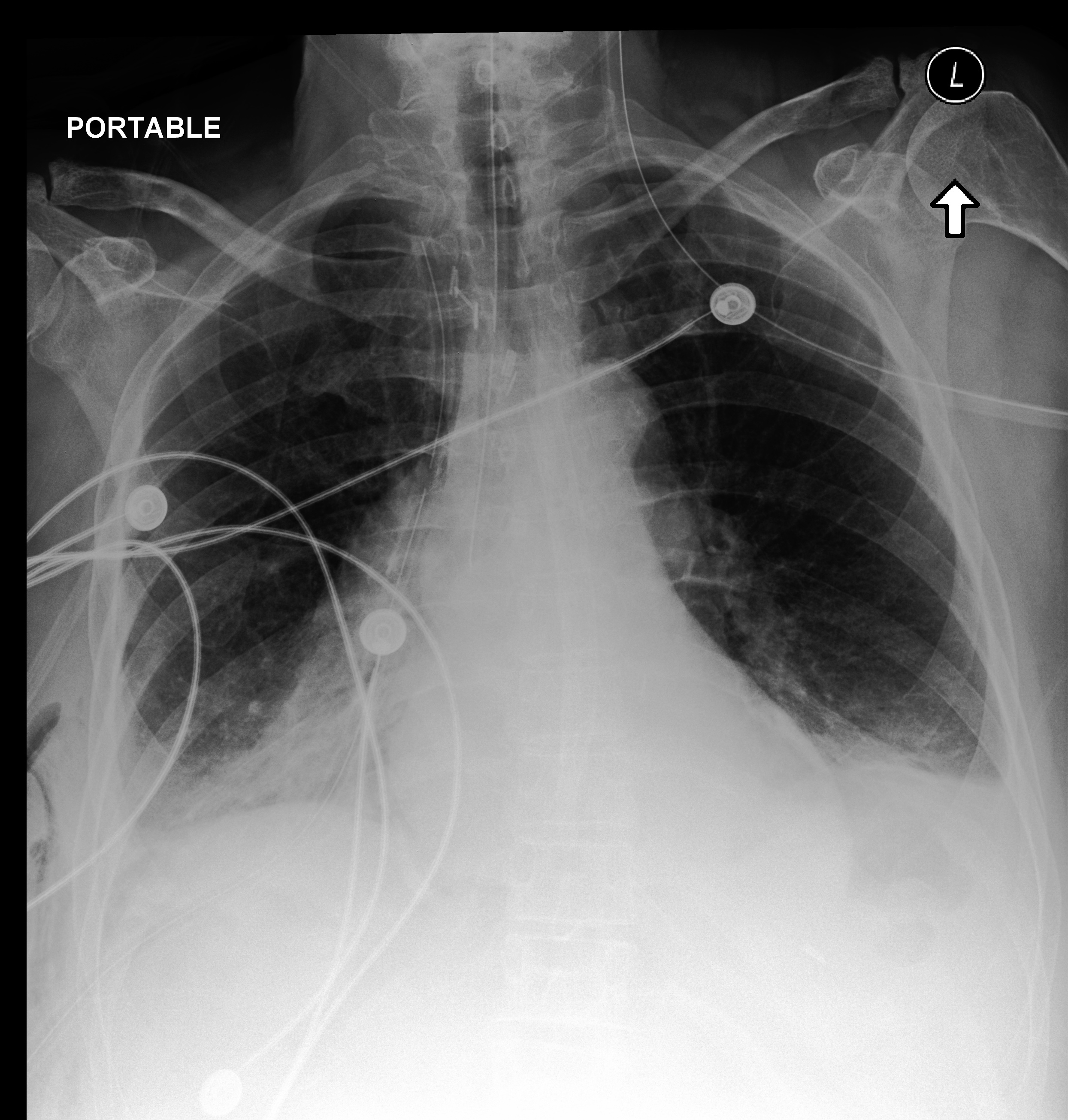}
\end{minipage}
\end{prompt}
\begin{prompt}[title={Student model output},colback=myblue_light!35!white,colframe=myblue_dark!55!white,colbacktitle=myblue_light,coltitle=myblue_dark,fonttitle=\bfseries,fontupper=\small,boxrule=0.7pt,arc=2pt]
\raggedright\setlength{\parindent}{0pt}\setlength{\parskip}{2pt}
\casestage{Initial reading from the failed rollout (before rollback)}
<think>\par
The user wants a step-by-step reasoning and a final boxed answer. Let me look at the radiograph carefully. It's a \appkey{portable AP chest radiograph} of a 67-year-old man post-esophagectomy. There's a right-sided chest tube. There are ECG leads overlying the chest. There's an NG tube or feeding tube coursing down the midline (post-esophagectomy, possibly gastric pull-up). The question asks about \appkey{bilateral lower lung opacities}. Looking at the image, the lower zones do not show convincing increased opacity; the costophrenic angles are not clearly blunted, and there is no definite meniscus-like contour. The hemidiaphragms remain relatively well defined. So pleural effusion is not strongly supported.\par

\par\smallskip{\apptopic{Navigation hint}\par
Before weighing the choices, \appkey{re-check the morphology of the basal opacities}: \appkey{costophrenic-angle blunting, a meniscus-like contour}, and partial obscuration of the hemidiaphragms would localize the process primarily to the...\par
}\smallskip
\casestage{Student continuation after the hint}
\par
Now, \appkey{let me think about the other options}: pulmonary embolism doesn't usually cause bilateral lower lobe opacities (may cause atelectasis or infarction, but not typical). Emphysema causes hyperlucency, not opacity. Cavitary TB causes upper lobe cavities, not bilateral lower lobe opacities. Tension pneumothorax causes hyperlucency and mediastinal shift, not opacity. So the answer is A, pleural effusion.\par
\par
The question says "bilateral lower lung opacities" — in a postoperative patient with a chest tube, bilateral pleural effusions (or a chylothorax after esophagectomy) are common. The radiographic appearance of basal opacities with meniscus and blunting of the costophrenic angles supports pleural effusion.\par
\par
Therefore the\par
\par
\appkey{Let me describe the radiographic findings in more detail}: bilateral lower zone opacities with a meniscus sign, blunting of the costophrenic angles, obscuration of the hemidiaphragms; the chest tube is on the right; the left side shows a similar opacity. No evidence of hyperlucency or pneumothorax; no cavitary lesions; no findings of emphysema (which would be lucent). PE is not a radiographic diagnosis on chest X-ray anyway and wouldn't present as bilateral basal opacities.\par
\par
Therefore the answer is A.\par
\par
I should present step-by-step reasoning: identify the findings, exclude the options, and conclude with A.\par
\par
\appkey{Let me double-check the image description once more}: the left costophrenic angle is blunted with a meniscus; the right base shows opacity consistent with effusion and possibly some atelectasis. The question asks for the "most likely cause" — pleural effusion.\par
\par
Also, after esophagectomy, patients often develop pleural effusions due to chylothorax or transudate from fluid shifts. So A is clinically plausible as well.\par
\par
Format: the answer should be moderately detailed, with step-by-step reasoning, and a final answer. Use headings or bullet points sparingly. Let me write a concise structured analysis: findings, differential exclusion, conclusion. Then give the final\par
\par
Let me write it like this:\par
\par
\appkey{Step 1 — Systematic reading:} portable AP film; cardiac silhouette is difficult to assess; bilateral lower zone opacities; blunting of the costophrenic angles with a meniscus-like upper border; obscuration of the hemidiaphragms; right-sided chest tube (likely for drainage of pleural fluid/air); no evidence of pneumothorax (no visceral pleural line with hyperlucency); no cavitary lesions; lung volumes are not hyperinflated.\par
\par
\appkey{Step 2 — Match with options:}\par
\par
- Pleural effusion: fits — dependent (basal) opacities with meniscus, blunting of the costophrenic angles, common after esophagectomy (including chylothorax).\par
- PE: not visible on CXR; wouldn't cause diffuse basal opacities.\par
- Emphysema: causes hyperlucency, flattened diaphragms — opposite of opacity.\par
- Cavitary TB: upper lobe cavitary lesions, not bilateral basal opacities.\par
- Tension pneumothorax: hyperlucent hemithorax, mediastinal shift away — not present.\par
\par
\appkey{Conclusion: A.}\par
\par
I might also mention that the chest tube on the right is for drainage of pleural fluid/air, and that the left effusion is likely also present.\par
\par
Let me double-check the arrow in the image: it points to the left shoulder/clavicle area — probably just a marker. Not~relevant.\par
\par
Final\par
\par
I'll write the response with clear reasoning. Format: short paragraphs or bullet points. Let me use a structured but concise~approach</think>\par
\par
\appkey{Most appropriate answer:} \textbackslash{}boxed\{A\}.\par
\end{prompt}

\subsection{Case 31: Elbow radiograph}
\label{app:case31}
Clinical abbreviations in this case: anteroposterior (AP) and fall on an outstretched hand (FOOSH).
\begin{prompt}[title={Original question and options},colback=myblue_light!35!white,colframe=myblue_dark!55!white,colbacktitle=myblue_light,coltitle=myblue_dark,fonttitle=\bfseries,fontupper=\small,boxrule=0.7pt,arc=2pt]
\raggedright\setlength{\parindent}{0pt}\setlength{\parskip}{2pt}
\begin{minipage}[t]{0.66\linewidth}\vspace{0pt}
A 32-year-old patient presents to the emergency department with acute right elbow pain, swelling, and limited range of motion following a fall onto an outstretched hand. An anteroposterior radiograph of the right elbow \mbox{is obtained (Figure A). What is the most likely diagnosis?\par}

Options: \par
A. Normal elbow radiograph\par
B. Isolated elbow dislocation\par
C. Elbow fracture\par
D. Advanced osteoarthritis\par
E. Osteochondritis dissecans\par
Please reason step by step, and put your final answer within \textbackslash{}boxed\{\}.\par
\end{minipage}\hfill\begin{minipage}[t]{0.30\linewidth}\vspace{0pt}\centering
\includegraphics[width=\linewidth,height=0.18\textheight,keepaspectratio]{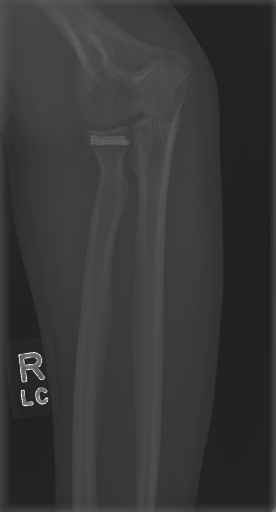}
\end{minipage}
\end{prompt}
\begin{prompt}[title={Student model output},colback=myblue_light!35!white,colframe=myblue_dark!55!white,colbacktitle=myblue_light,coltitle=myblue_dark,fonttitle=\bfseries,fontupper=\small,boxrule=0.7pt,arc=2pt]
\raggedright\setlength{\parindent}{0pt}\setlength{\parskip}{2pt}
\casestage{Original context and initial reading}
<think>\par
The user wants a step-by-step reasoning for a radiograph interpretation question. \appkey{Let me carefully look at the image}. The image is labeled "R" (right) and shows an elbow radiograph. The question states it's an AP radiograph, though the image looks somewhat like an oblique or lateral-ish view, but let's focus on the pathology. Looking at the radiograph: I see the \appkey{distal humerus, the proximal radius, and proximal ulna}.\par
\par

\par\smallskip{\apptopic{Navigation hint}\par
Before committing, \appkey{trace the radial head and neck cortices} to verify that the horizontal lucency causes \appkey{true cortical interruption or fragment displacement} rather than projectional overlap, then \appkey{assess whether the radiocapitellar and ulnohumeral relationships} are...\par

}\smallskip
\casestage{Student continuation after the hint}
\par
\appkey{The radiocapitellar alignment appears normal} (the radial head is opposite the capitellum), and \appkey{the ulnohumeral articulation is also maintained}, suggesting no dislocation. In case of an isolated dislocation, the radial head would be displaced from the capitellum. Here the alignment looks preserved, so an isolated dislocation is unlikely. The presence of a fracture line at the radial head suggests a radial head fracture, so the answer is C (elbow fracture). Osteoarthritis wouldn't present acutely after trauma, and there are no osteophytes. Osteochondritis dissecans is a chronic condition with a loose body, not an acute trauma presentation.\par
\par
Therefore, the most likely diagnosis is C: elbow fracture (radial head fracture). Let's describe the reasoning steps:\par
\par
\appkey{1. Verify patient history:} 32-year-old, fall on an outstretched hand (FOOSH) -> typical mechanism for radial head/neck fractures, coronoid fractures, lateral collateral ligament injuries, etc.\par
\par
\appkey{2. Radiographic findings:} AP view of the right elbow shows a lucent line with cortical disruption at the radial head/neck, with some displacement/fragmentation - consistent with a radial head fracture. The radiocapitellar and ulnohumeral joints remain congruent, excluding dislocation. No evidence of advanced osteoarthritis (no joint space narrowing, osteophytes, subchondral sclerosis). No chronic loose body or cartilage defect suggestive of osteochondritis dissecans.\par
\par
\appkey{3. Therefore the answer is C.}\par
\par
Also, it's worth mentioning that the "fat pad sign" may be present (elevated anterior fat pad), indicating an intra-articular fracture/hemarthrosis, supporting the diagnosis of a fracture. In the image, there may be a faint anterior fat pad sign. But the key is the radial head fracture.\par
\par
Let me double-check the image description once more: the image shows the humerus, ulna, and radius. In the proximal radius region, there's a clear horizontal white line - a fracture of the radial head. The radial head may also be comminuted. So the answer is C.\par
\par
Let me also consider the possibility of a coronoid fracture. Regardless, it's an elbow fracture.\par
\par
Thus, the final\par
\par
Format: The answer should be moderately detailed, including step-by-step reasoning: mechanism, radiographic findings, exclusion of other options, conclusion. Use headings or bullet points. Final\par
\par
Let me carefully describe the image: In the image, the radial head shows a transverse lucency consistent with a fracture; the radial head fragment is slightly displaced; the radiocapitellar relationship is preserved; the ulnohumeral angle is normal; no dislocation.\par
\par
Let me also mention that the lateral fat pad sign supports an intra-articular fracture.\par
\par
Let me write the response.\par
\par
Also, the question states "anteroposterior radiograph," but the image is somewhat lateral; I won't overemphasize the view.\par
\par
Final\par
\par
**\appkey{Step 1 – Mechanism of injury:}** A fall onto an outstretched hand (FOOSH) is the classic mechanism for a radial head/neck fracture, coronoid fracture, or lateral collateral ligament injury in a young adult.\par
\par
**\appkey{Step 2 – Systematic review of the radiograph:}**\par
- **\appkey{Joint alignment:}** The radiocapitellar and ulnohumeral relationships are preserved – the radial head remains opposite the capitellum and the ulna articulates normally with the trochlea. This argues **against an isolated elbow dislocation** (option B), which would show obvious displacement of these relationships.\par
- **\appkey{Bony detail:}** There is a transverse lucent line with cortical disruption and a small displaced fragment at the radial head/neck region – i.e., a **fracture of the radial head**. A faint anterior fat‑pad sign (sail sign) is also suggestive of an intra‑articular fracture/hemarthrosis.\par
- **\appkey{No degenerative changes:}** There is no joint‑space narrowing, osteophyte formation, or subchondral sclerosis, excluding **advanced osteoarthritis** (option D), which would also be unlikely to present as an acute traumatic event.\par
- **\appkey{No chronic lesion:}** There is no well‑circumscribed subchondral defect with a loose body, which would suggest **osteochondritis dissecans** (option E); that condition is chronic/overuse‑related, not acute post‑traumatic.\par
- The film is clearly **not normal** (option A) given the visible fracture line and fragment.\par
\par
**\appkey{Step 3 – Synthesis:}** Acute trauma + FOOSH mechanism + visible fracture at the radial head with preserved joint alignment = a fracture of the elbow (radial head fracture), not a dislocation or degenerative/chronic lesion.\par
\par
**\appkey{Most likely diagnosis: C. Elbow fracture}**</think>\par
\par
\appkey{Most appropriate answer:} \textbackslash{}boxed\{C\}.\par
\end{prompt}

\FloatBarrier

\begingroup
\setlength{\parskip}{2pt plus 1pt minus 1pt}
\makeatletter
\renewcommand\subsection{\@startsection{subsection}{2}{\z@}%
  {-1.5ex\@plus .3ex\@minus .2ex}{.5ex\@plus .2ex}%
  {\normalfont\large\bfseries}}
\renewcommand\paragraph{\@startsection{paragraph}{4}{\z@}%
  {1.2ex\@plus .3ex\@minus .2ex}{-1em}%
  {\normalfont\normalsize\bfseries}}
\makeatother
\Needspace{4\baselineskip}
\section{Additional Related Work}
\label{app:broader-related-work}
This section expands the main-text review of medical evaluation, supervision quality, localized correction, and alternative forms of self-evolution. It also covers complementary work in agent safety, robotics, and data generation.

\subsection{Medical Reasoning, Distillation, and Data Selection}

\appkey{Medical vision-language models} use biomedical instruction tuning~\citep{li2023llavamed,moor2023medflamingo}, generalist reasoning with \appkey{Lingshu}~\citep{xu2025lingshu}, and English--Arabic medical image understanding in \appkey{BiMediX2}~\citep{mullappilly2024bimedix2biomedicalexpertlmm}. Verifier-guided search, knowledge-enhanced synthesis, and reinforcement learning\citep{chen2024huatuogpto1,DBLP:journals/corr/abs-2509-15279,DBLP:conf/acl/LiJZYZW26}, together with parallel reasoning and multi-agent collaboration\citep{DBLP:conf/acl/ChenYXALLZLCY26,DBLP:journals/corr/abs-2602-07905,xia2025mmedagentrl}, improve reasoning capabilities. Process-level diagnostics and distillation audits distinguish answer correctness from reliable visual recognition, knowledge application, and evidence integration\citep{DBLP:journals/corr/abs-2505-11462,DBLP:journals/corr/abs-2606-14697,DBLP:journals/corr/abs-2605-28301}.

MMVistaReason~\citep{lin2026mmvistareason} combines open-data curation with capacity-aware post-training for multimodal reasoning.

\appkey{Medical hallucinations} expose evaluation limits. In controlled experiments, Farrow et al.~\citep{farrow2026rubricsfail} show that clinical hallucinations outside predefined criteria can leave rubric scores unchanged. This motivates explicit factuality checks.

\appkey{Process supervision} provides intermediate-step feedback\citep{ICLR2024_aca97732}, while trajectory synthesis and selection address rationale quality, difficulty, and distribution\citep{DBLP:journals/corr/abs-2601-21821,chandra2025shape}. \appkey{ChainPrune}~\citep{pan2026chainprune} organizes self-generated reasoning paths into a tree and selects paths that reduce redundancy while preserving essential reasoning steps. Its distinction between token length and structural efficiency motivates content-based trajectory selection. \appkey{On-policy distillation} emphasizes compatibility between student states and teacher capabilities\citep{li2026rethinkingopd}; reference-induced shortcuts, reasoning paths, and state coverage also shape supervision value\citep{DBLP:journals/corr/abs-2607-02234,hou2026whatmatters,fu2026rethinkingii}.  \appkey{MedEvo} connects medical error diagnosis with student-centered supervision, guiding correction with minimal hints while preserving student reasoning.

\subsection{Learning from Failed Trajectories and Localized Repair}

Failed trajectories contain reusable reasoning as well as errors. Self-Refine and Reflexion improve subsequent attempts through self-feedback and verbal memory without weight updates~\citep{NEURIPS2023_91edff07,DBLP:conf/nips/ShinnCGNY23}. \appkey{Learning from Mistakes (LEMA)} constructs mistake--correction pairs, \appkey{TrajFusion} combines incorrect and correct trajectories, and Exploring Expert Failures (EEF) extracts useful actions from failed demonstrations~\citep{an2023lema,DBLP:conf/acl/DengTLLWLX26,lan2025eef}. Hindsight Hint Distillation (\appkey{HHD}) and FailForge guide recovery with concise hints or skills instead of full teacher solutions~\citep{DBLP:journals/corr/abs-2605-11556,lv2026failforge}.

Local recovery also depends on where guidance enters and how it becomes supervision. \appkey{Error-localization} studies, backtracking-based distillation, Test-Time Scaling via Error Localization (\appkey{TTEL}), and Selective Regenerative Decoding investigate localization, backtracking corrections, and prefix reuse~\citep{tyen2023errors,DBLP:journals/corr/abs-2605-19433,DBLP:journals/corr/abs-2607-21453,DBLP:journals/corr/abs-2608-24338}. Self-Correction via Reinforcement Learning (\appkey{SCoRe}) learns self-correction through reinforcement learning; Project Aletheia and Pyligent derive supervision from verified backtracking traces or search trees; micro-reflective self-distillation combines micro-reflective trajectories with capability-dependent selection~\citep{DBLP:conf/iclr/KumarZASCSBIBRZ25,dixit2026aletheia,DBLP:journals/corr/abs-2607-07492,DBLP:journals/corr/abs-2606-18844}. EvoThink and Reflective On-Policy Self-Distillation (ROSD) instead use preference learning and reflective on-policy self-distillation~\citep{DBLP:conf/ijcai/DaiXHRJXDDZ26,DBLP:journals/corr/abs-2605-28014}.

 \appkey{MedEvo} uses medically grounded, \appkey{same-student local navigation}: preserving supported prefixes, guiding continuation with gold-blind auditing and tag-conditioned hints, and selecting learnable supervision.

\subsection{Data Evolution and Recursive Self-Improvement}

\appkey{STaR} and ReST$^{\mathrm{EM}}$ select correct self-generated reasoning for supervision\citep{zelikman2022star,singh2023restem}. EvoThink and \appkey{ROSD} use preference learning and reflective self-distillation\citep{DBLP:conf/ijcai/DaiXHRJXDDZ26,DBLP:journals/corr/abs-2605-28014}; \appkey{capability-dependent selection} matches supervision and students~\citep{DBLP:journals/corr/abs-2606-18844}.

\appkey{Recursive self-improvement} examines how feedback changes subsequent learning experience~\citep{duan2026lastrsi}. In clinical interaction, \appkey{MediSkill-Evo}~\citep{wu2026mediskillevo} evolves validated clinical, process, symbolic, and visual knowledge banks without fine-tuning the backbone, using process constraints to guide evidence-grounded actions. \appkey{MedEvo} instead updates model weights through SFT on repaired student trajectories: it diagnoses medical groups and error patterns on a fixed feedback set, then generates and repairs training-pool trajectories for subsequent \appkey{self-distillation}.

\subsection{Failure Feedback beyond Medical Reasoning}
Failure-aware approaches also organize recovery using structured failure feedback in agent safety and robotics~\citep{yin2026fate,yan2026redflow,zhao2026flare}.

\subsection{Analysis of Frontier-Model Reasoning}
Luo et al.~\citep{luo2026capable} study frontier-model reasoning efficiency, step types, and tree structure alongside outcomes.

\subsection{Broader Data and Environment Evolution}
Difficulty-guided chart synthesis and multimodal generation workflows expand the supervision distribution~\citep{DBLP:conf/acl/LiuLWGLCZZPPSHC26,greco2026nemo}. Data-strategy optimization, evolving environments, and autonomous medical curricula provide broader settings for recursive improvement~\citep{DBLP:journals/corr/abs-2607-25886,zheng2026dreamrsi,DBLP:journals/corr/abs-2608-16211}. MedEvo evolves data for medical reasoning repair and supervision.

\FloatBarrier
\section{Additional Baseline Details}
\label{app:baseline-details}
Tab.~\ref{tab:main} compares external models and trajectory-supervision baselines.

\subsection{Backbones and External Comparison Models}
We use Qwen3.5-9B, scaling to Qwen3.5-35B and Qwen3.5-397B~\citep{qwen3.5}. A Base result refers to the corresponding initial checkpoint before MedEvo training. Later-round models are trained from that \appkey{same initial checkpoint} using the updated supervision dataset; the current checkpoint generates its training trajectories.

The external comparison set includes general-purpose models and models specialized for medicine. Medical models include LLaVA-Med~\citep{li2023llavamed}, HuatuoGPT-Vision~\citep{DBLP:journals/corr/abs-2406-19280}, BiMediX2~\citep{mullappilly2024bimedix2biomedicalexpertlmm}, Lingshu~\citep{xu2025lingshu}, and HealthGPT~\citep{DBLP:conf/icml/LinZLYY0H00ST0L25}. InternVL2.5 and InternVL3 provide general-purpose multimodal comparisons~\citep{DBLP:journals/corr/abs-2412-05271,DBLP:journals/corr/abs-2504-10479}. Tab.~\ref{tab:main} groups models by proprietary access and open-source parameter scale. These comparisons contextualize performance; they differ from MedEvo's \appkey{controlled ablations}.

\subsection{Trajectory-Supervision Baselines}
\paragraph{Baseline 1: HHD.} Hindsight Hint Distillation uses concise hints to scaffold recovery~\citep{DBLP:journals/corr/abs-2605-11556}. Its role in the comparison is to assess hint-based supervision alongside MedEvo's \appkey{medical diagnosis, prefix preservation, and feedback-driven data updates}. The method follows the cited work; Tab.~\ref{tab:main} reports our medical-benchmark evaluation.

\paragraph{Baseline 2: TrajFusion.} TrajFusion combines incorrect and correct trajectories to construct training supervision~\citep{DBLP:conf/acl/DengTLLWLX26}. It provides a comparison for learning from failed reasoning. MedEvo instead constructs a \appkey{corrected suffix} from the same student's retained prefix and uses medical feedback to prioritize later repairs.

\paragraph{Baseline 3: Teacher SFT.} This baseline uses SFT on supervision generated by Qwen-3.8-Max. It compares \appkey{teacher-generated supervision} with MedEvo's repaired student trajectories. This isolates supervision.

\subsection{Reading the Comparisons}
The medical icon identifies models specialized for medicine, and the repair icon identifies trajectory-repair methods. Blue shading identifies MedEvo. Bold and underlining indicate the highest and second-highest distinct scores within each model group and column, with ties sharing a rank. The \appkey{seven-benchmark average} weights each benchmark equally. It is distinct from repair success, the fraction of submitted questions corrected.

Component ablations test full rewriting, suffix continuation, medical-tag feedback, and error-type feedback. Their purpose is to isolate design choices within the reported setup. Training configuration and information-access rules are provided in Appendices~\ref{app:manifest} and~\ref{app:permissions}, respectively.

\endgroup

\end{document}